\documentclass{article}
\usepackage{amsmath, amsthm, amssymb}
\usepackage{adjustbox}
\usepackage{graphicx}
\usepackage{subcaption}     
\usepackage{relsize}
\usepackage{multicol}
\usepackage{booktabs}
\usepackage{geometry}
\usepackage{xcolor}
\usepackage{url}

\newcommand{\remove}[1]{}

\title{
Towards Combinatorial Interpretability of Neural Computation  
}
\author{Micah Adler \\ {\it \small MIT} \\ \tt \footnotesize micah@csail.mit.edu  \and Dan Alistarh \\ {\it \small ISTA 
\& Red Hat AI
} \\ \tt \footnotesize Dan.Alistarh@ist.ac.at  \and Nir Shavit \\ {\it \small MIT \& Red Hat AI
} 
\\ \tt \footnotesize shanir@mit.edu}

\date{}

\usepackage{xcolor}

\begin{document}

\maketitle
\begin{abstract}
\smaller[1]

We introduce {\em combinatorial interpretability}, a methodology for understanding neural computation by analyzing the combinatorial structures in the sign-based categorization of a network's weights and biases.  We demonstrate its power through \textit{feature channel coding}, a theory that explains how neural networks compute Boolean expressions and potentially underlies other categories of neural network computation. According to this theory, features are computed via feature channels: unique cross-neuron encodings shared among the inputs the feature operates on.  Because different feature channels share neurons, the neurons are polysemantic and the channels interfere with one another, making the computation appear inscrutable.

We show how to decipher these computations by analyzing a network's feature channel coding, offering complete mechanistic interpretations of several small neural networks that were trained with gradient descent. Crucially, this is achieved via static combinatorial analysis of the weight matrices, without examining activations or training new autoencoding networks. Feature channel coding reframes the superposition hypothesis, shifting the focus from neuron activation directionality in high-dimensional space to the combinatorial structure of codes.
It also allows us for the first time to exactly quantify and explain the relationship between a network's parameter size and its computational capacity (i.e. the set of features it can compute with low error), a relationship that is implicitly at the core of many modern scaling laws. 

Though our initial studies of feature channel coding are restricted to Boolean functions, we believe they provide a rich, controlled, and informative research space, and that the path we propose for combinatorial interpretation of neural computation can provide a basis for understanding both artificial and biological neural circuits.

\end{abstract}

\section{Introduction}

An exciting recent body of research in the interpretability of neural networks has been focused on viewing their computation as collections of independent features whose directions in high dimensional activation space can be analyzed. This has led, using novel approaches such as the use of sparse autoencoders \cite{anthropic2023autoencoder}, to the discovery of polysemanticity, the superposition hypothesis \cite{toy}, and a body of empirical evidence of the correlation between features and directionality of neuronal activation space vectors \cite{Ameisen2025,anthropic2025circuit,Bricken2023,Cunningham2024,transcoders,Dunefsky2024,ge2024automatically,anthropic2023autoencoder,Rajamanoharan2024}.  Those techniques (surveyed in detail in the related work section) have provided ingenious ways of capturing the representation of features and some information on their interdependencies, but so far have not elucidated the full dependency graph, the underlying function being computed nor the mechanism for computing it.

In this paper, we take a combinatorial approach to interpretability, rather than one based on the geometry of the activation space. As we explain below, we focus on the computation of Boolean expressions, a meaningful class of computation that makes it easier to take first steps into interpreting combinatorial representations. We use it to propose a theory explaining how neural networks compute (at least for Boolean formulae) by showing how features and their computation are represented as combinatorial structures encoded in the weights and biases of the trained network. In particular, this theory offers explanations of how polysemanticity and the superposition hypothesis arise through codes associated with features.
 
 In this combinatorial view, we focus much less on the actual values involved in these learned parameters and more on the sign-based categorization (positive, negative, or zero) of those values.  The two views into neural computation (the existing vector space one and the new combinatorial one) are not mutually exclusive, but we show here that the combinatorial approach, at least in some cases, provides insights into a number of important facets of interpretability that are harder to analyze with the vector space approach, including the logic of the circuits that compute the features.  We also demonstrate, currently in a number of small examples, but hopefully in many more elaborate ones in the future, that the combinatorial approach can be used to extract the  features and their exact computation directly from the structure of the learned parameters of the network. This is done without resorting to training sparse autoencoders, transcoders, or similar approaches \cite{anthropic2025circuit,transcoders,ge2024automatically, anthropic2023autoencoder}.  In other words, our work aims to reveal the exact low level mechanics of the computation, the {\em how}, not just the when and where of the relationships among features.

At the heart of our combinatorial approach is what we call the {\em feature channel coding hypothesis}. This hypothesis specifies both how a feature is represented (which has a natural analogue to the superposition hypothesis), as well as how computation proceeds between those features (which does not).  In feature channel coding, a feature is represented in a layer of the network as a code using neurons at that layer. The individual "bits" of the code specify the neurons whose weights bring about an activation for that feature in that layer.  If the codes for any pair of features at a given layer do not overlap much, the corresponding activation vectors (where precise values are taken into account) are almost-orthogonal, similar to the superposition hypothesis. 

The real differentiator of feature channel coding is its theory of how computation proceeds.  To compute a feature $f_i$, all of the features from the prior layer of the network needed for that computation are mapped to the feature code for $f_i$.  Then, the network computes $f_i$ individually on each neuron of the code.  For example, if $f_i = x_{i_1} \land x_{i_2}$ (where $x_{i_1}$ and $x_{i_2}$ are Boolean variables) then, since $x_{i_1} \wedge x_{i_2}$ can computed as ReLU$(x_{i_1}+x_{i_2}-1)$, that computation is performed on every neuron of the code for $f_i$.  When the right combination of features is present at the previous layer to activate $f_i$, each of the neurons of the code for $f_i$ becomes active through their individual computations, and as a result, $f_i$ itself will be active.  When a feature at the prior layer is used for multiple features at the current layer, it gets mapped to each of the codes for the features it is used in.  Note that the channel's code is somewhat unrelated to the function being computed by the feature - the goal of the channel's code is to bring together the inputs that need to be combined to compute this function. There may be noise on each channel due to the overlap between codes, but as long as the overlap is small, the computation still produces a result that approximates the correct code sufficiently well.  Uncovering the codes for a given neural network will allow us to describe its ``circuit,'' and to do so statically directly from the weight matrices of the network, without learning the directionality of neurons in activation space.  We provide a more in depth description of feature channel coding below.

Here is a brief summary of the results of this paper.  We demonstrate the effectiveness of combinatorial interpretability through  contributions that can be broadly categorized into three areas:

\begin{enumerate}
\item {\it Establishing the Feature Channel Coding Hypothesis and Demonstrating its Manifestation}.

We introduce the feature channel coding hypothesis, proposing that neural networks internally represent and compute with features using a specific, interpretable coding scheme. We provide substantial evidence supporting this hypothesis, showing its emergence in MLP networks trained via backpropagation and gradient descent.  Most of this evidence is for single hidden layer networks, where we show the following problems exhibit feature channel coding:
\begin{itemize}
    \item Disjunctive Normal Form (DNF) Boolean formulas with only positive variables.
    \item DNF formulas with both positive and negative variables.
    \item Formulas consisting of just the OR of Boolean formulas.
    \item A counter-intuitive example of a network learning Conjunctive Normal Form (CNF) formulas via double application of DeMorgan's Law to arrive at a DNF-like coding.
    \item A one-dimensional analogue of a vision task, demonstrating the potential breadth of this coding phenomenon.
\end{itemize}
We also demonstrate how to fully extract feature channel codes, and the computations they enable, from the weight matrices of the networks computing DNF Boolean formulas with positive variables as well as from the weight matrices of the network computing the one-dimensional vision task.  These extractions provide a complete mechanistic interpretation of the network's computation directly from its weights.

\item {\it Quantifying Feature Channel Coding, Scaling, and the Learning Process}.
We quantify the frequency with which the combinatorial structures of feature channel coding appear in the weight matrices.  By demonstrating that these structures appear much more often than random, we provide further evidence for feature channel coding.  We also apply this combinatorial approach to quantitatively analyze scaling laws based on feature coding. This yields, to our knowledge, the first concrete mechanistic explanation for why model accuracy can break down as problem complexity (e.g., the size of the underlying Boolean formula) increases: there is a combinatorial limit to the ability to ``pack in'' the necessary structures for feature channel codes, leading to representational collapse.  The network also does not seem to have a ``plan B'' when increasing complexity makes feature channel coding no longer possible.  These results suggest that tracking feature coding across scales could lead to more refined scaling laws. Furthermore, we use this quantitative approach to demonstrate how feature channel codes emerge dynamically during gradient descent training, showing a gradual structuring of weights from initial randomness. This observational capability opens avenues for quantitatively studying the impact of various training paradigms like batch normalization and dropout on code formation. 

\item {\it Introducing a Framework for Deeper Networks}.
All of the above results are for networks with a single hidden layer.  To facilitate the analysis of deeper networks, we propose a novel disentangled perspective on neural network computation, building on techniques from \cite{adler2024}.  This framework conceptualizes each layer's weight matrix ($W_i$) as a product of two distinct matrices: $D_i$, which transforms the previous layer's feature codes into a monosematic representation, and $C_i$, which projects each monosematic feature into its feature channel codes required by the features of the subsequent layer.  Ignoring non-linearities and biases, the network's transformations can thus be viewed as a sequence $W_d \cdot \ldots \cdot W_2 \cdot W_1 = C_d D_d \cdot \ldots \cdot C_2 D_2 \cdot C_1 D_1$. This perspective not only clarifies why methods like sparse auto-encoders and transcoders \cite{anthropic2025circuit, transcoders, anthropic2023autoencoder} are effective (as they approximate the decoding matrices $D_i$) but, more critically, enables direct combinatorial analysis of the weight matrices themselves without requiring auxiliary trained models. Building on this, we introduce {cascading feature disentanglement}, a process to incrementally reverse this matrix sequence combinatorially. This yields a decodable representation of the feature channel codes used by the $C_i$ matrices, starkly contrasting with the often indecipherable patterns in raw weight matrices ($W_i$) or activation space vectors, and making it possible to combinatorially analyze the weight matrices directly.  We show (admittedly somewhat limited) preliminary work on applying cascading disentanglement by decoding single hidden layer MLPs with an added initial embedding layer, as well as single hidden layer MLPs with multiple output neurons.
\end{enumerate}

There were several reasons we chose to focus on neural computing of Boolean formulas. One of the motivations for the feature channel coding hypothesis comes from a few theoretical papers on computing Boolean functions in superposition (including one by these authors) \cite{adler2024,CircuitsSuperposition2024,vaintrob_superposition}. Our paper \cite{adler2024} specifically provides a feature channel coding algorithm for computing the AND function, and therefore we know that there exist provably accurate feature channel coding algorithms for AND and other Boolean functions for any size network and any number of features subject to certain asymptotic bounds.  The initial question we asked in this work (and have answered in the affirmative) was whether a neural network ever converges during training to find anything similar to that algorithm.

Perhaps more importantly, Boolean formulas are a versatile, predictable, and quantifiable laboratory for studying neural networks.  Generating training and testing examples is trivial, it is straightforward to quantify dataset sizes, and many of the problems that are encountered with training real world problems (over-fitting / generalization, etc.) show up and again are simple to quantify. Moreover, with Boolean formulas, one can accurately control the number of features and the complexity of their interactions, a crucial tool we use to analyze scaling laws. 

As the reader will see, in the networks we studied, we found strong evidence that features are computed using soft Boolean logic \cite{softboolean}. In soft Boolean logic, values for $x_i$ and $x_j$ are reals, and $x_i \wedge x_j$ for example is computed as ReLU$(x_i+x_j-b)$, where $b$ is not 1 but rather a real number sized by the variables to deliver a 0 or positive outcome. ReLUed dot products computed by neurons are a natural way of computing soft (and therefore differentiable) Boolean computations. 

More speculatively, this leads us to hypothesize that soft Boolean formulas, though they do not account for all neural computation, are at the heart of many of the more complex computations where neural networks are used.  
This may also be true for biological neurons after some initial layers translating analog data to soft Boolean computation.  

The rest of this paper is organized as follows. In Section~\ref{section: related work} we add some broader background on interpretability as it relates to our work. In Section~\ref{section: intro to coding} we begin with a simple example of feature channel coding and how to interpret it combinatorially. After that, in Section~\ref{section: scaling}, we show a detailed quantitative analysis of the underlying combinatorial structures of feature channel coding, and demonstrate how they explain scaling and scaling laws.  In Section~\ref{section : further features}, we show various additional examples of feature channel coding and how one can interpret networks combinatorially.  Following that, Section~\ref{section: theory} presents a formal model of combinatorial interpretability that lays out the foundation of how one could systematically disentangle a multi-layer superposed neural network, together with our first steps towards such disentanglement.  In Section~\ref{section multioutput} we examine slightly more complex networks, and present results for single layer networks with multiple outputs.  We finish by laying out directions for future research.

\section{Background and Related Work}
\label{section: related work}

Early research in neural network interpretability has treated neurons as geometric vectors in activation space, visualizing the set of inputs that activate a given neuron as a single ``semantic'' direction, e.g.~\cite{nguyen2016multifaceted, simonyan2013deep}. However, it has quickly become clear that many neurons are \emph{polysemantic}, in the sense that they correspond to multiple unrelated features or concepts, e.g.~\cite{cunningham2023sparse,toy, goh2021, scherlis2022}, making interpretability difficult~\cite{goodfellow2016,Belinkov}. 
This phenomenon has been linked to neural networks that pack more ``features'' into their latent space than there are neurons, via \emph{superposition}, e.g.~\cite{adler2024,hanni2024mathematical,  CircuitsSuperposition2024,vaintrob_superposition}. One hypothesis is that in superposition, features are not assigned unique basis vectors; instead, they are encoded in combinations of neurons.


Polysemantic neurons have been empirically observed in various architectures, significantly in larger models where the efficiency of representation is paramount \cite{goh2021}. For example, neurons in multimodal models like CLIP respond to disparate concepts such as images of spiders and textual references to ``Spider-Man,'' illustrating the complex and overlapping feature encoding by these neurons \cite{goh2021}. Insights by Elhage et al.~\cite{toy} and Scherlis et al.~\cite{scherlis2022} indicate that overlapping features are a deliberate strategy to utilize limited neuron capacity efficiently, packing more information into fewer dimensions.  Explicit combinatorial coding was theoretically analyzed in~\cite{adler2024, hanni2024mathematical}, and the construction of feature channel coding was described and theoretically analyzed for the case of pairwise AND in \cite{adler2024}. This differs from sparse coding and similar approaches~\cite{olshausen1997sparse} by having the computed function be similar across the code neurons in the feature channel and having the code serve as a combinatorial way of reducing interference. 
Scherlis et al.~\cite{scherlis2022}  quantified this trade-off with the notion of capacity allocation, where they found that the network ``decides'' how many dimensions (fractionally) to give each feature. 




Recent advances in the use of autoencoders for network interpretability have led to enhanced methods for isolating and manipulating feature directions \cite{anthropic2025circuit,transcoders,ge2024automatically,anthropic2023autoencoder}. In these approaches, features in neural networks are represented as vectors in activation space, where each dimension can encode different aspects of the input data.  This vectorial representation supports complex feature combinations, essential for the network’s ability to generalize from seen to unseen data \cite{bengio2013}. The techniques allow the creation of attribution graphs that map the dependencies among features for the computation of a specific input \cite{anthropic2025circuit,cunningham2023sparse,transcoders}. 

Yet, the circuits extracted from neural networks using methods such as sparse autoencoders (SAEs) \cite{cunningham2023sparse} and transcoders \cite{transcoders} are best described as targeted approximations of the original model, rather than elucidating the underlying computational processes. Specifically, the resulting circuits capture which features influence subsequent features {\em for a specific input}, but they do not capture (a) the overall feature dependence (across all inputs), (b) the function that is being computed to realize that dependence, and (c) the underlying mechanism for computing that function.

Due to their nature, SAE-like techniques require new trained components  which inherently simplify or omit certain aspects of the original model’s complexity. 
This includes focusing on a subset of features, possibly ignoring less interpretable yet significant elements, and applying transformations like gating or thresholding that modify the original computational pathways. Furthermore, these methods often involve training new versions of network components---like using transcoders to replace dense layers with sparser ones. While this helps in understanding how inputs are processed, it is fundamentally a reconfiguration rather than a direct observation of the original network’s internal representation. In addition, recent work has raised doubts concerning the generalization ability of such  approaches~\cite{heindrich2025sparse}. Consequently, while these extracted circuits are highly valuable for gaining insights into model behavior and assisting in debugging or educational purposes, they do not fully capture all the intricate details and interactions of the original model. In contrast, the idea behind our combinatorial interpretabiliy approach, though at a much earlier phase than the study of SAEs \cite{cunningham2023sparse} and transcoders \cite{transcoders}, is to directly interpret the computation from analyzing the structure of the model itself. 

Techniques such as principal component analysis (PCA) and canonical correlation analysis (CCA) have also been employed to analyze and optimize feature alignments,  simplifying the feature landscape by reducing polysemanticity and enhancing clarity in the roles of individual neurons \cite{kornblith2019, raghu2017}. However, the overlapping or intersecting of these feature vectors can complicate the isolation and interpretation of individual features \cite{goodfellow2016}.


Other prior work has turned to Boolean function tasks as a laboratory for  interpretability, in particular tasks in disjunctive normal form (DNF) and Conjunctive Normal Form (CNF) tasks. 
Mechanistic interpretability analyses have shown that, surprisingly, even without an architecture explicitly designed for logic, networks learn internal circuits that resemble symbolic algorithms. Specifically, Palumbo et al.~\cite{palumbo2024mechanistically} trained a transformer to solve 2-SAT and reverse-engineered the network’s algorithm. 
Nanda et al.~\cite{nanda2023progress} analyzed small networks trained on modular arithmetic and identified the exact algorithm to be a form of discrete Fourier transform to add numbers.

More broadly, Tishbi et al. \cite{tishby2015deep} proposed the Information Bottleneck theory, which offers a principled approach to understanding how neural networks compress input data into compact and relevant representations. Carleo et al. have employed sophisticated statistical mechanics approaches to dissect how correlations between features in input data influence the structure and dynamics of the learned representations in neural networks \cite{carleo2019machine}. This body of work complements our empirical findings by highlighting the theoretical underpinnings and potential limitations of using superposition as a feature encoding strategy in neural networks.

The superposition hypothesis is also linked to  scaling laws in neural networks, initially identified by Rosenfeld et al. \cite{rosenfeld2019scaling} and later popularized by Kaplan et al. \cite{kaplan2020scalinglaws}. These laws suggest that as networks scale—increasing in model size, data size, and computational resources—they tend to exhibit more distinct and specialized neuron representations, which might reduce the prevalence of polysemantic neurons \cite{olah2023superposition, kaplan2020scalinglaws,rosenfeld2019scaling}.

\section{An Introduction to Feature Channel Coding}
\label{section: intro to coding}

This section introduces feature channel coding and shows how it can be used for combinatorial interpretability. 

\subsection{Explicit Feature Channel Coding}

The technique of feature channel coding was first suggested in \cite{adler2024}, where it was used to provide an explicit, theoretical construction of a multi-layered perceptron (MLP) that computes multiple pairwise ANDs simultaneously.  The resulting algorithm demonstrates that in theory, feature channel coding is effective, and can be used to produce neural networks that are provably correct.  We here provide a simplified version of the technique from \cite{adler2024} to serve as a more in-depth introduction into how feature channel coding works.

\begin{figure}[h]
  \centering
  \begin{subfigure}[b]{0.9\textwidth}
    \centering    \includegraphics[trim=0in 1in 0in 1.65in, clip=true, width=\textwidth]{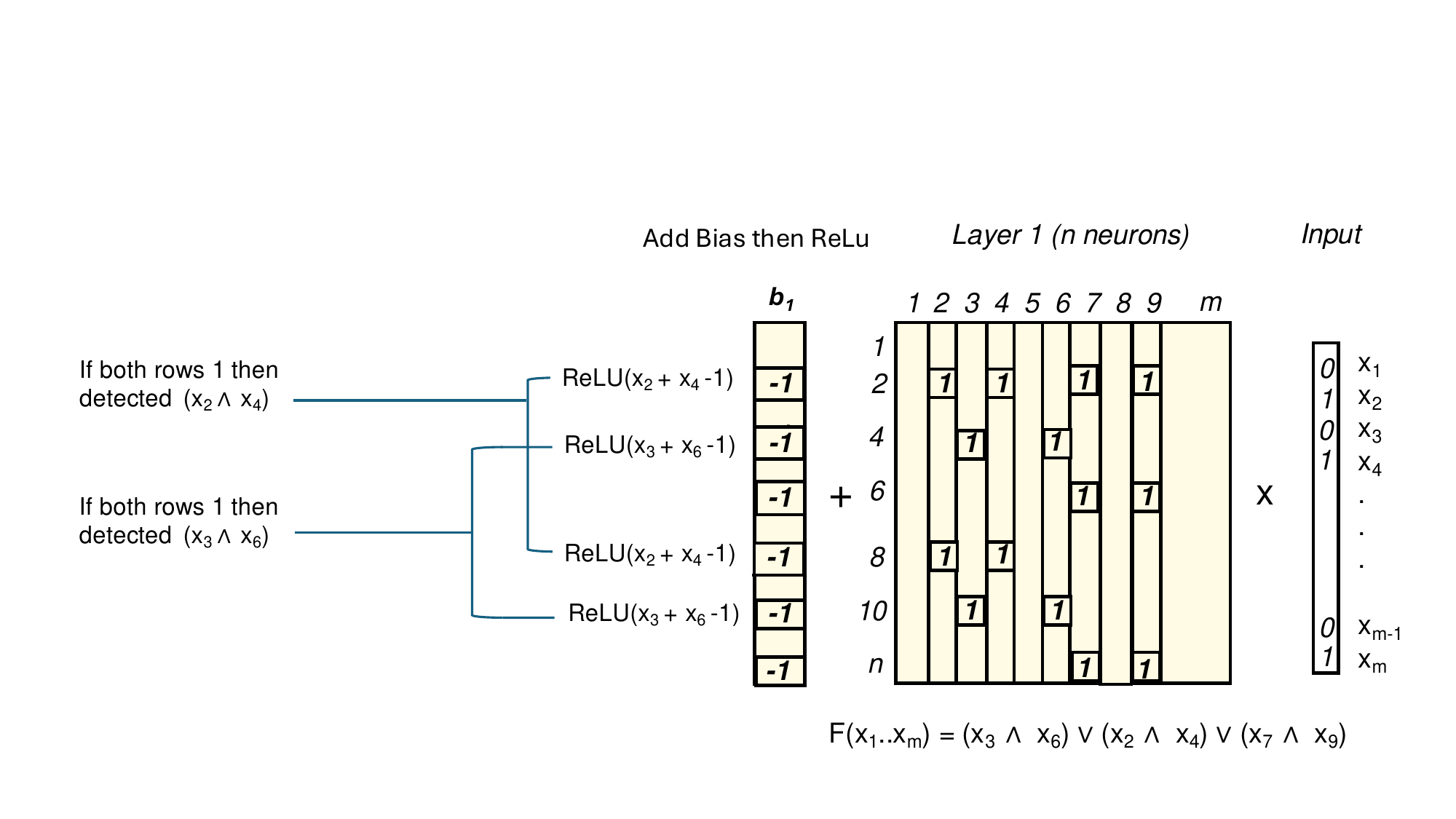}
  \end{subfigure}
  \caption{A neural network that computes $(x_3 \land x_6)$, $(x_2 \land x_4)$, and $(x_7 \land x_9)$ using Feature Channel Coding for each AND \cite{adler2024}.}
  \label{feature channel NN1}
\end{figure}

Figure~\ref{feature channel NN1} shows a schematic of this use of feature channel coding. This first layer of a neural network computes the three following Boolean expressions: $(x_3 \land x_6)$, $(x_2 \land x_4)$, and $(x_7 \land x_9)$, given inputs to the network that are Boolean sequences of 0's and 1's representing the truth values of $x_1,\ldots, x_m$. So if locations 3 and 6 in the input are $1$, we would expect the network to detect that $(x_3 \land x_6)$ is satistied. 

Consider how the feature $x_3 \land x_6$ is computed. The network does this by using a code for the two inputs $x_3$ and $x_6$ in columns 3 and 6 of the weight matrix. Specifically, the code places a 1 value in both of those columns for the two neurons 4 and 10.  All other entries in the two columns are 0.
The key to feature channel coding is that those two codes are the same (columns 3 and 6 are identical), and so the set of rows (neurons) where 1s appear in those columns can be used to compute the AND of those two inputs.    Consider now what happens when the input consists of $x_3 = x_6 = 1$ (True), and all other variables are 0.  When all $n$ neurons of the layer compute the dot product with this input, neurons 4 and 10 will produce a sum of 2 each (before the bias) while all other neurons will produce 0. If we then add the bias of $-1$ and perform a ReLU for those neurons, we obtain a 1, as desired.  More generally, for any combination of 0s and 1s for inputs $x_3$ and $x_6$, both neuron 4 and neuron 10 will produce a post ReLU value of $\mbox{ReLU}\,(x_3+x_6 - 1) = x_3 \land x_6$.  We can think of the code consisting of those two neurons as serving as a ``computational channel'' for computing $x_3 \land x_6$, and this can be detected by the next layer of the network (which is not depicted in the figure).

Similarly, we use a code to detect the output feature $(x_2 \land x_4$) and that code does not overlap with the code for $(x_3 \land x_6)$. However, as the number of features grows, the codes will begin to overlap, as is the case of the codes for $(x_2 \land x_4)$ and $(x_7 \land x_9)$, which overlap in neuron 2. This introduces "noise" into the computational channel, and for that reason it can be helpful to use larger codes (i.e., codes with 1s in a larger number of neurons), so as to reduce the impact of this noise.  \cite{adler2024} shows how to select random codes that will create feature channels that provably avoid having too much noise from overlaps so as to ensure there are no false positive detections and that the signal of the correct feature codes is not too diminished.

\subsection{Neural Networks Learn Feature Channel Coding}

One of the main open questions from \cite{adler2024} was whether these kinds of codes occur naturally in networks trained via gradient descent. In other words, does an SGD trained neural network learn to use feature channel coding?

We here demonstrate that this does actually happen: a similar algorithm emerges naturally from training using gradient descent.  Let us turn to a specific example of this. We constructed a two layer MLP, with 16 inputs, 16 neurons at the first layer, and a single neuron at the second layer, with a Layer 1 bias but no Layer 2 bias.  We then generated a Boolean formula as the hidden function for the network to learn; this was chosen randomly from the set of pairwise ANDs of 16 Boolean variables with the variables paired up so each appears in exactly one AND. We here examine the result (which was typical of all such formulas) of the formula 
$$f(x_0,\ldots,x_{15}) = (x_3 \land x_{11}) \lor (x_9 \land x_{13}) \lor (x_2 \land x_6) \lor (x_0 \land x_4) \lor (x_5 \land x_{12}) \lor (x_7 \land x_{15}) \lor (x_8 \land x_{10}) \lor (x_1 \land x_{14}).$$ 
We trained the network on this formula, with random inputs, each chosen with equal probability of being a True or False instance of the formula.  There were 30,000 random inputs chosen.  Each input had between two and six True variables (with equal probability).  The training process utilized the BCE loss function and the Adam optimizer with a learning rate of 0.01. Training was conducted over 2000 iterations over the training data (epochs).

\begin{figure}[t]
  \centering
  \begin{subfigure}[b]{1.1\textwidth}
    \centering
    \includegraphics[trim=1in 0in 0in 0in, clip=true, 
    width=\textwidth]{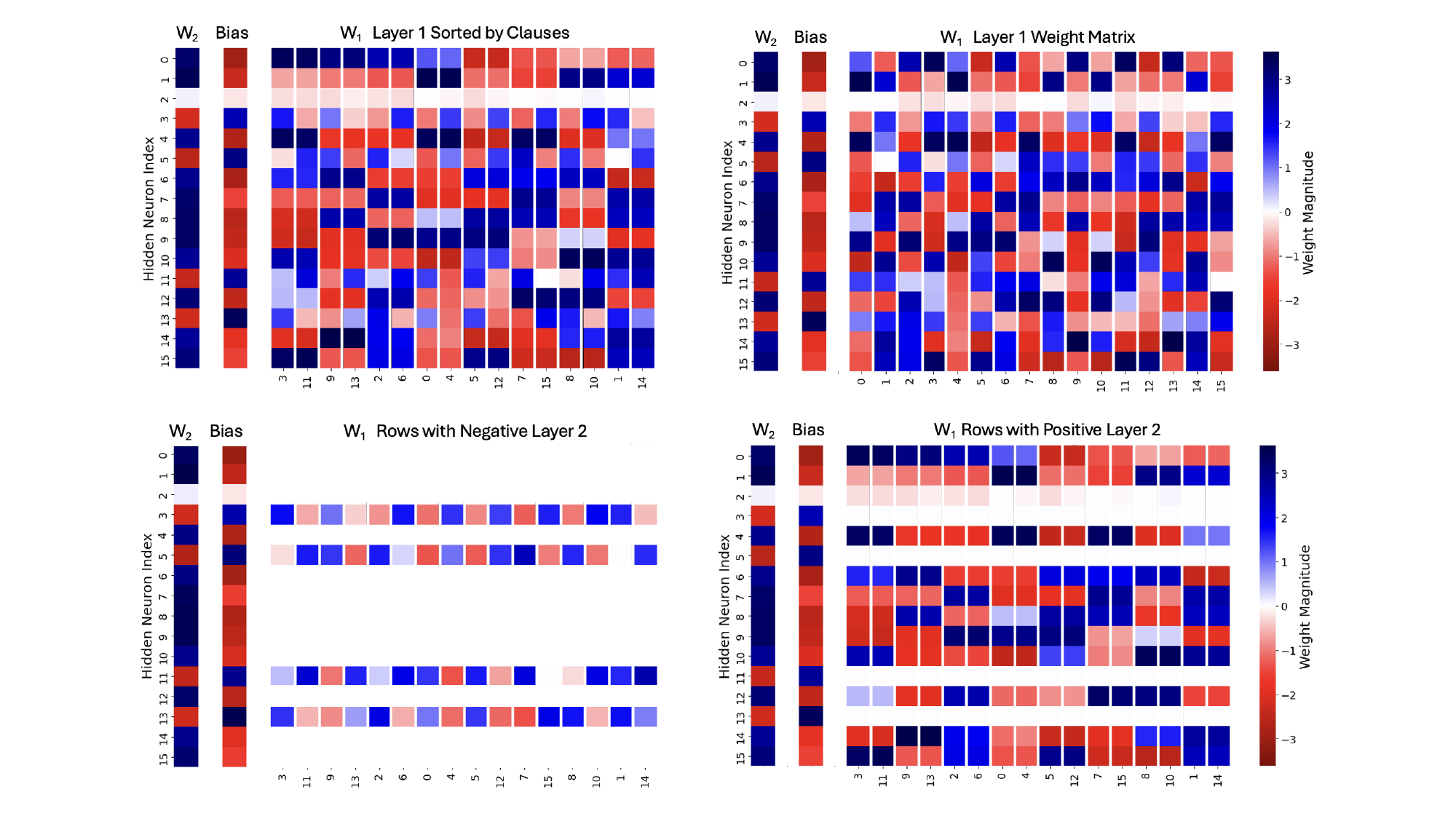}
  \end{subfigure}
  \caption{All weights and biases for a neural network trained on the Boolean formula\\ $(x_3 \land x_{11}) \lor (x_9 \land x_{13}) \lor (x_2 \land x_6) \lor (x_0 \land x_4) \lor (x_5 \land x_{12}) \lor (x_7 \land x_{15}) \lor (x_8 \land x_{10}) \lor (x_1 \land x_{14})$}
  
  \label{paired}
\end{figure}
In the upper right hand corner of Figure~\ref{paired}, we present the full set of learned parameters of one representative training run of the neural network: Layer 1 weights ($W_1$), Layer 1 bias, and Layer 2 weights ($W_2$). $W_2$ is placed upright for convenience. The colors correspond to the sign of the weights, the bluer they are the more positive, the redder the more negative. Initially, let us focus on $W_1$.  The weight matrix in the figure is in its raw form without alteration. Its columns correspond to input features, the input values of the variables $x_i$, and its rows correspond to the neurons in this first hidden layer. The matrix $W_1$ does not seem to reveal much, other than the fact that neuron 2 does not really participate in the computation.  In the upper left corner, we sorted the columns of $W_1$ by the features (variables) so that the pairs that are joined by an AND are next to each other. So for example, columns 3 and 11 are next to each other because $(x_3 \wedge x_{11})$ is a clause in the Boolean expression the network is trained on. This image starts to reveal that there is something interesting going on with these pairs: in many positions, but not all, adjacent columns have the same values.

In the lower right hand corner, features in $W_1$ are sorted the same way, but now we only consider the rows (neurons) of the weight matrix $W_2$ that correspond to positive Layer 2 weights. We will justify this filtering below.  This image is striking in that in each and every row, the columns corresponding to an AND feature pair seem to be virtually identical. Moreover, if we look at the positive, blue colored weights, we see an emerging pattern. 
Looking across all pairs of input features (variables), every pair that appears in the same AND output feature is in fact brought together into a feature channel code.  That set of neurons allows the AND feature's computation to propagate to the next layer.

For example, in the underlying computation, input features 3 and 11 participate in the computation of the feature $(x_3 \land x_{11})$.  Thus, they are mapped to the same feature channel code, which consists of positive weights on the set of neurons $\{0, 4, 6, 10, 12, 15\}$.  Feature $(x_9 \land x_{13})$ is represented by the code $\{0, 6, 8, 14\}$, and feature $(x_5 \land x_{12})$ by the code $\{6, 8, 9, 10, 15\}$.  

However, the neural network in Figure~\ref{paired} does not compute $x_i \wedge x_j$ using exact Boolean logic as in our theoretical example, but rather using soft Boolean logic via ReLU$(x_i+x_j-b)$ for some $b$ (regular Boolean variables would have $b=1$).  We see in the lower right hand side that for every single row with a positive Layer 2 weight, the bias is significantly negative, representing the $-b$.  Thus, like in the pure Boolean algorithm we showed, if two input features in the same AND, say $(x_3 \land x_{11})$, are both set to True then their sum (calculated by the dot product of each neuron by the input) will make it through the negative bias and the ReLU as a positive value for all the rows of the code.  Consequently, the result from the network will be a "True" after the sigmoid. 

Note that this network is polysemantic.  For example, neuron 4 reacts positively to four different AND features: $(x_3 \land x_{11})$, $(x_0 \land x_{4})$, $(x_7 \land x_{15})$, and $(x_1 \land x_{14})$.  Thus, the result of neuron 4 will be similar when the input consists of 1s for only variables $x_3$ and $x_{11}$ (a True instance) to what it is when the input consists of 1s for only variables $x_3$ and $x_4$ (a False instance).  This is where feature channel coding plays a crucial role. The outcome of the computation for a given AND output feature is dependent on all of the feature's coding rows, and these codes are learned in such a way that even if they overlap in some rows, they don't overlap in too many of them. For example, the codes for inputs $x_3$ and $x_4$ overlap in neurons 0 and 4, and so when both are 1s they will produce some positive signal due to these two rows, but not nearly as much as when $x_3$ and $x_{11}$ are both 1s, since the codes for those two inputs overlap in 6 rows.

However, since some positive signal does get through even with the False instance, we see that the negative rows do play a role in the computation.  We provide quantitative evidence of this below for various different problems; the degree of importance of the negative roles seems to depend on the specifics of the problem being computed.  Here, we see that a strong negative signal is provided by neuron 3 in the case where the input is 1s only on variables $x_3$ and $x_4$. 
Specifically, neuron 3 passes a positive signal to Layer 2 because row 3 is positive in both column 3 and column 4, and that positive value, when multiplied by the negative value in $W_2$, results in a strong negative value received by the second layer output neuron, and this offsets the positive signal received from neurons 0 and 4.  However, when the input is 1s only on $x_3$ and $x_{11}$, that negative signal is greatly diminished by the fact that row 3 of the Layer 1 weight matrix has opposite signs in columns 3 and 11.   This explains why the network has a strong learned preference for the inputs that appear in the same clause to have opposite Layer 1 weights for negative rows.

With this as motivation, let us get back to the separation of the weight matrix into the rows that have positive Layer 2 weights and those that have negative Layer 2 weights (bottom of Figure~\ref{paired}).  In the network, the weights of the Layer 2 neuron $W_2$ defines whether a row of the $W_1$ matrix contributes to the final result as a positive or a negative value. Essentially, the positive $W_2$ weights correspond to features in Layer 1 that we call  {\em positive witnesses}: they provide evidence that the input is a 1 (True) instance and provide a strong positive signal to support that.  Negative $W_2$ weights are {\em negative witnesses} and provide a negative signal due to evidence that the input is a 0 (False) instance. The lower left image in Figure~\ref{paired} shows that the negative witnesses are essentially computing an XOR: if one variable of a pair of variables is present but not the other to offset it, that provides a signal that this is a No instance.

As we discuss in the next section, feature channel coding allows the neural network to compute in ``superposition'' \cite{adler2024}, and in a different example than the one presented in this section, the number of features encoded can be much larger than the number of neurons. 

\subsection{Combinatorial Interpretation Using Feature Channel Coding}
We next point out that our techniques go beyond just understanding the structure of the underlying network, and in fact can, at least in some cases, be used to interpret the computation explicitly.  
From Figure~\ref{paired}, it seems likely that we can determine the underlying randomly chosen Boolean formula from only the learned parameters of the network.  Here we address the question of whether we can do this for formulas chosen from the distribution described for Figure~\ref{paired}, although we found the technique we describe to be effective across a broad range of formula distributions.  We demonstrate that even just the Layer 1 weight matrices are sufficient to reconstruct the entire formula, as long as the trained network was provided with enough training samples to generalize fairly well.  In fact, we found that a straightforward algorithm allows us to reconstruct the formula (for this 2-variable AND special case): we take the absolute values of all entries of the Layer 1 weight matrix, and then find for each column of the resulting matrix which other column it is most closely correlated with. If column $i$ is most closely correlated with column $j$, then we assume that input $i$ is paired with input $j$ in the Boolean formula.  Note that this algorithm does not necessarily return symmetric results.

\begin{figure}[h!]
\centering
\makebox[\textwidth][c]{%
\includegraphics[width=0.6\linewidth]{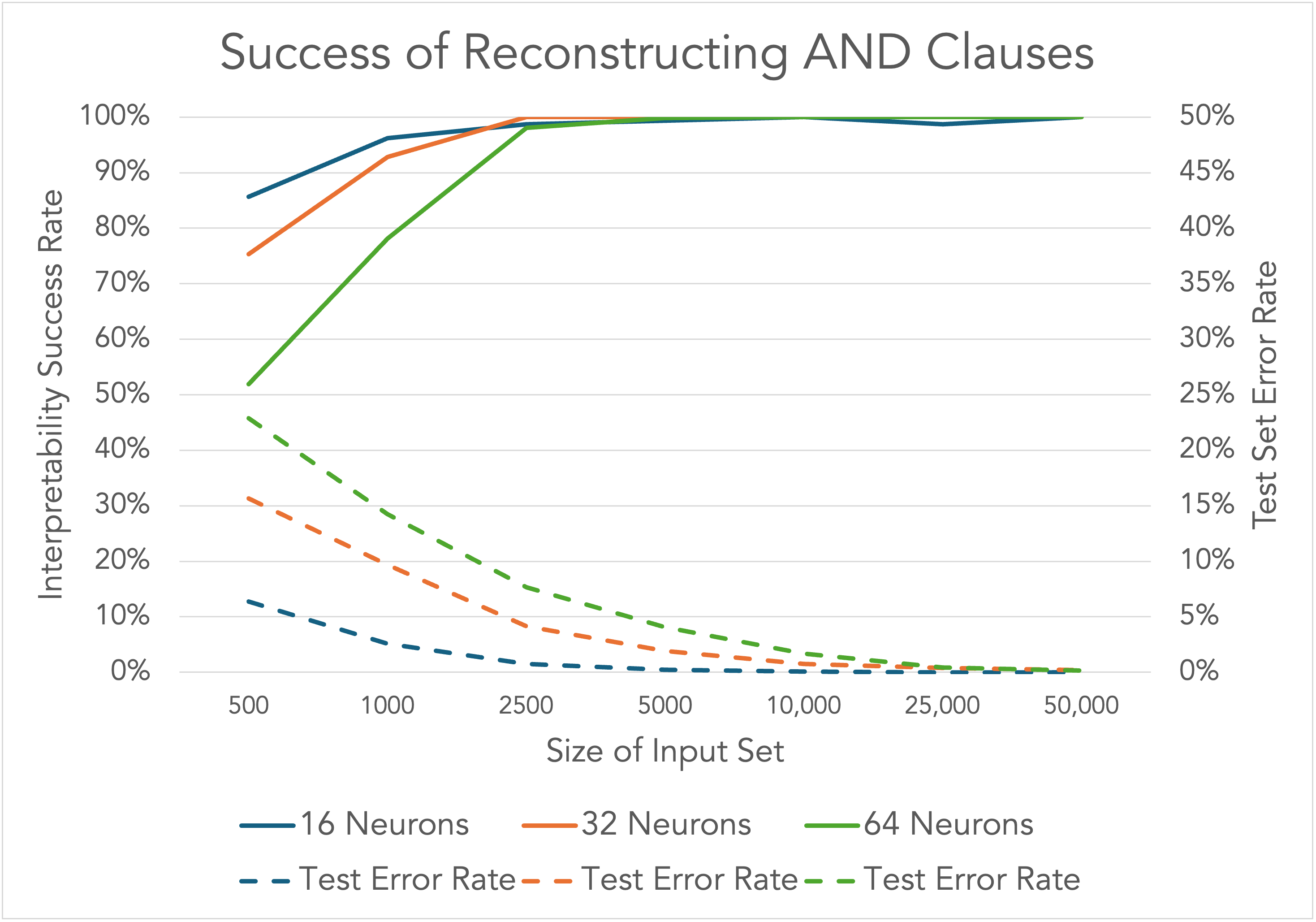}
}
\caption{Results on our algorithm to recreate a Boolean formula from its Layer 1 weight matrix.}
\label{correlation test}
\end{figure}

The results of our tests on this algorithm are depicted in Figure \ref{correlation test}.  We tested a range of sizes of the hidden layer (and corresponding size of Boolean formula), as well as a range of sizes of the training set.  For each of these, we chose ten random formulas, and then trained each formula on a set of random inputs.  We then performed two tests: we measured the error rate of the trained network on a new test set of inputs, chosen from the same distribution as the training set, and we tested the accuracy of the correlation of absolute values algorithm.  Our measure of success for the correlation algorithm was how many of the pairings it was able to correctly identify.  We see that for sufficiently large sets of training data, our algorithm was almost always able to reconstruct the entire formula correctly.  Furthermore, the measure of "sufficiently large" here seems to tie closely to the network correctly learning the function (as measured by the test set): when there was enough training data for the error rate on the test set to be small, the correlation algorithm performed nearly perfectly.

Putting all this together, the gradient descent training process of the network has learned to perform ANDs of the input channels in the first layer, and an OR of the AND clauses coming out of the first layer in the second layer. 

\subsection{Polysemanticity and Superposition}

Let us touch a bit more on the notions of polysemanticity and superposition \cite{adler2024,toy,hanni2024mathematical,olah2023superposition}. When one looks at the set of neurons of a feature channel code as in our example above, we see that there is overlap. The same neuron is polysemantic and fires in response to multiple features. The accepted idea for how superposition is implemented is that neurons represent features using polysemanticity. To quote \cite{toy}: ``in the superposition hypothesis, features are represented as almost-orthogonal directions in the vector space of neuron outputs. Since the features are only almost-orthogonal, one feature activating looks like other features slightly activating.'' This is true, and yet our theory of feature channel coding suggests that superposition can also be explained by the combinatorial property of using feature channel codes that overlap. The noise of one feature slightly lighting up others is explained not just by alignment in vector space, but also by the use of multiple rows in the feature channel coding so the relative importance of a given neuron is lowered \cite{adler2024}. Moreover, we don't need to decipher directionality in activation space or geometric representations in order to map the features captured by a collection of neurons: the feature channel coding in the weight matrices allows us to do it combinatorially without any activation information.  Furthermore, as we will see in Section~\ref{section: scaling}, using the combinatorial interpretation also allows us to analyze the code pattern distributions of networks to explain scaling laws in ways that seem impossible to capture by looking at features through their directionality in activation space. 

Getting back to our example network, if overlapping codes introduce noise, given that our network was trained on 8 clauses using 16 neurons in the first layer, the training could have found a coding that had one code row per clause: one monosemantic neuron per feature. In this way, there would be no overlaps among codes, and thus no cross-feature noise.  However, as in the matrices resulting from our training, we see that the training settled on multiple overlapping codes and the neurons are polysemantic. In other words, polysemanticity appears even when there are fewer features than neurons. 

We believe that there is a combinatorial explanation for this. Gradient descent searches the state space for local minima.  There are many more encodings of the desired computation using overlapping codes than using non-overlapping monosemantic ones, and thus likely many more local minima involving feature channel coding using multiple overlapping polysemantic neurons.  Thus, gradient descent is more likely to find such an encoding than the monosemantic one, and then not be able to move from that local minimum.  This is somewhat reminiscent to the findings of \cite{Lecomte2024WhatCauses}.  
As witnessed in our example, coding (and with it polysemanticity) does appear in cases where one does not need to compute in superposition.  It would be interesting to obtain more evidence or a proof of this hypothesis, which could be done using an approach similar to the one we take in Section~\ref{section: emerge}. 

We will provide other examples of feature channel coding in the context of other training problems in Sections~\ref{Section: 3P1N} and \ref{section : further features}.  For example, the interested reader can skip to take a look at an example of a one dimensional``vision problem'' in Section~\ref{section: vision}. Before we do so, we provide a quantitative analysis of feature channel coding. This will provide numerical evidence of feature channel coding and also allow us to better understand the constraints imposed by this technique.  These constraints in turn will allow us to provide a combinatorial explanation of a scaling law in neural networks.

\section{Combinatorial Analysis of a Scaling Law}
\label{section: scaling}

In this section we use our combinatorial interpretability framework to analyze scaling laws around both {\em how} a model computes Boolean formulas, as well as {\em why} the model breaks down as a result of increasing complexity. At the core of our analysis is a study of the strong correlation between a model's accuracy and its ability to encode features.  This analysis serves a number of purposes:
\begin{itemize}
    \item It provides quantitative and convincing evidence that feature channel coding is being used by networks trained via gradient descent to evaluate Boolean formulas.
    \item It provides insights into the constraints faced by such systems, and why and when they break down.  
    \item It provides an example of the power of the combinatorial approach to analyze such systems.
\end{itemize}

For this analysis, we will focus on DNF formulas consisting entirely of 4-variable clauses.  We will consider two different cases of such formulas: one where all variables appear in their positive form (this section), and one where each clause has 3 positive variables and 1 negated variable (Section~\ref{Section: 3P1N}).  One of our primary goals is to see how the system reacts to increasing complexity of the formula being learned.  To study this, we start with a formula with a small number of clauses, and gradually increase the number of clauses in the formula, until the point the model is no longer able to learn the formula accurately.  Note that this does not change the size or structure of the network or the size of its inputs in any way.  It merely creates a Boolean formula with a more complex underlying truth table for the model to learn.

In more detail, in all experiments, the network consists of 32 input variables and a single output neuron, computed via a sigmoid.  In between is a single hidden layer, and we test $j = 16, 32$, and $64$ neurons in that layer.  Each hidden layer neuron has a bias followed by a ReLU, and we use an output layer bias as well, 
and the hidden layer is fully connected to the inputs, and (post ReLU) is fully connected to the output layer.  We test the following numbers $k$ of clauses: 2, 4, 8, 12, 16, 24, 32, 42, 56, and 64. 

In all cases, we average results over 10 trials, each with a different random Boolean formula (an OR of clauses in 32 variables). For each trial, we train for up to 100 epochs (with patience‑based early stopping) on 20,000 samples. The samples are drawn from the following distribution: with 50\% probability, we assign a label of 1 (True), randomly select one of the formula’s clauses, set its variables to satisfy that clause, and then choose between zero and two additional variables to set to 1 while preserving the clause’s satisfiability. Otherwise (label 0, a value of False), we again pick a clause but now deliberately “break” it so that only three of the four variables are set to satisfy that clause.  We also again choose additional variables to set to 1, so that the total number of set variables will be between four and six while the value will still be False. 
We use the Adam optimizer with a learning rate of 0.001, a batch size of 64, and a patience of 10 epochs for early stopping. We note that the distribution matters quite a bit for the overall numbers of patterns we report on, but not the general scaling trends that we describe. 

Our analysis of these cases are depicted in the figures below.  In Figure \ref{training_error}, we examine training error for the three sizes of the hidden layer for the case of clauses of four non-negated variables, and see that it starts to accumulate when the number of clauses reaches the number of neurons in the hidden layer.  In the subsequent analysis, we will look at how the network is using feature channel coding, and why that leads to that number of clauses being the starting point of training error.

\begin{figure}[hbt!]
\begin{center}
\includegraphics[width=0.7\textwidth]{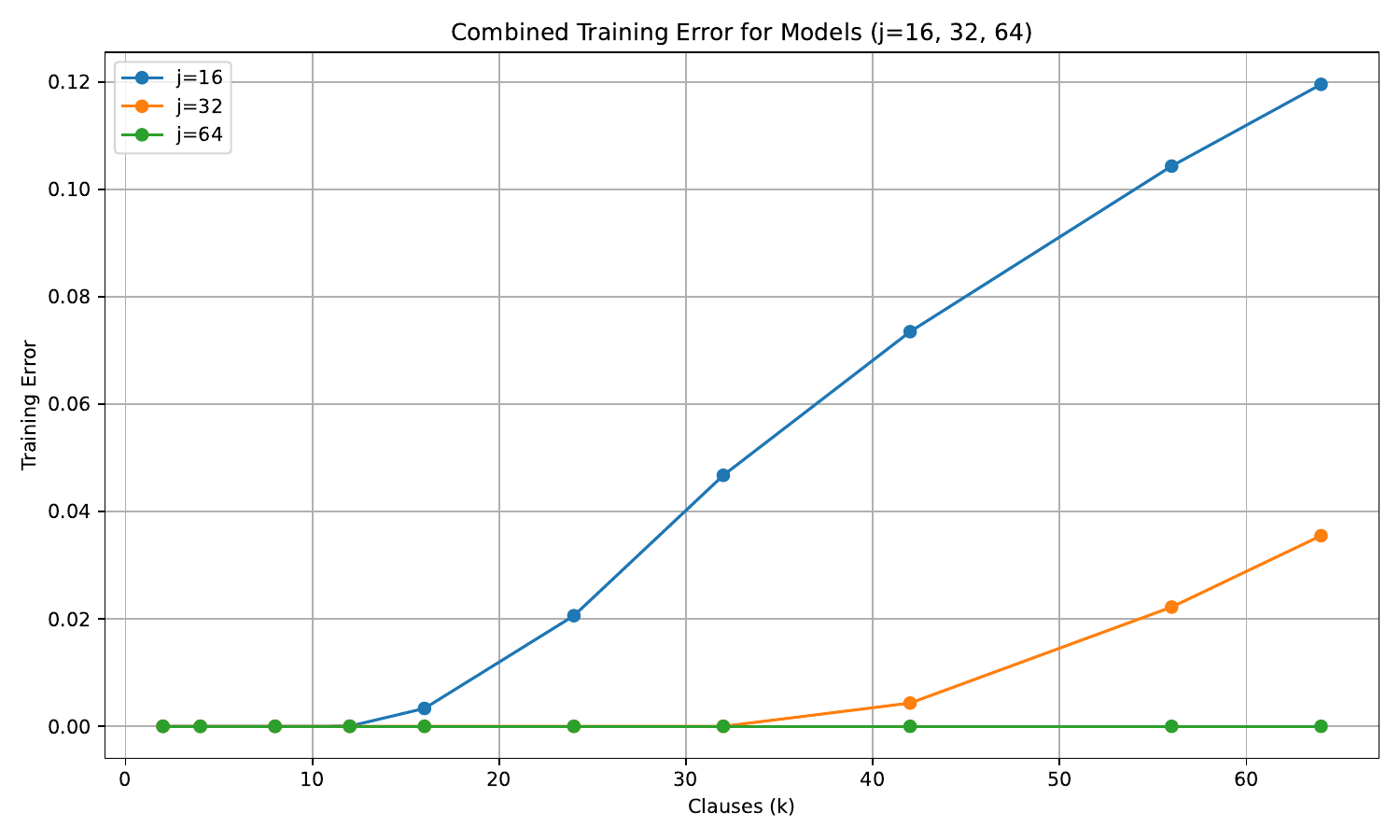}
\caption{Onset of training error as number of clauses increases.}
\label{training_error}
\end{center}
\end{figure}

\begin{figure}[hbt!]
  \centering
  \begin{subfigure}[b]{0.32\textwidth}
    \centering
\includegraphics[width=\textwidth]{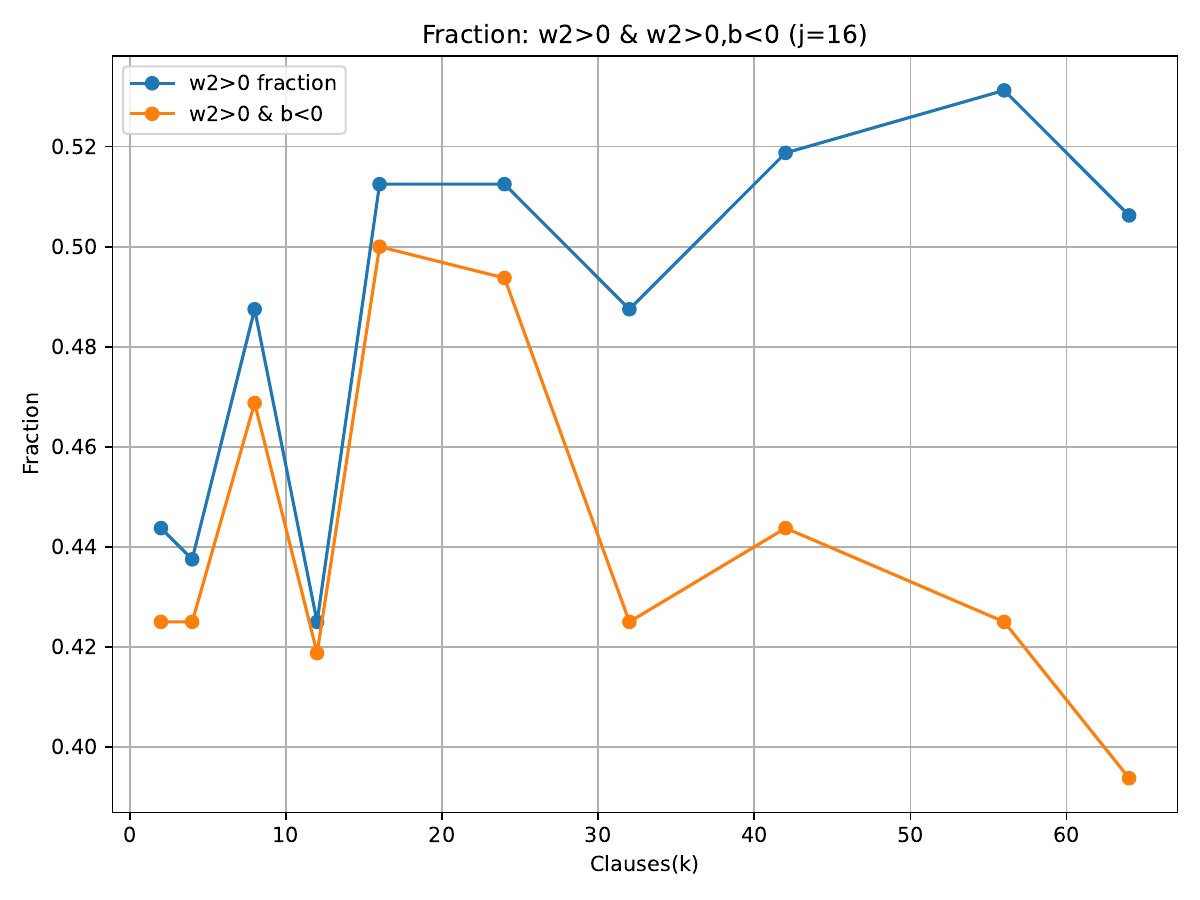}
  \end{subfigure}
  \begin{subfigure}[b]{0.32\textwidth}
    \centering
\includegraphics[width=\textwidth]{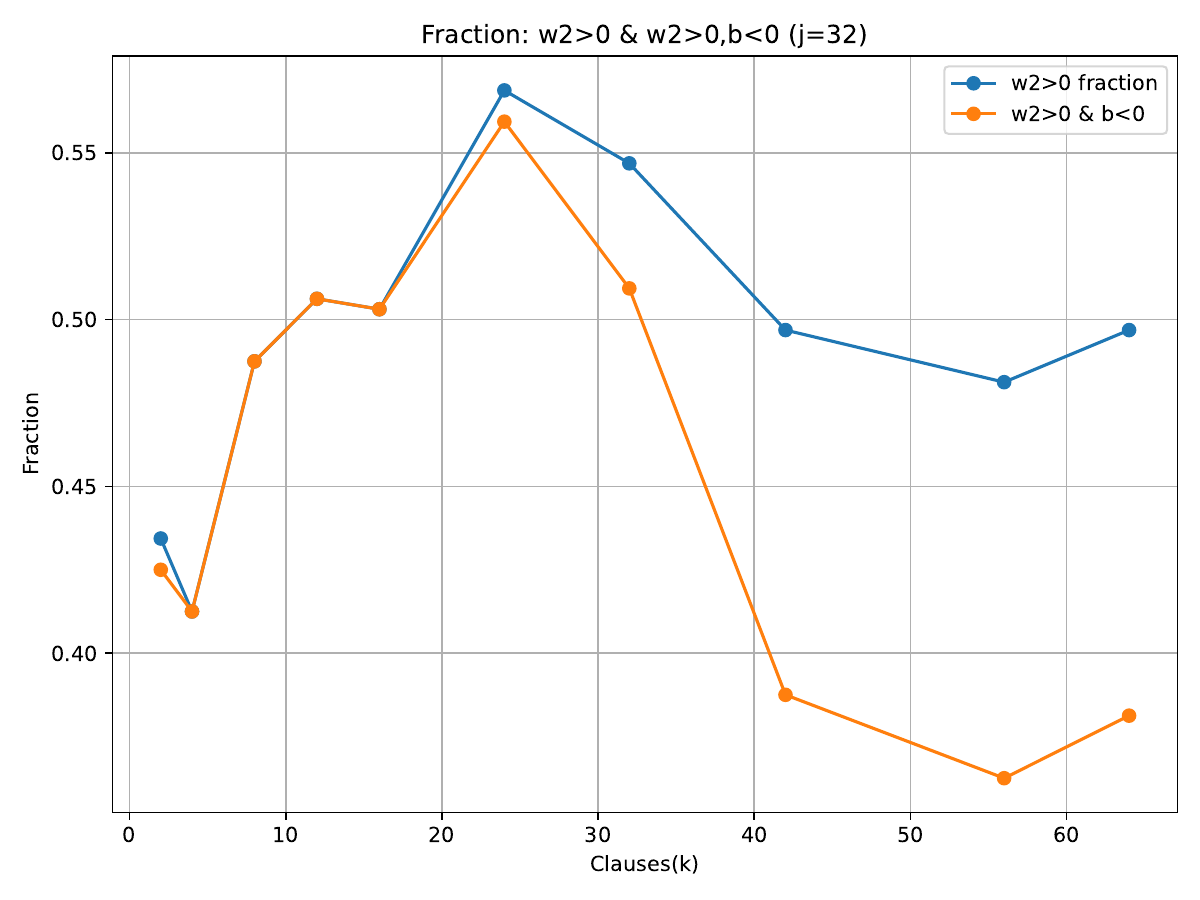}
  \end{subfigure}
  \begin{subfigure}[b]{0.32\textwidth}
    \centering
\includegraphics[width=\textwidth]{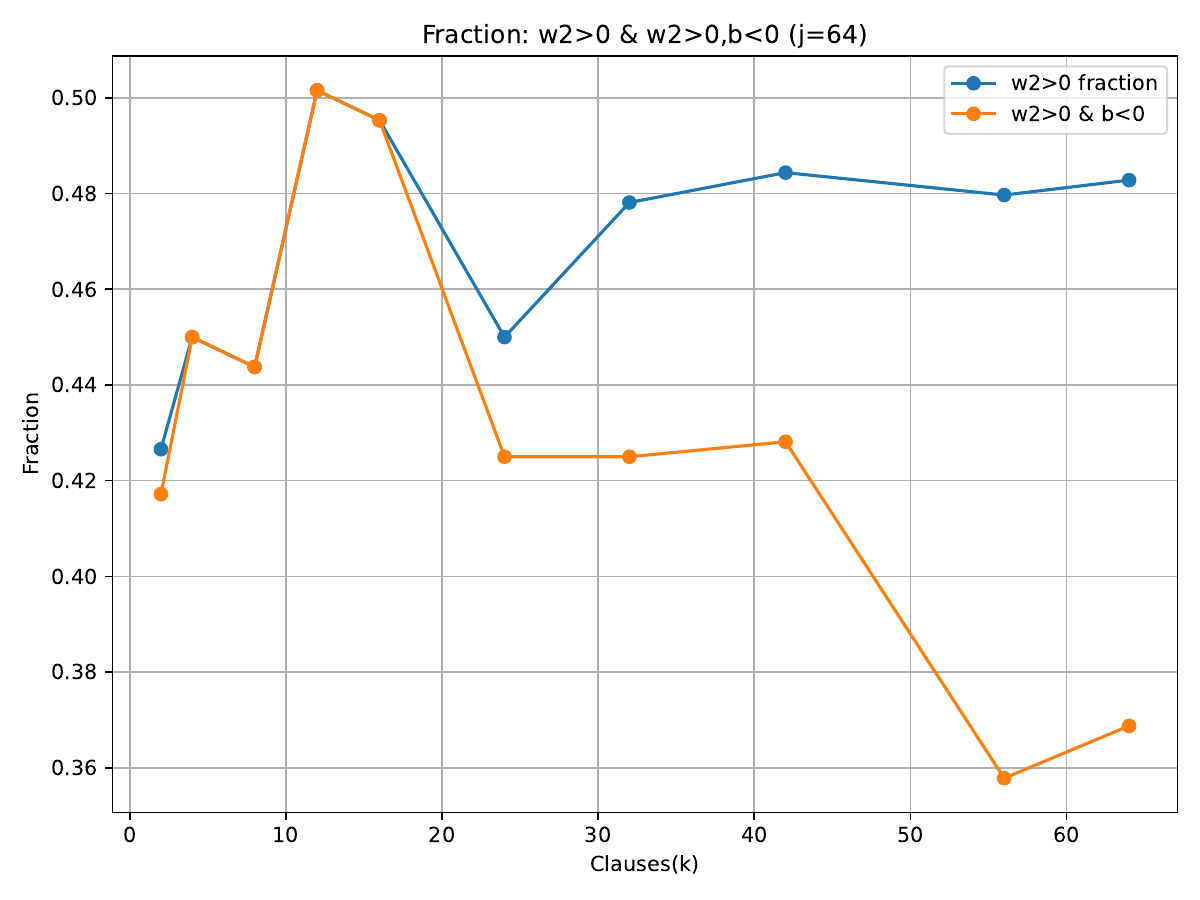}
  \end{subfigure}
  \caption{Fraction of positive Layer 2 weights and negative Layer 1 bias.}
  \label{negative_rows}
\end{figure}

Next, in the three versions of Figure \ref{negative_rows}, we graph the percentage of hidden layer neurons that are connected to the output via positive weights.  As we saw in Figure \ref{paired}, the sign of that weight differentiates how the model uses that neuron, for positive or negative witnesses.  We see that in these tests, the learned model does not stray far from having half of the neurons in positive rows and half in negative rows.  In other experiments, not described in this paper, we see that the fraction of negative and positive rows is strongly influenced by the fraction of positive and negative examples in the training set.  For example, if the training set has 90\% positive examples instead of 50\% positive examples, there will be very few negative rows.  This is perhaps not that surprising if viewed through the lens of gradient descent: positive examples can only either increase or not change Layer 2 weights of our network, and negative examples can only decrease or not change Layer 2 weights.  Fully understanding and quantifying this phenomenon is an interesting open problem, but for this work, we maintained a balanced distribution of positive and negative examples.

Figure \ref{negative_rows} also  depicts how many of the positive rows also have a negative bias (which is what feature channel coding would predict).  Interestingly, we see that for the (simpler) functions with a smaller number of clauses, the system does indeed have all (or almost all) of the positive Layer 1 neurons associated with a negative bias.  However, as the function becomes more complex, more and more of the second layer neurons have positive bias.  And this does not line up with the point where the network training accuracy falls off - it happens around 15 to 20 clauses for all three values of $j$, and so seems independent of the size of the hidden layer.  It is also past the point where accuracy falls off for 16 hidden neurons, but before it for the two cases of a larger hidden layer, and so it seems that having negative bias is helpful, but not crucial to learning the Boolean function.  We will discuss below a possible reason for why the percentage of positive rows with negative bias decreases the way it does.

\begin{figure}[hbt!]
  \centering
  \begin{subfigure}[b]{0.32\textwidth}
    \centering
\includegraphics[width=\textwidth]{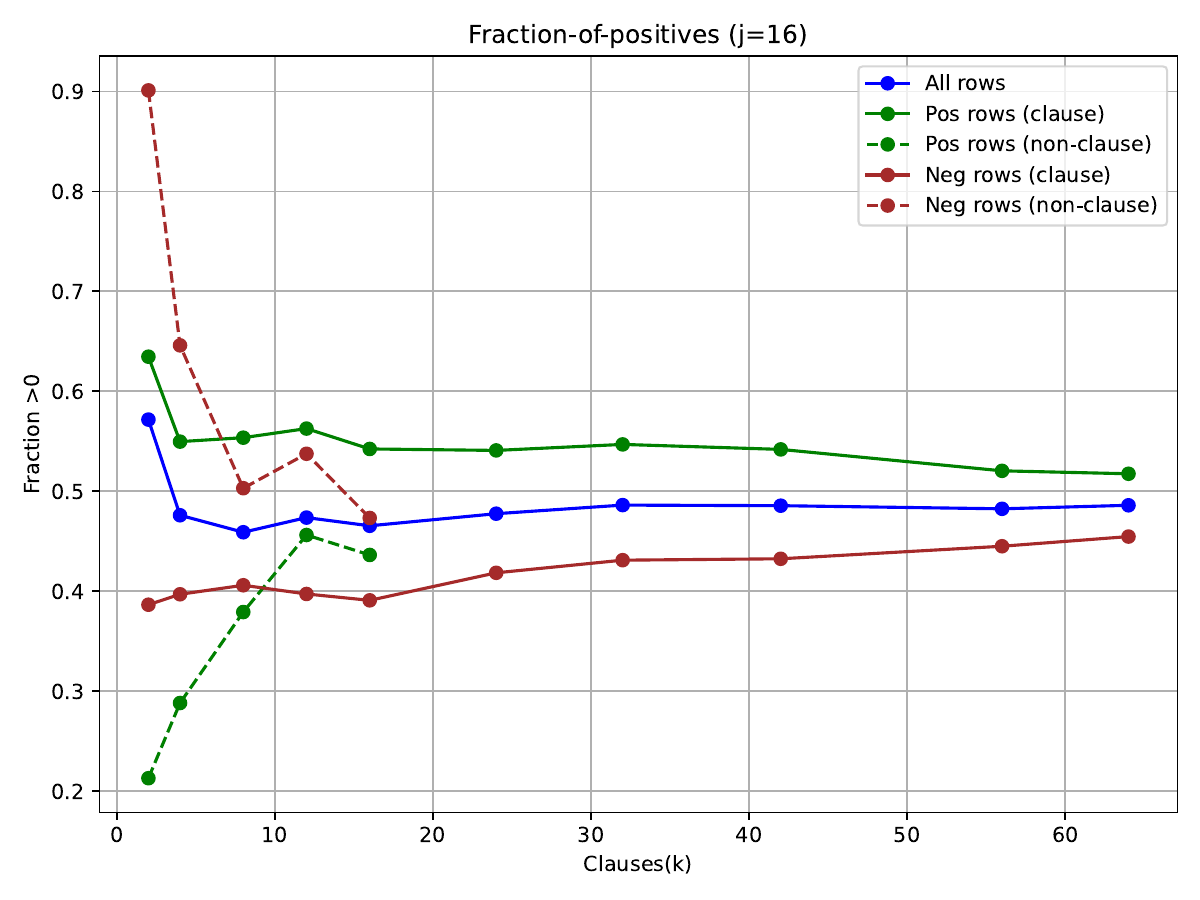}
  \end{subfigure}
  \begin{subfigure}[b]{0.32\textwidth}
    \centering
\includegraphics[width=\textwidth]{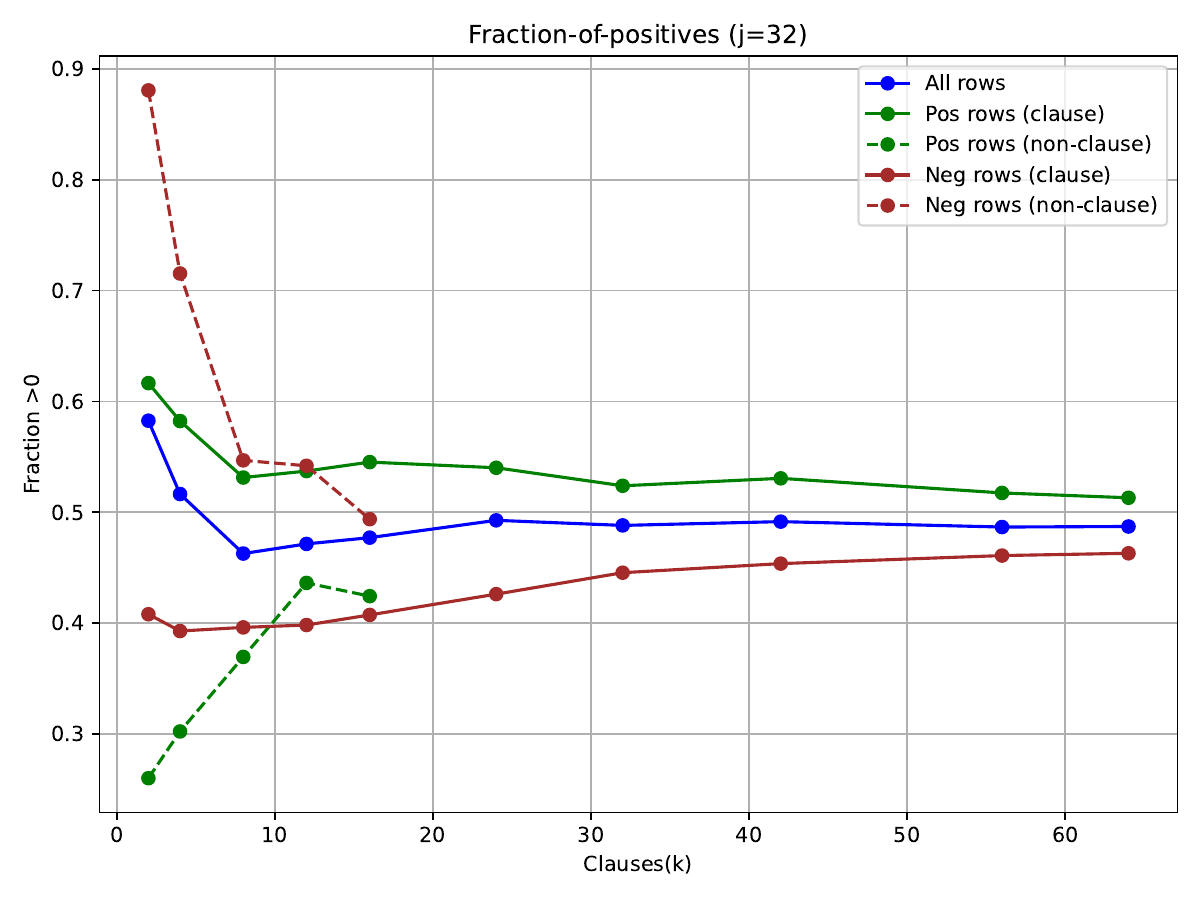}
  \end{subfigure}
  \begin{subfigure}[b]{0.32\textwidth}
    \centering
\includegraphics[width=\textwidth]{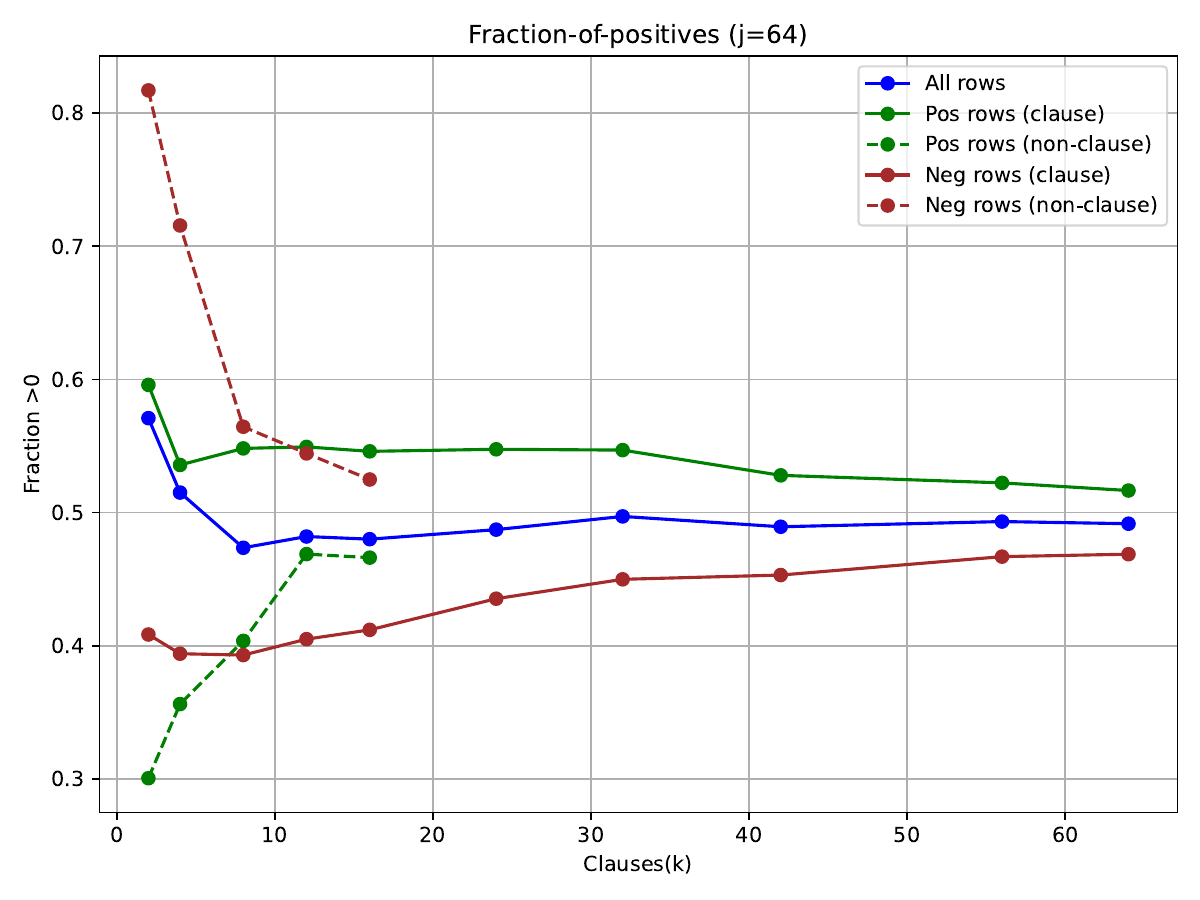}
  \end{subfigure}
  \caption{Bias of Layer 1 matrix weights, overall and by row and variable type}
  \label{matrix_bias}
\end{figure}

We next turn to the three versions of Figure \ref{matrix_bias}.  This is looking at the fraction of entries of the hidden layer weight matrix that are positive and negative.  We examine the overall number, as well as the fraction in positive rows that are in inputs used in at least one clause, positive rows that are in inputs not used by any clauses, and similarly for negative rows.  We see that overall, very close to half of the matrix entries are positive, and that there is a small positive bias for the positive rows and a small negative bias for the negative rows.  This small bias decreases as the number of clauses increases.  We also see that the non-clause variables have the opposite bias of the clause variables, although closer examination (not depicted in the graph) reveals that the magnitude of these weights is much smaller than that of the clause variables, and so we do not believe this impacts how the neural network learns very much.  Both the fact that there is so little bias in the clause variables and the fact that the three graphs are so similar (with no discernable difference where training error occurs) leads us to believe that it is useful for the network to have close to an even mix of positive and negative values, but this measure does not explain training error.

\begin{figure}[hbt!]
  \centering
  \begin{subfigure}[b]{0.45\textwidth}
    \centering
\includegraphics[width=\textwidth]{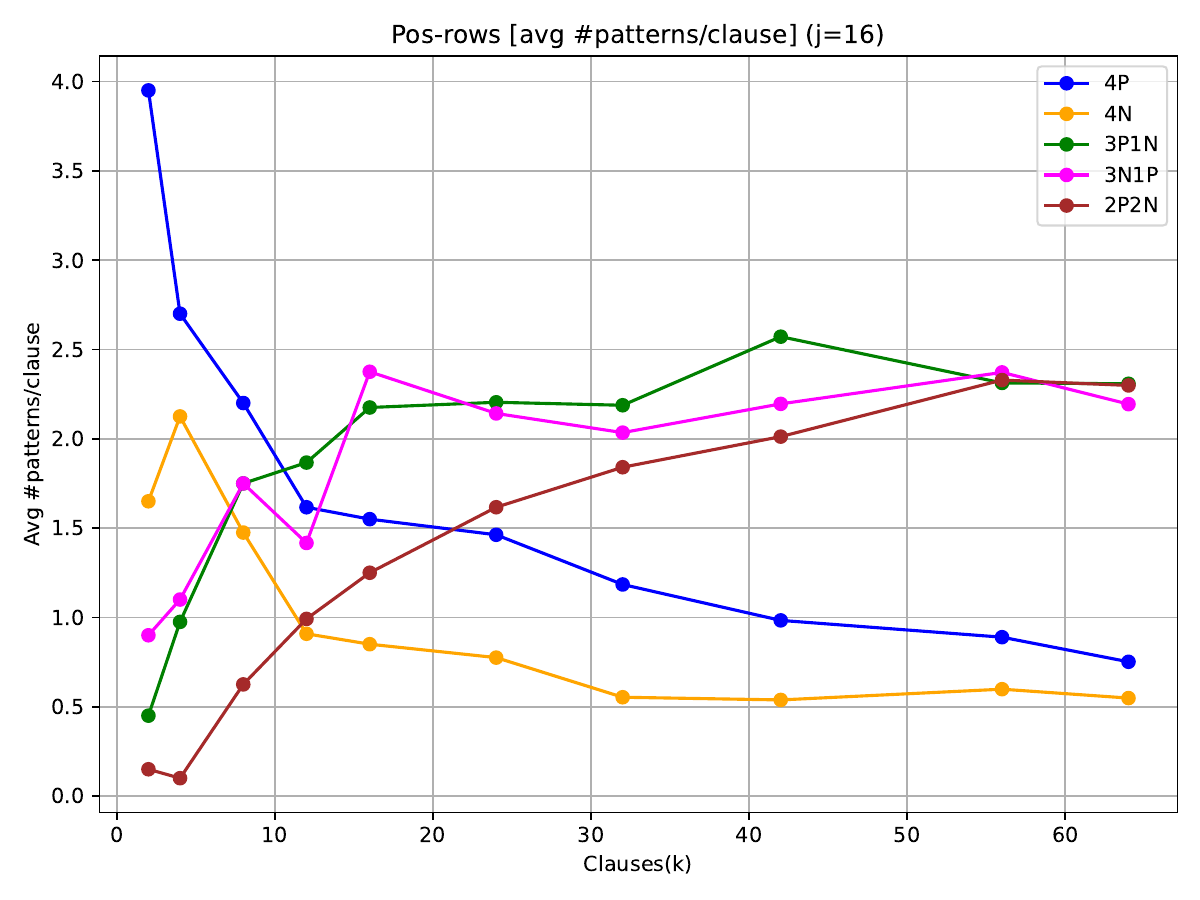}
  \end{subfigure}
  \begin{subfigure}[b]{0.45\textwidth}
    \centering
\includegraphics[width=\textwidth]{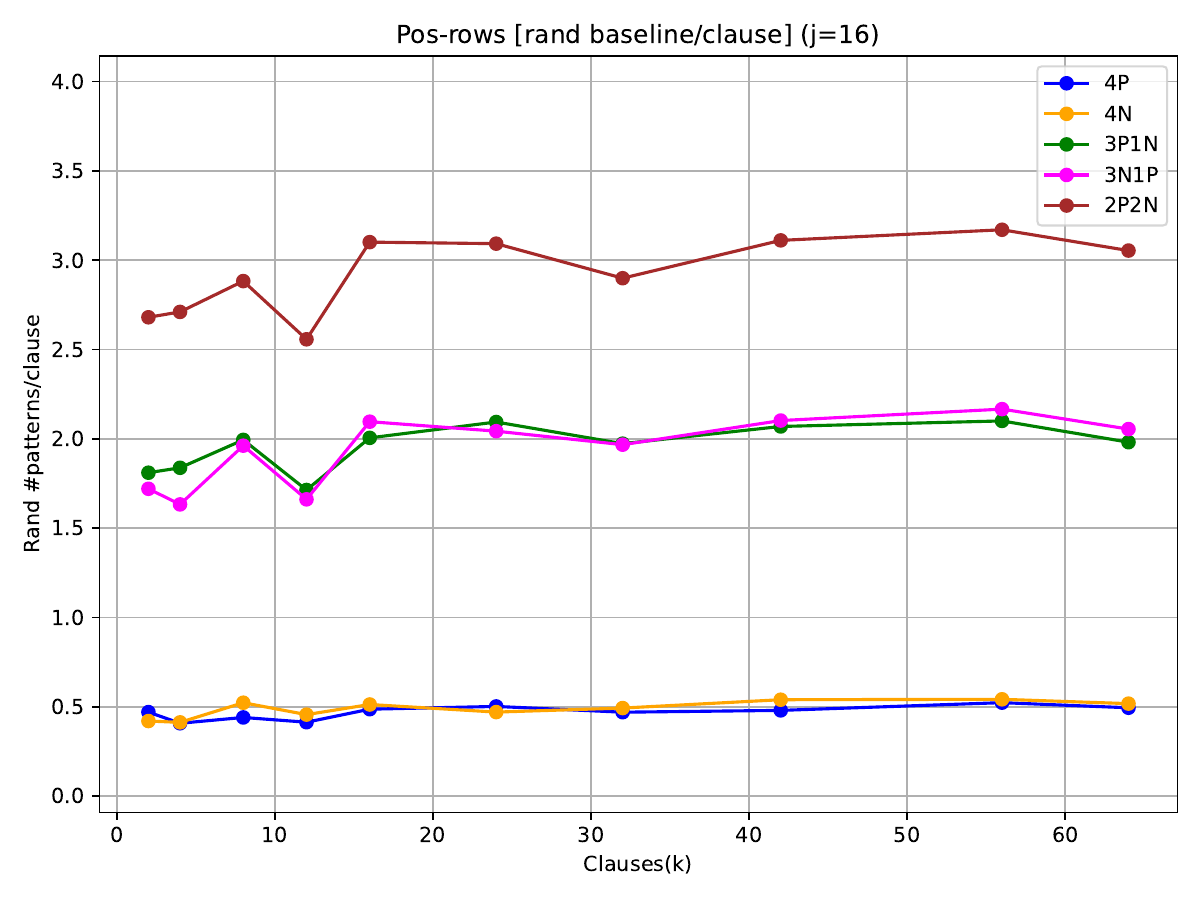}
  \end{subfigure}
  \begin{subfigure}[b]{0.45\textwidth}
    \centering
\includegraphics[width=\textwidth]{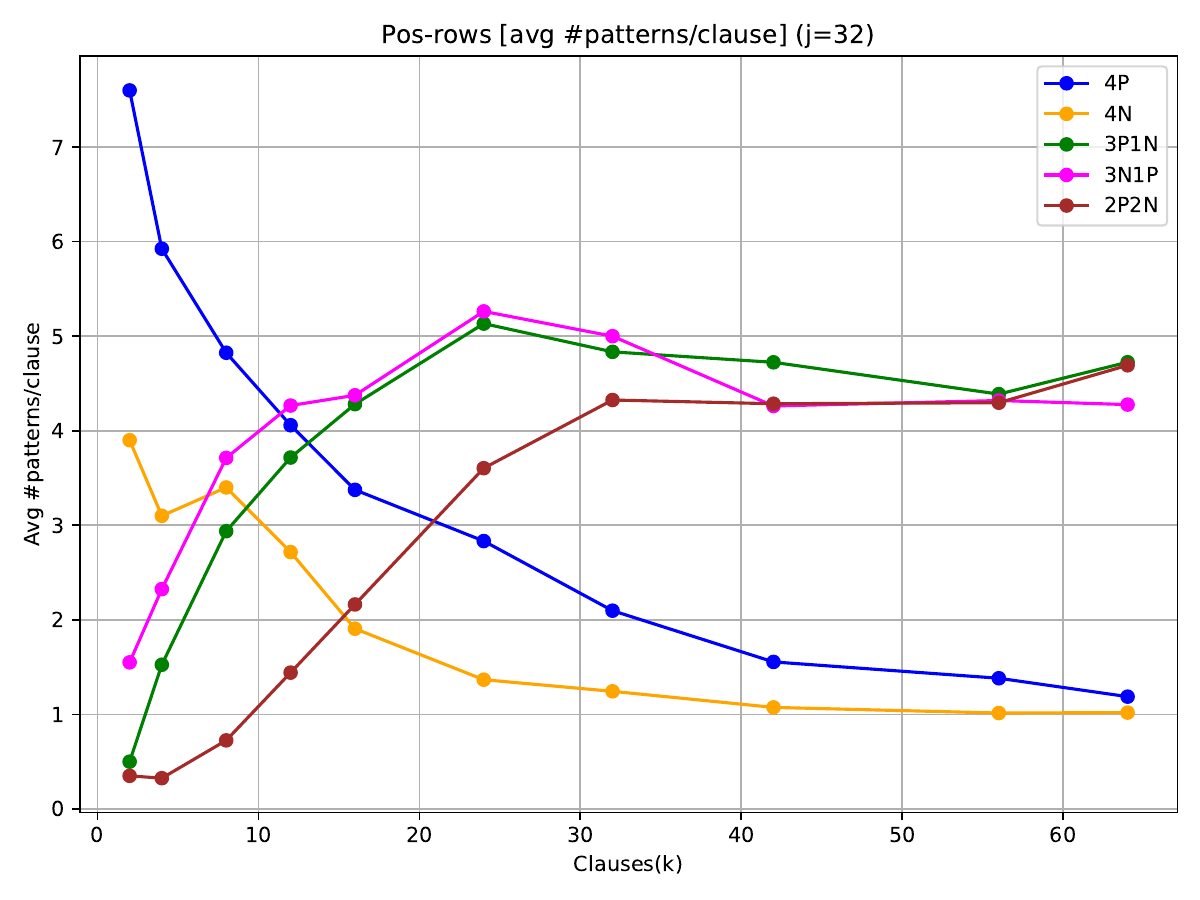}
  \end{subfigure}
  \begin{subfigure}[b]{0.45\textwidth}
    \centering
\includegraphics[width=\textwidth]{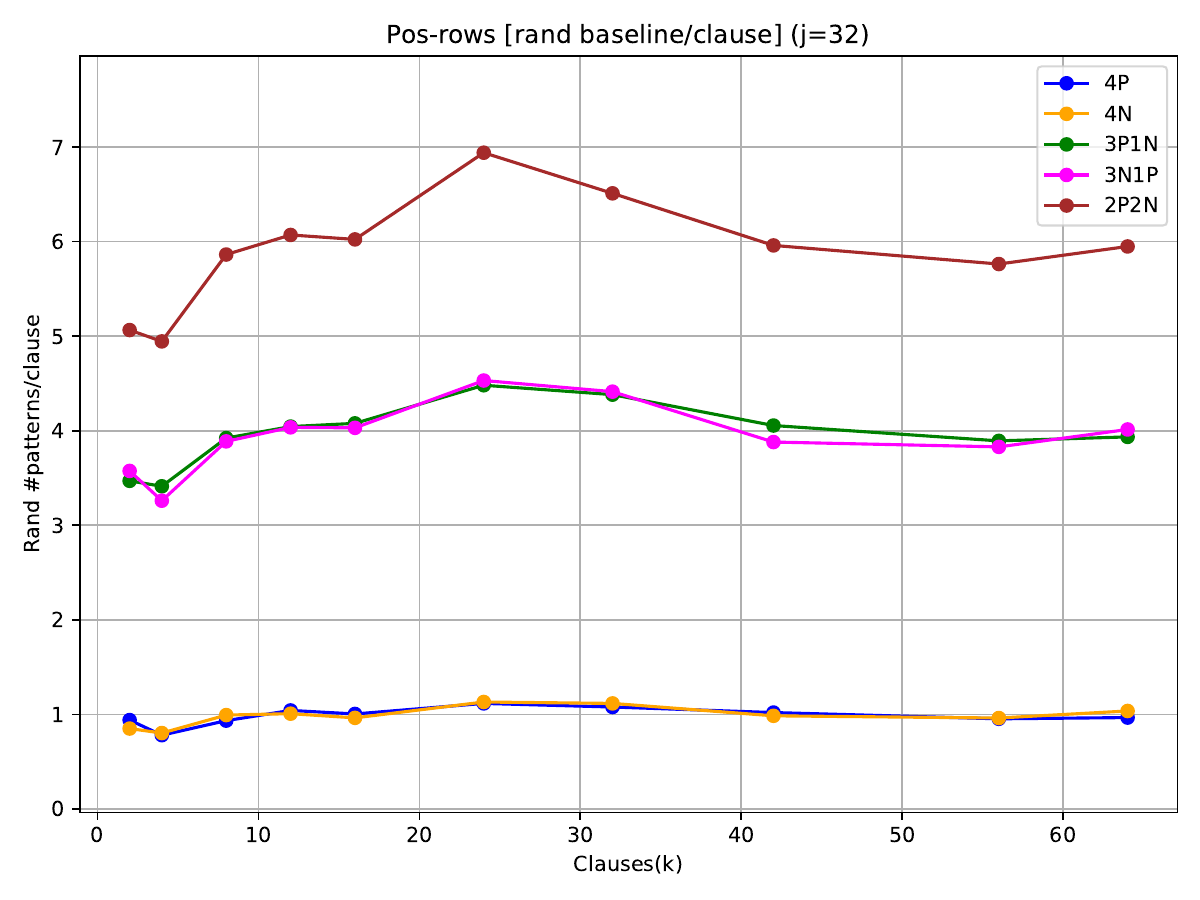}
  \end{subfigure}
  \begin{subfigure}[b]{0.45\textwidth}
    \centering
\includegraphics[width=\textwidth]{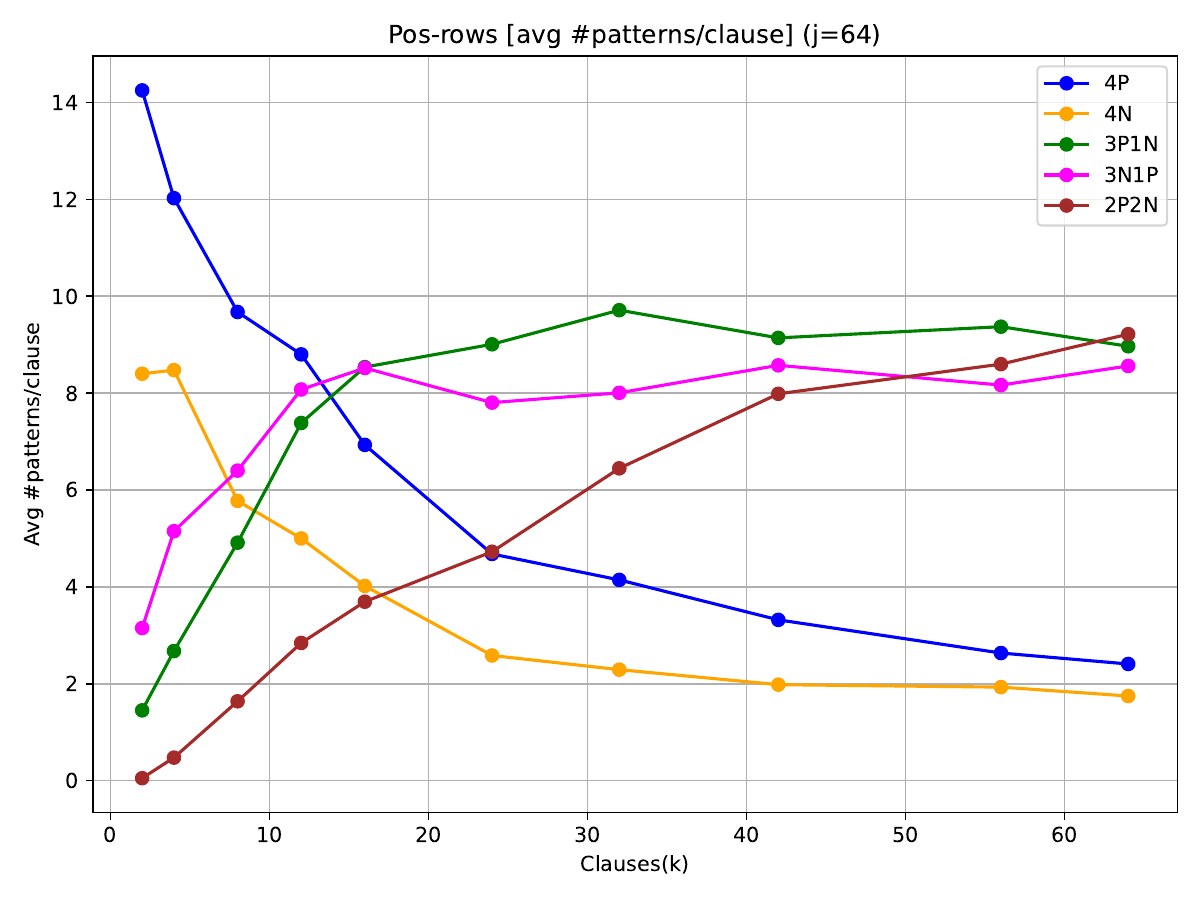}
  \end{subfigure}
  \begin{subfigure}[b]{0.45\textwidth}
    \centering
\includegraphics[width=\textwidth]{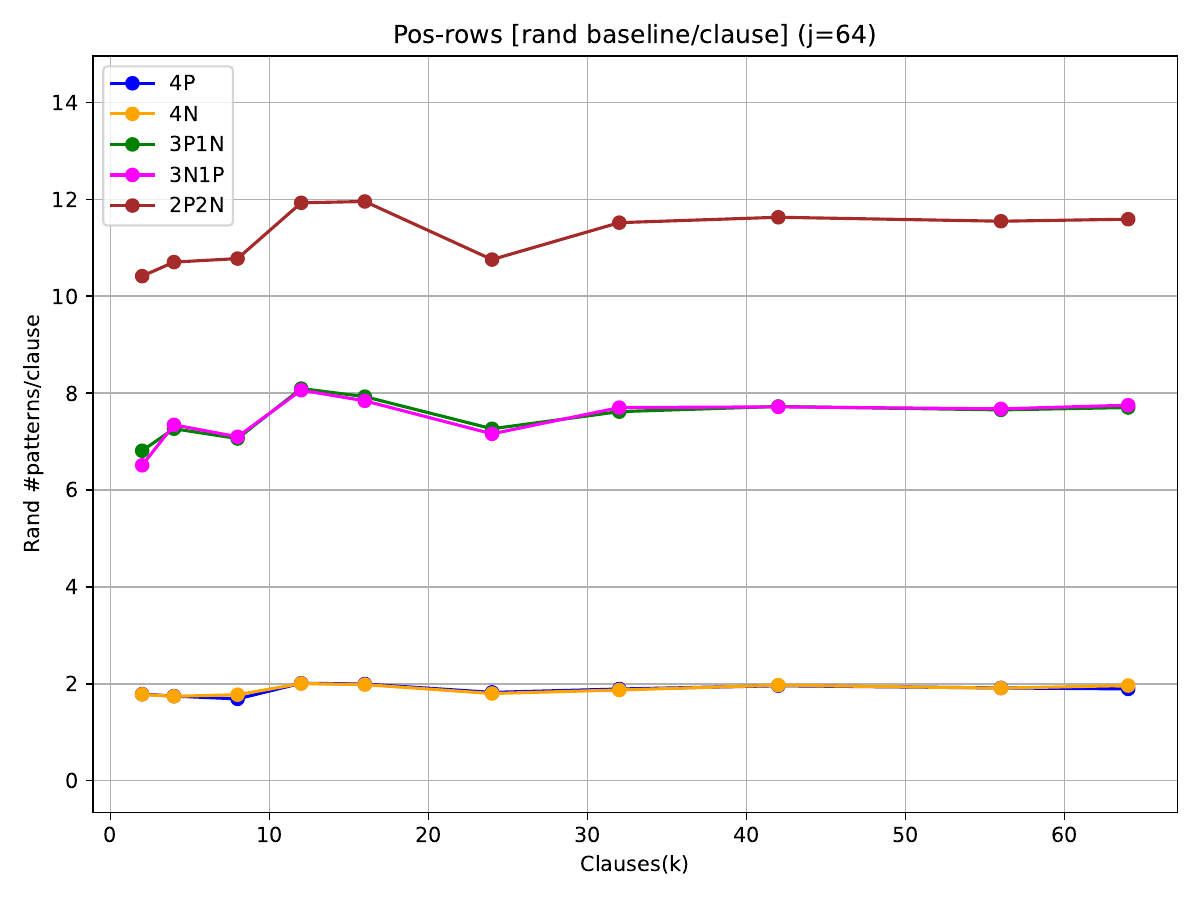}
  \end{subfigure}
  
  \caption{Prevalence of coding patterns in positive rows: trained networks versus random networks.}
  \label{positive_vs_random}
\end{figure}

\subsection{Feature Channel Coding is Real}
We next turn to Figure \ref{positive_vs_random}.  We shall see that the graphs there provide strong evidence that the system is using feature channel coding and also help us understand the limitations of what the network is and is not able to learn, and so we describe those graphs in a fair bit of detail. We study what patterns of positive and negative values in the weight matrix the clauses of the Boolean formula line up with.  We partition the patterns into five different types: 4P, 4N, 3P1N, 3N1P, and 2P2N, where the number before the P (respectively N) represents the number of positive (respectively negative) weights in the pattern.  So, for example, 3P1N represents the four variables of a clause lining up with a row of the weight matrix with 3 positive values and 1 negative value.  Each type of pattern has a line in the figure, where the vertical value of any point represents the average, over all clauses and all runs, of the number of positive witness (i.e., positive Layer 2 weight) rows of the Layer 1 weight matrix where that pattern type appears for the four variables of that clause.  We will examine negative witnesses in Figure \ref{negative_vs_random} below.

The three graphs on the left side are what we see in the 10 runs of a trained network, for the three different hidden layer sizes we study.  We see that when the number of clauses is small, the clauses exhibit a large number of 4P patterns (as depicted in the blue lines on the left - for example, we see 14 4P patterns per clause when there are 4 clauses and $j=64$ hidden neurons.)  However, this would not be possible when the number of clauses increases.  We show below that there is not enough ``room'' in the weight matrix to simultaneously (a) have all the clauses line up with a large number of 4P patterns, and (b) maintain a fraction of almost half negative weights in each row. This is exactly what we see in the line for 4P: as the number of clauses increases, the average number of occurrences of the 4P pattern per clause declines.   Note that if the network learned to increase the fraction of positive weights in a row, it would be possible to maintain a large number of 4P patterns, but in that case the network would lose its ability to differentiate between 4 positive variables all in the same clause, and 4 positive variables in different clauses.  

To further understand the values in the three graphs on the left side of Figure \ref{positive_vs_random}, we also provide the three graphs on the right side of Figure \ref{positive_vs_random}.  For these three graphs, the weights of the hidden layer matrix are chosen randomly, and this allows us to differentiate how the frequency of the patterns in the trained network deviates from random.  We expect that patterns that show up more frequently than random are important to the training, and patterns that show up less frequently are either a hindrance or not useful.  In more detail, for each number of clauses, we use the number of positive rows (neurons) we saw in Figure \ref{negative_rows}, and the bias of the weight matrix from Figure \ref{matrix_bias} to determine a bias towards positive entries in the matrix. Using that bias, we choose random entries for the weight matrix, and then we count how many times each pattern type appears as we did for the trained matrix.  We take the average over 10 samples of the random choices.

When the number of clauses is very small, we see that both 4P and 4N are much more frequent than random.  For 4N, this falls off fairly quickly, and it shows up with essentially the same frequency as random starting at around 30 clauses.  The 4P pattern also seems to be approaching random, but does not get there by 64 clauses.  This is evidence that these patterns are central to how the network has learned the Boolean function, which is exactly what the feature coding hypothesis predicts. The places where the 4P patterns show up are the codes for that clause, and they show up in multiple rows.  The 4N patterns show up to offset the 4P patterns, and to allow the network to differentiate between 4 true variables in a clause and 4 true variables in different clauses.

We also point out that with a small number of clauses, both the 3P1N and 3N1P patterns are suppressed, but as the number of clauses grow, they actually seem to occur slightly more than random.  We believe that this is due to 4P and 4N being preferable and leading to more precise results when there is room for them, but when there is not such room, the network uses 3P1N and 3N1P as an approximation to the better patterns, since more of them can fit into the network.  We also note that the point where the frequency of 3P1N and 3N1P crosses the random values coincides with a decrease on the reliance on a negative layer one bias, as seen in Figure \ref{negative_rows}.  We believe this is due to the single negative weight in 3P1N providing an effective enough replacement for that bias, which would explain why, as the number of clauses grows, the network no longer requires a negative bias on all positive rows.  2P2N is always suppressed, although it becomes less so as the clauses increase.  This is not surprising, since 2P2N is not helpful for feature channel coding of positive rows.  However, as the number of clauses increases, through random chance some of the clauses will increasingly see them showing up.

Perhaps most importantly, we also want to make clear that with all three network sizes, the error starts to accumulate right around the time that the number of 4P patterns per clause decreases below a threshold between 1.5 and 2.  This is also consistent with the feature coding hypothesis, since once we pass that point, these clauses are not successfully coded for, and the network no longer is able to use feature coding. We believe that this is really the crux of why the error starts to accumulate.  Once the network is no longer able to "pack in" enough of the 4P patterns, it is no longer able to properly learn the function.  We will next provide an analysis of why this happens at the point it does.  We also point out that
this seems to indicate that once feature coding is no longer feasible, there is no "Plan B" - the network does not have another fallback plan to make up for the lack of feature codes.

\subsection{Theoretical Limits of the 4P Coding Pattern}
\label{packing_theory}

We next dive deeper into the limit on how many 4P patterns can be packed into the weight matrix.  We looked at a similar coding capacity problem in \cite{adler2024}, although both the problem setup and the techniques used for its analysis were significantly different than what we look at here.  In \cite{adler2024} we used Kolmogorov complexity to show that $n$ neurons can compute no more than $O(n^2 / \log n)$ features in superposition, where the features were pairwise AND functions.  Here, we do not assume superposition, nor do we really achieve it, since the error starts to accumulate right around the time the number of features passes the number of neurons.  Also, we are concerned here with precise numbers, as opposed to asymptotic growth rates, and the setup is a bit different: here we are keeping the number of input variables fixed.  Finally, here we analyze the limits of a specific encoding technique (albeit, the only one the network seems to be able to find), as opposed to the information theoretic bounds proven in \cite{adler2024}.

As pointed out above, the saturation of the 4P patterns seems to be a real limitation of the network. We first point out that the positive rows have roughly half positive (coding) values, and half of them are negative. This is necessary to group the inputs contained in a feature together - if all inputs were positive it would not be possible to distinguish 4 inputs from the same AND clause being true from 4 inputs from different AND clauses being true.  Thus, we assume here that if we have $\ell$ input variables, each row will have roughly $\ell / 2$ positive values that can go towards 4P patterns.  Let $\rho$ be the fraction of positive rows.  If there are $j$ hidden neurons, then the total number of positive entries in positive rows will be roughly $\frac{j\ell\rho}{2}$.  

Each 4P pattern requires 4 of those positive entries.  There may be some overlap between the 4 positive entries of different 4P patterns, but since the clauses are chosen randomly, that overlap will be small, and we shall ignore that effect here (but we note that for clauses of size 2 that overlap is much more significant, and in fact we see that networks trained on such smaller clauses can handle a much larger number of clauses).  As a result, the average number of 4P patterns per clause, when there are $k$ clauses, will be $\frac{j\ell\rho}{8k}$.  Since $\rho$ is approximately $1/2$, this will be roughly $\frac{j\ell}{16k}$  This explains the saturation point we see in the above graphs.  For example, when $j=k=\ell=32$, according to this formula, we would not expect to see more than 2 occurrences of 4P per clause.  Comparing this prediction to Figure \ref{positive_vs_random}, we see that our prediction is accurate: the value of the blue curve in the $j=32$ graph of the right hand column, at the $32$ clause point is almost exactly 2.  In other words, the network in practice was able to achieve this maximum packing of 4P patterns, but not better.

If we increase $k$ to 64 and hold the other variables fixed, our formula says there is not room for more than an average of one 4P per clause; again Figure \ref{positive_vs_random} shows us the network is able to achieve that but not much higher (we suspect the slight overperformance is the result of increasing overlap of clauses).  Conversely, when we decrease $k$, the network does not quite keep up with the theoretical maximum.  But in those cases, the network does not have an incentive to keep up with that maximum: there is plenty of room for coding and so it does not have to saturate the network.  We see a similar agreement with this packing limit for the $j=16$ and $j=64$ curves for 4P in Figure \ref{positive_vs_random}.  In all cases, when the network is no longer able to achieve an average of approximately 2 coding rows per clause, it starts to accumulate error.

\begin{figure}[hbt!]
  \centering
  \begin{subfigure}[b]{0.45\textwidth}
    \centering
\includegraphics[width=\textwidth]{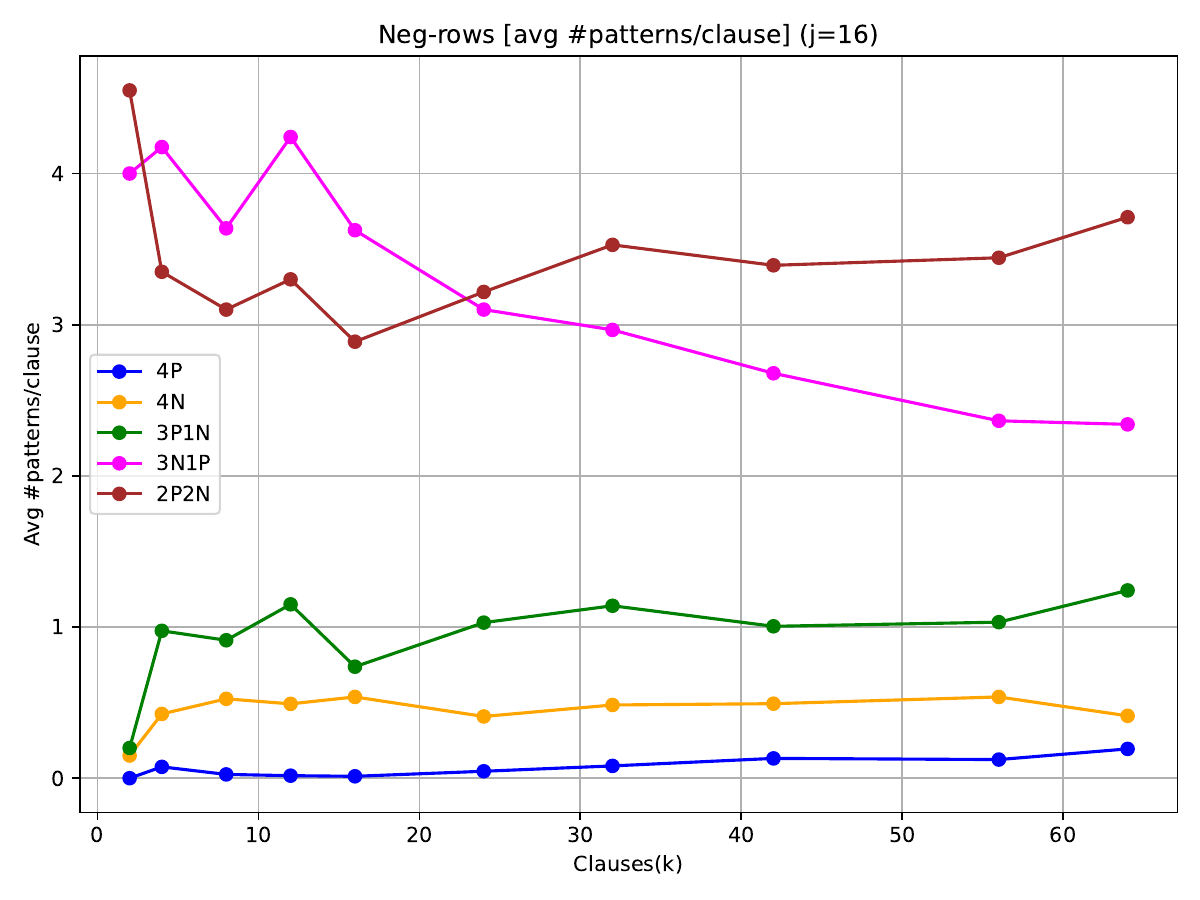}
  \end{subfigure}
  \begin{subfigure}[b]{0.45\textwidth}
    \centering
\includegraphics[width=\textwidth]{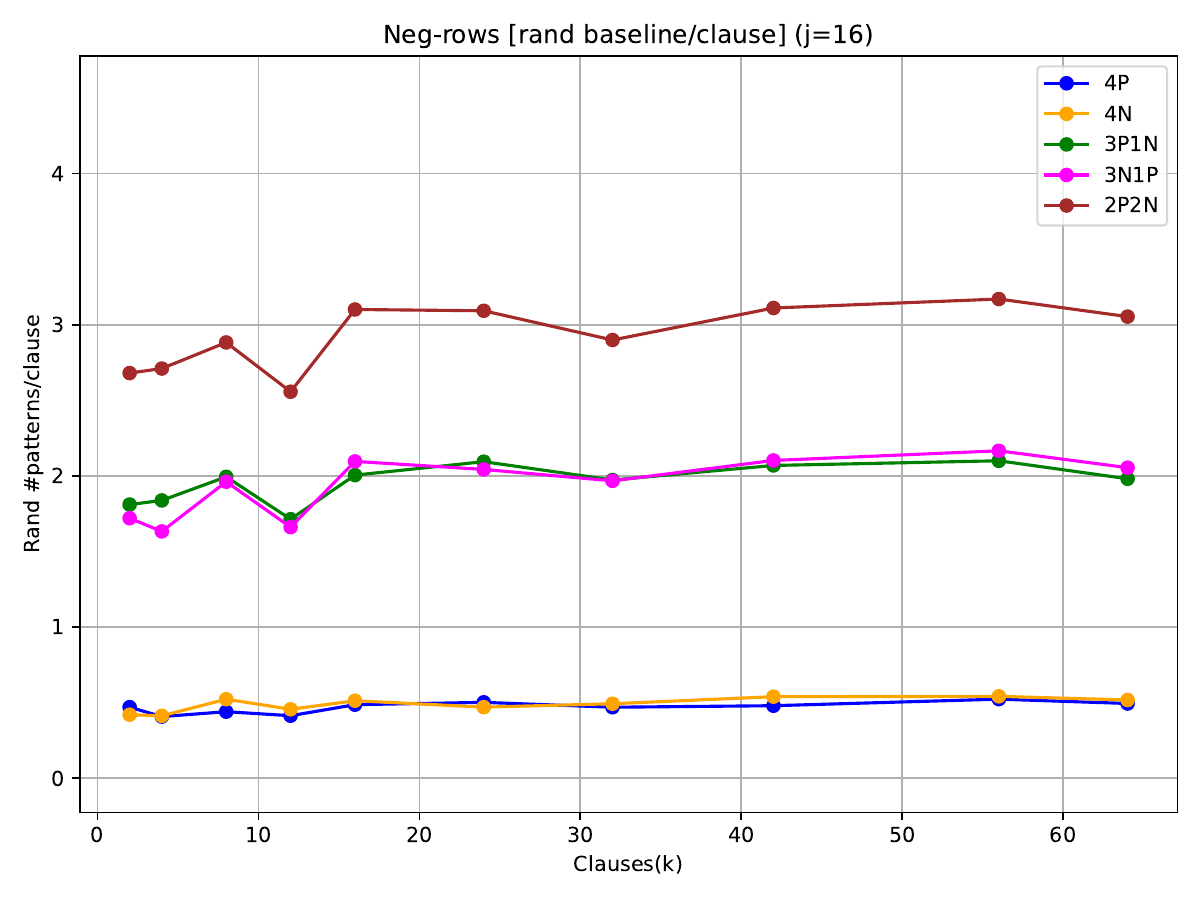}
  \end{subfigure}
  \begin{subfigure}[b]{0.45\textwidth}
    \centering
\includegraphics[width=\textwidth]{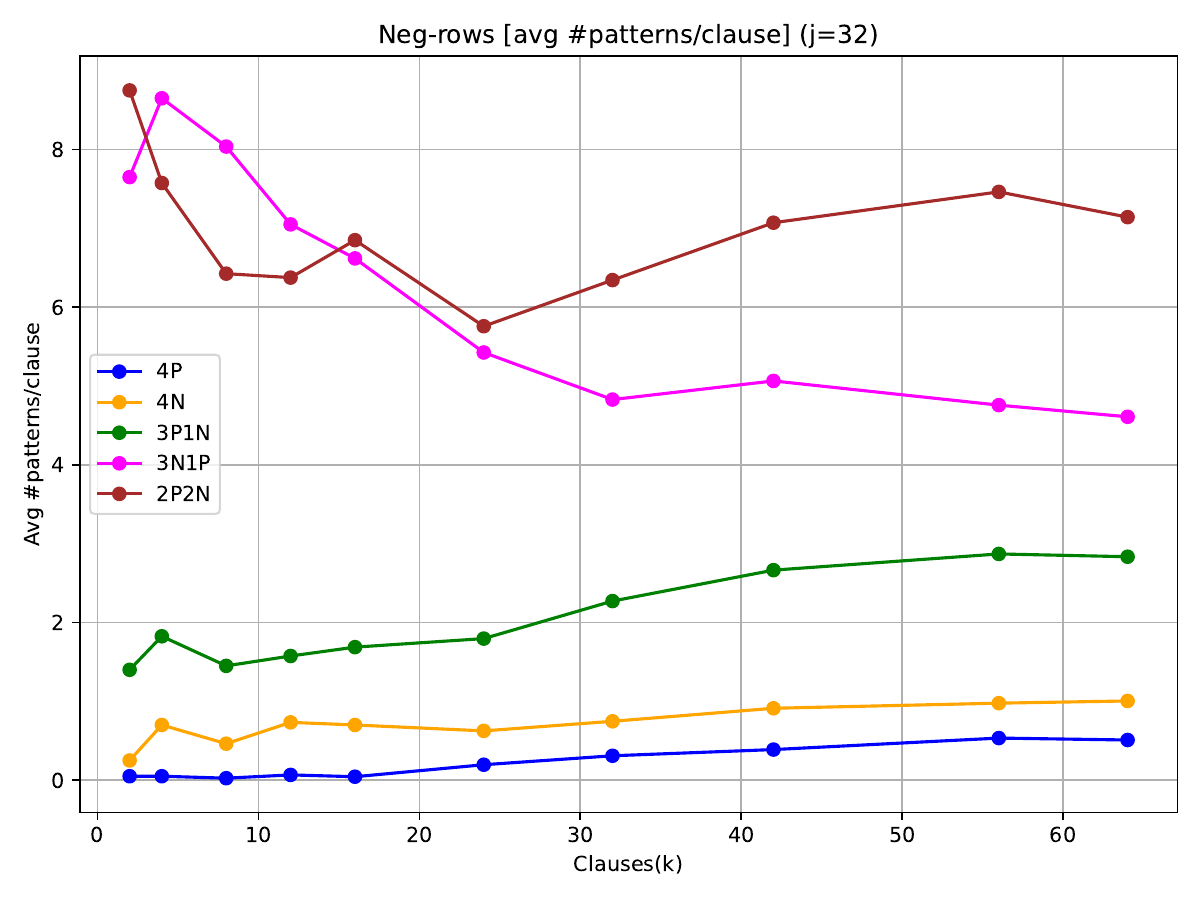}
  \end{subfigure}
  \begin{subfigure}[b]{0.45\textwidth}
    \centering
\includegraphics[width=\textwidth]{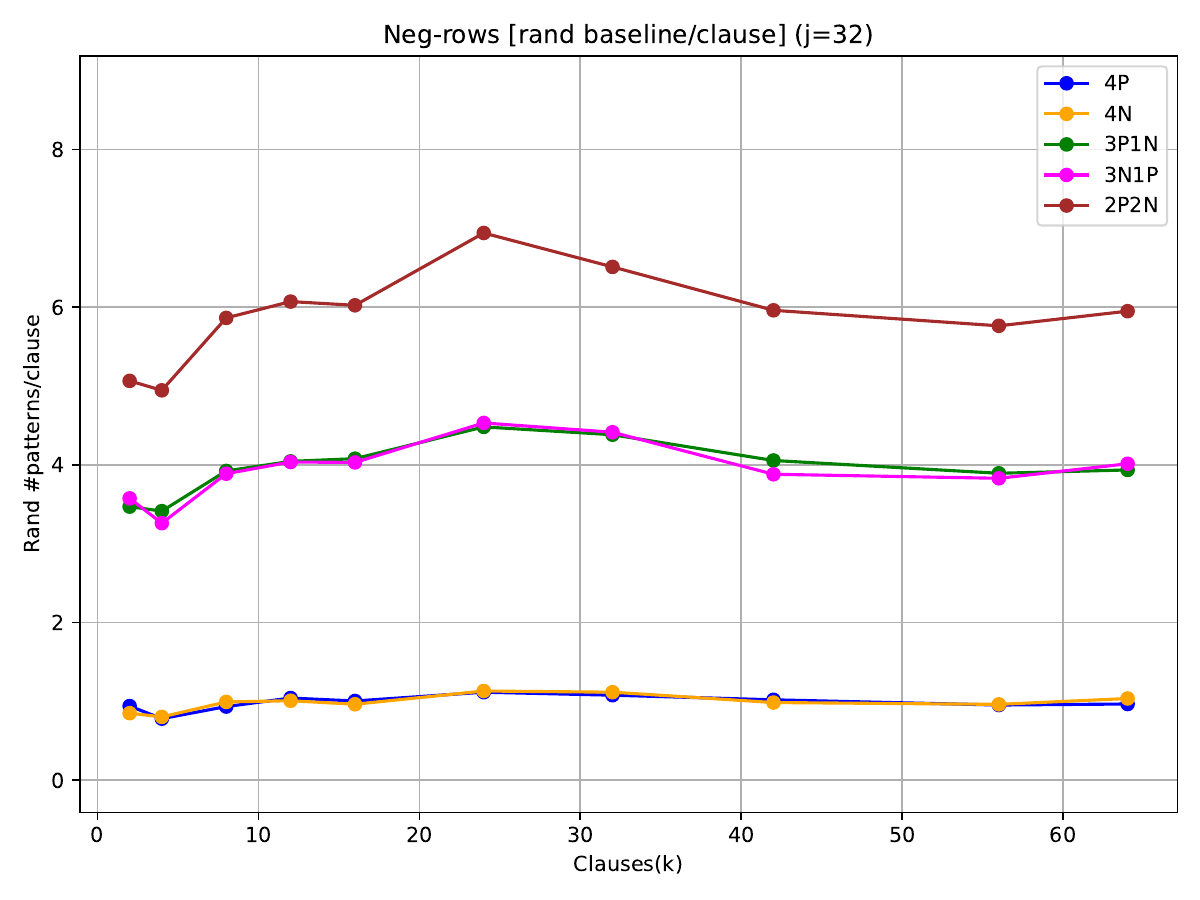}
  \end{subfigure}
  \begin{subfigure}[b]{0.45\textwidth}
    \centering
\includegraphics[width=\textwidth]{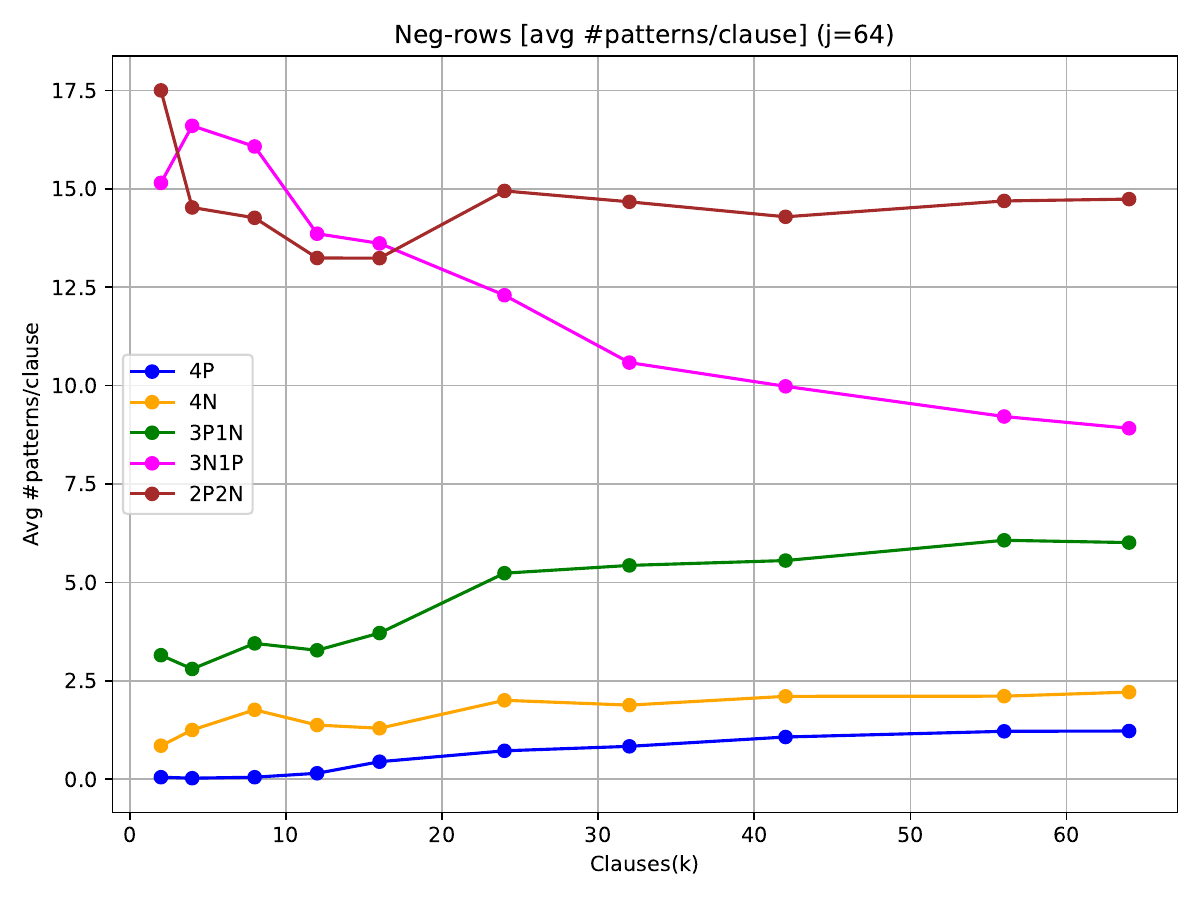}
  \end{subfigure}
  \begin{subfigure}[b]{0.45\textwidth}
    \centering
\includegraphics[width=\textwidth]{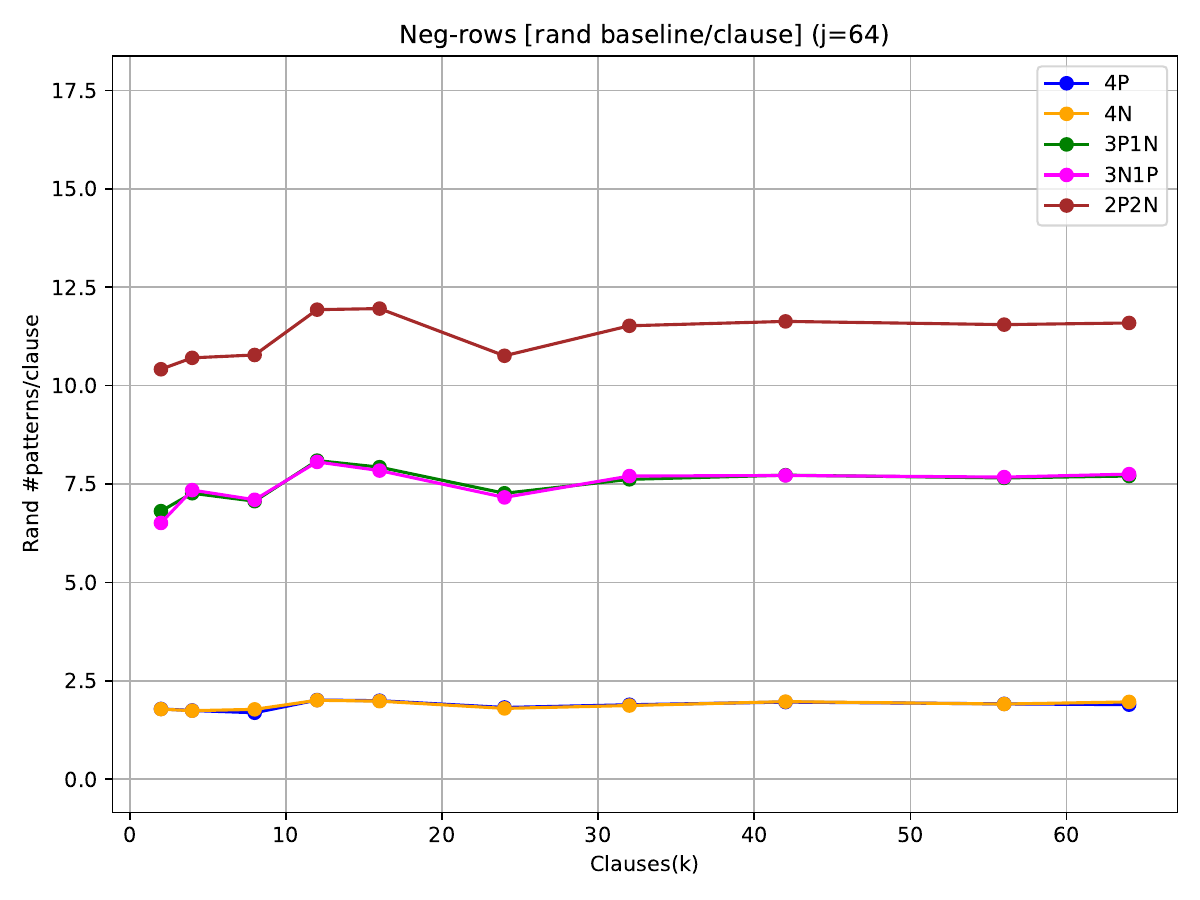}
  \end{subfigure}
  
  \caption{Prevalence of coding patterns in negative rows: trained networks versus random networks.}
  \label{negative_vs_random}
\end{figure}

\subsection{Codes as Negative Witnesses}

We next turn our attention to the negative rows of the weight matrix, as depicted in Figure \ref{negative_vs_random}. These graphs are analogous to those in Figure \ref{positive_vs_random}, but focus on patterns emerging in negative rows, again using a random matrix baseline configured with the corresponding row counts and biases. For a small number of clauses, the 3N1P pattern exhibits the strongest signal in these negative rows. However, as the number of clauses increases, the dominant pattern transitions to 2P2N. Both patterns appear designed to detect instances where two or three positive variables, but not all four, are present in a clause. This detection generates a positive post-activation signal at Layer 1, which is then inverted to a negative signal by the negative Layer 2 weight. The 3N1P pattern achieves this using a positive Layer 1 bias, roughly equal negative Layer 1 weights, and ensuring the sum of the bias and the positive weight approximately offsets the sum of the three negative weights (e.g., weights: -1, -1, -1, +2; bias: +1). In this configuration, if the variable corresponding to the positive pattern column is 1, all three variables corresponding to negative columns must also be 1 to suppress a positive post-activation value. However, if any of the three negative weight variables are 0 (indicating the clause condition is not met), a positive Layer 1 activation results, leading to a negative final signal via Layer 2. As the clause count increases, there is no longer sufficient room to fully utilize the 3N1P pattern, and the mechanism shifts towards the 2P2N pattern.  This coincides with a reduced reliance on positive bias in the negative rows. Additionally, the 3P1N, 4P, and 4N patterns are consistently suppressed in negative rows, indicating they are not effective for feature channel coding to be performed by these rows.

\subsection{Code Interference}

\begin{figure}[hbt!]
  \centering
  \begin{subfigure}[b]{0.45\textwidth}
    \centering
\includegraphics[width=\textwidth]{images/plot_pdfs/Combined_Train_Error.pdf}
  \end{subfigure}
  \begin{subfigure}[b]{0.45\textwidth}
    \centering
\includegraphics[width=\textwidth]{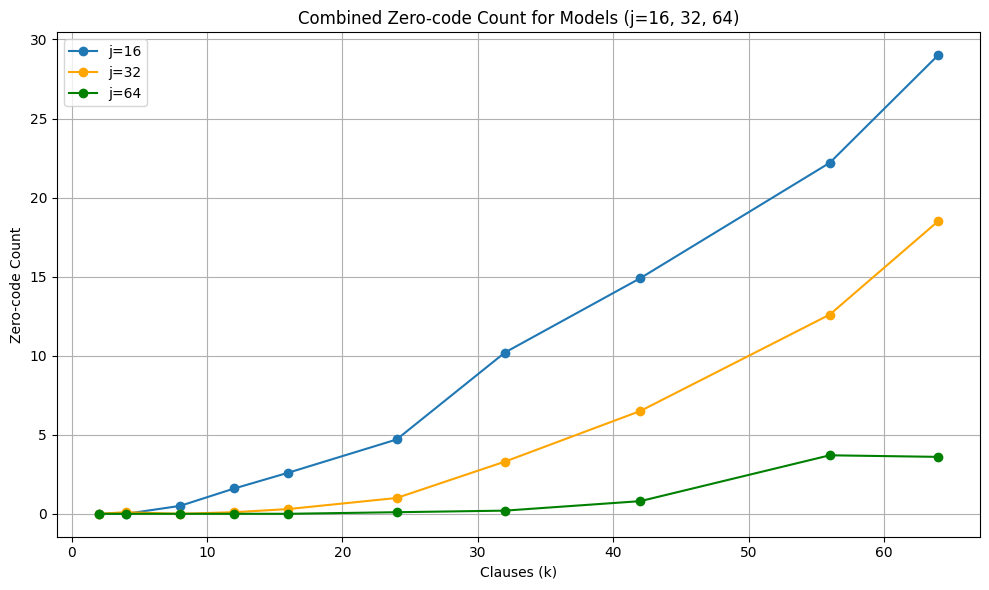}
  \end{subfigure}
  \caption{Scaling of training error and number of clauses without codes}
\label{zero_clauses}
\end{figure}

As described above, the number of 4P coding rows a clause coincides with seems to have a significant impact on the network's ability to learn the Boolean function.  We take this a step further in Figure \ref{zero_clauses}, where we depict the number of clauses that do not coincide with any 4P rows.  This is depicted on the right side - the left side is a duplicate of Figure \ref{training_error}, shown again here for convenience.  We see here that there is, in fact, a close alignment between training error and the number of clauses with no 4P coding rows.  However, it does seem like the network can tolerate a small amount of clauses without 4P coding rows.

Next, we examine the overlap between the codes that the network is using.  As shown in \cite{adler2024}, this is a central aspect of coding.  Codes usually overlap and always will if there is sufficient saturation of the network by the number of features being coded.  However, as long as the overlap between pairs of codes is not too large, this is fine.  One of the main advantages of using codes is that spreading out a signal across multiple neurons creates tolerance to noise that appears on some of those neurons.  This is a core tenant of the feature channel coding hypothesis.

\begin{figure}[hbt!]
  \centering
  \begin{subfigure}[b]{0.32\textwidth}
    \centering
\includegraphics[width=\textwidth]{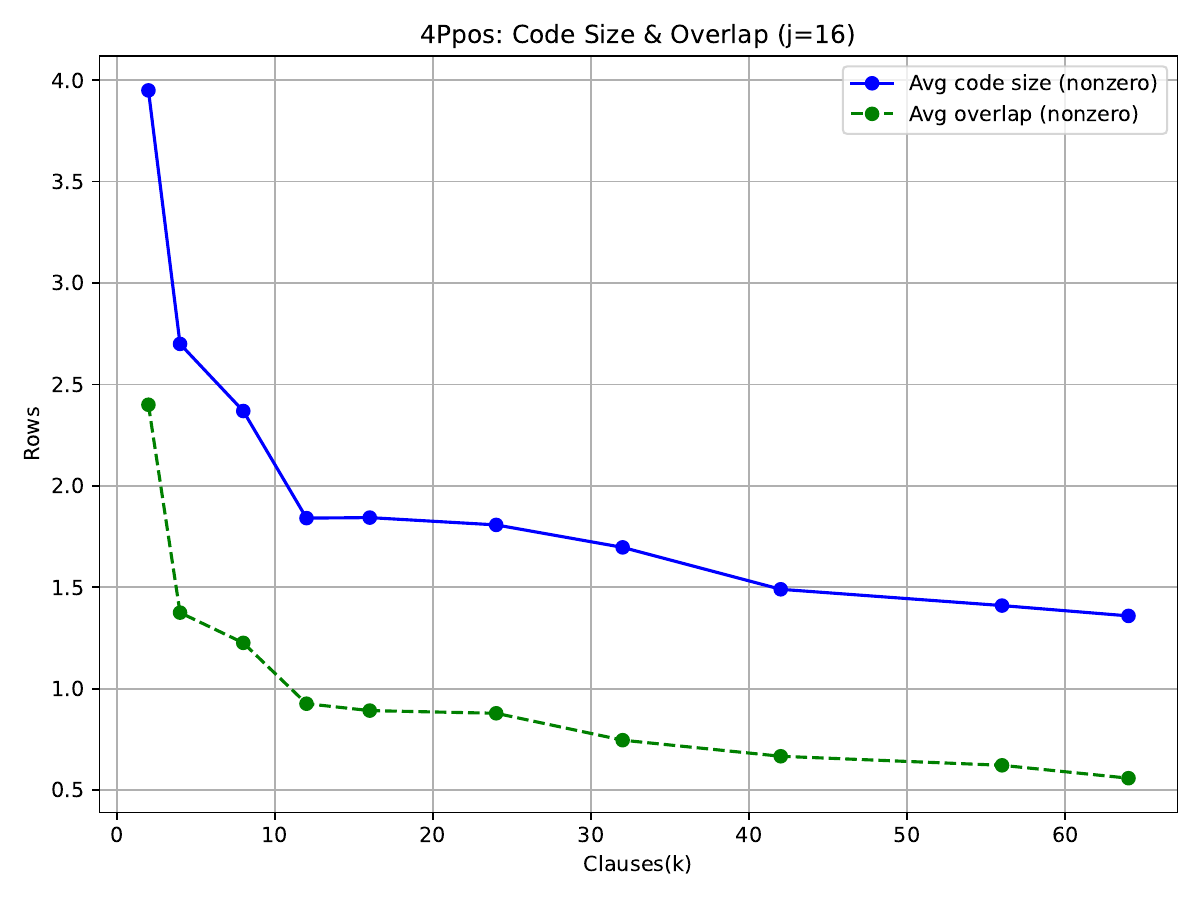}
  \end{subfigure}
  \begin{subfigure}[b]{0.32\textwidth}
    \centering
\includegraphics[width=\textwidth]{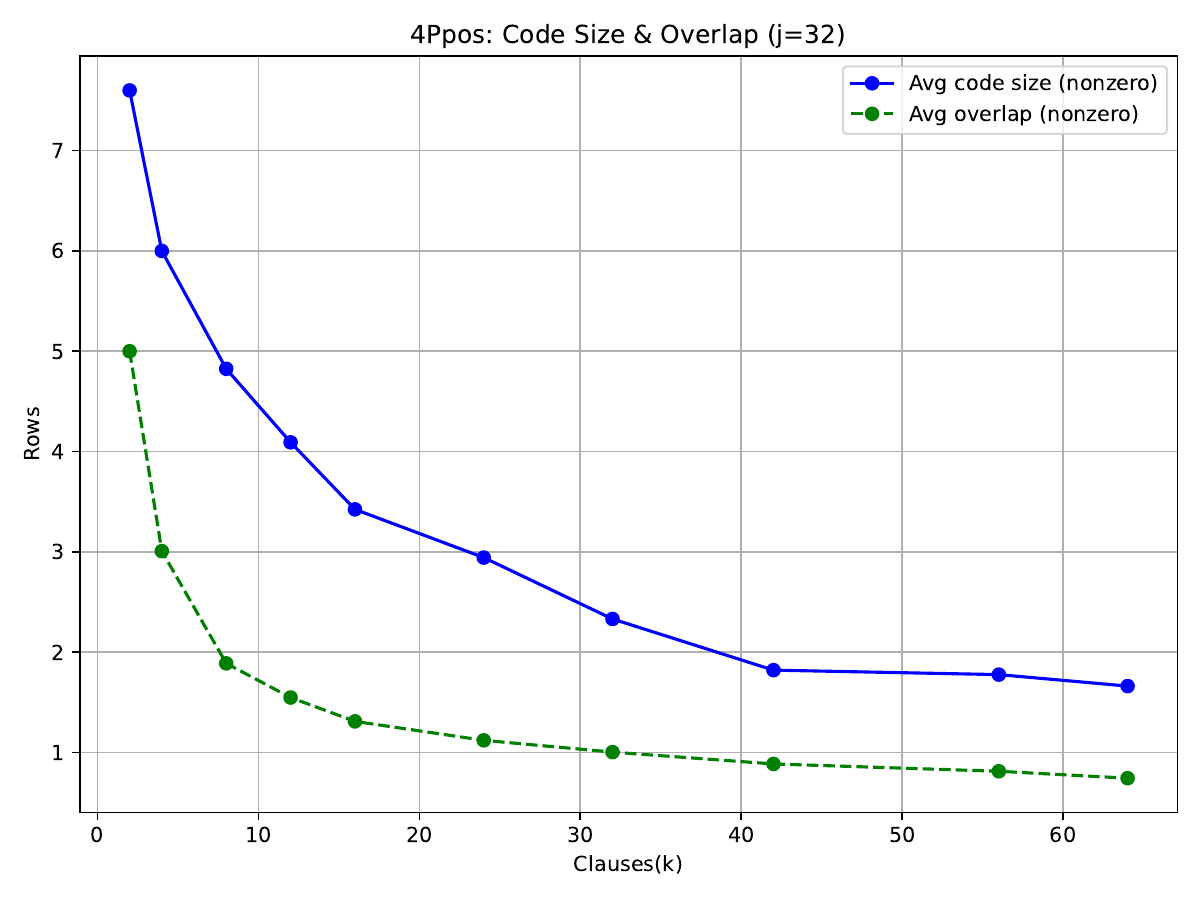}
  \end{subfigure}
  \begin{subfigure}[b]{0.32\textwidth}
    \centering
\includegraphics[width=\textwidth]{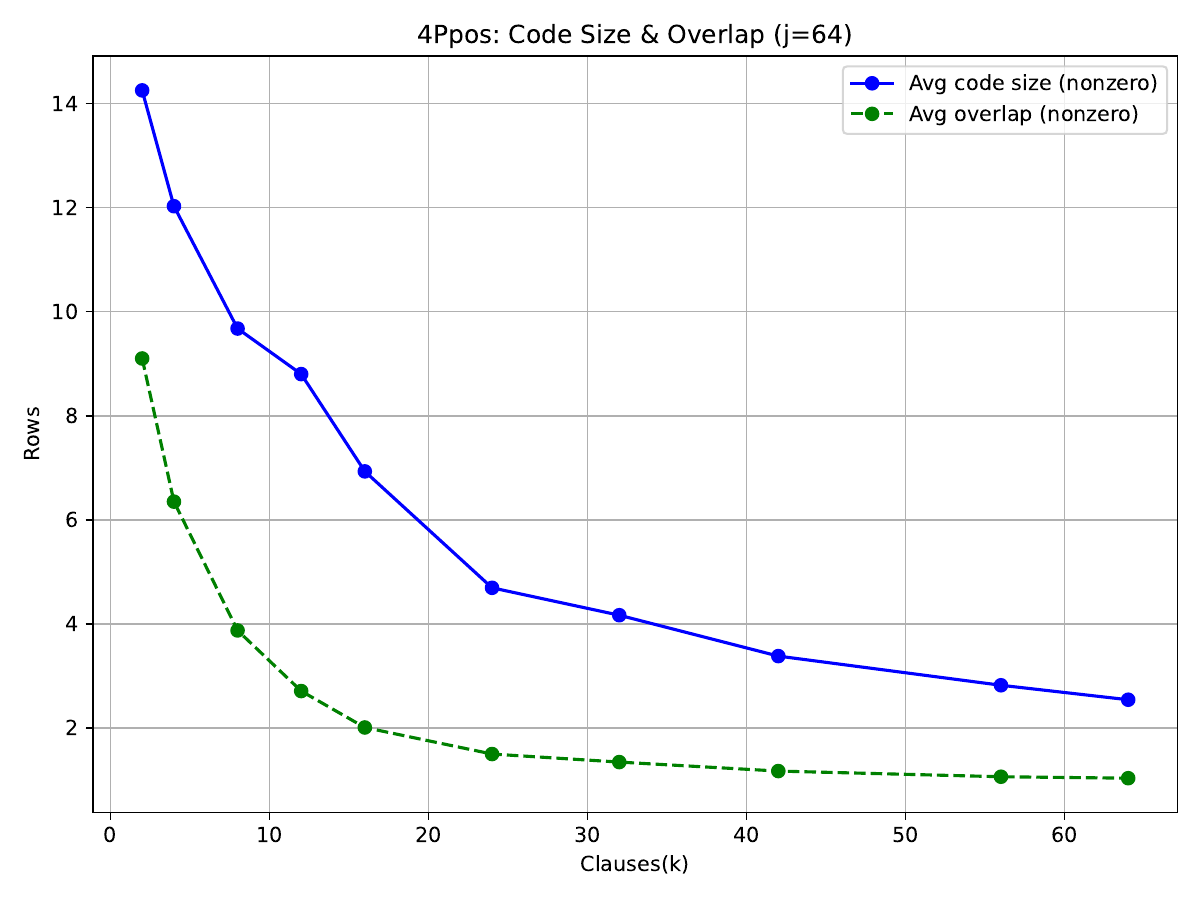}
  \end{subfigure}
  \caption{The average, over all pairs of clauses in each training run, of the overlap between the feature channel codes. Average code size is shown to give the reader a measure of the relative portion of the overlaps.}
  \label{clause_overlap}
\end{figure}

We show the code overlap measured from our experiments in Figure~\ref{clause_overlap}. We plot the average, over all pairs of clauses in each training run, of the overlap between the codes, where an overlap means that both codes use the same row as a 4P positive row.  We also show the total number of 4P coding rows. For both values, we do not include clauses without coding rows in the count, so the blue curve (code size) will differ from that of Figure \ref{positive_vs_random}, which includes those clauses.  These curves are almost identical (except for scaling) for the three hidden parameter sizes, and consistent with what we would expect from the feature channel coding hypothesis.  Also, we do not see any indication that this overlap is responsible for the network not being able to train.

\subsection{Towards Combinatorial Scaling Laws}

\begin{figure}[hbt!]
  \centering
  \begin{subfigure}[b]{1 \textwidth}
    \centering
    \hspace{1.5in}
\includegraphics[trim=2.5in 2in 1.8in 2in, clip=true, width=\textwidth]{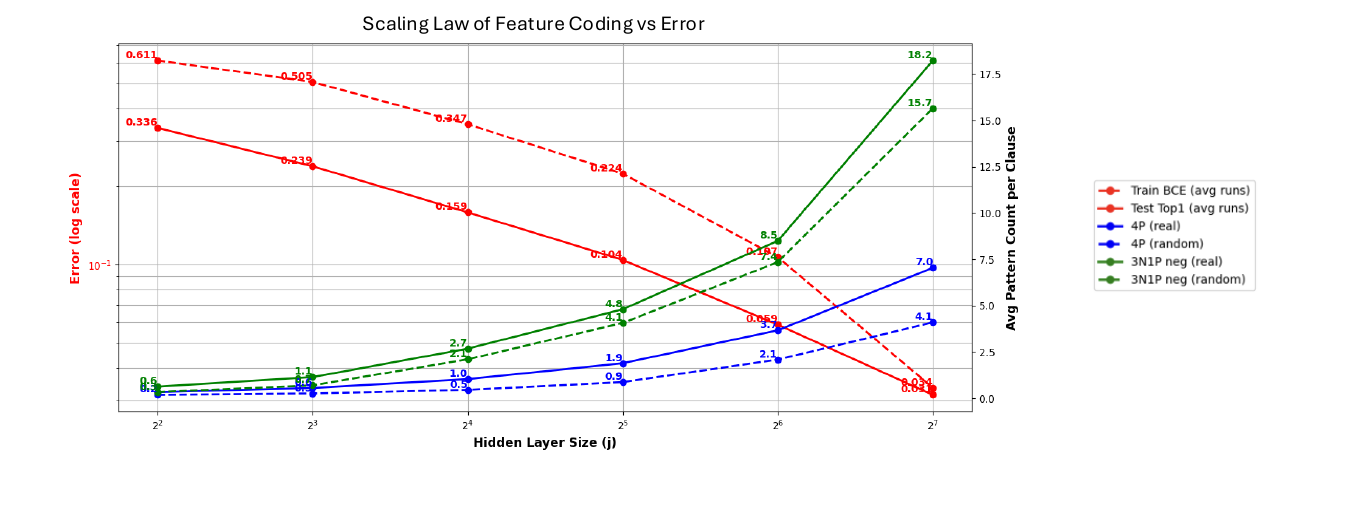}
  \end{subfigure}
  \caption{Feature channel coding combinatorially explains the neural network's scaling law.}
  \label{Scaling} 
\end{figure}

We next show that the properties of feature channel coding explain the shape of traditional scaling laws.  Unlike traditional scaling law graphs, our focus on Boolean formulas allows us to control the number of features explicitly, and we did so to construct Figure \ref{positive_vs_random}, where we studied the impact of increasing the number of features for three different sized networks.  Here, in Figure~\ref{Scaling}, we show similar data from a more traditional scaling law perspective, where we keep the number of features (clauses) fixed at $k=64$, and increase the size of the hidden layer.  We plot both the error and the frequency of 4P codes in positive rows (blue) and 3N1P codes in negative rows (green) versus the size of the hidden layer.  We see that the decrease in error is closely related to the increase in code frequency.
We believe that this relationship is due to the capacity of the network to perform feature channel coding.  As discussed in Section \ref{packing_theory}, there is a limit to how much of the 4P pattern can fit into the network, and the network roughly achieves that limit.\footnote{As before, when the hidden layer size increases and there is more space, the network starts to not need to take full advantage of that extra space.} 
This provides an easily explainable scaling law: the network's size implies a capacity to code the feature channels, with smaller networks being more limited in their coding ability, which explains the higher error rate.\footnote{Note that we trained here to half the epochs of prior examples and there is a larger error per model size.} 

We also point out that in all cases, both patterns appear more frequently than random (solid lines versus dashed lines).  The 4P positive coding is potentially more dominant as a code, hinted at by its relative ratio to the random distribution. As networks grow, the ratio of 4P to 4P random starts around 2, and decreases, but still is at almost 1.7x with $2^7$ hidden neurons, while the ratio of 3N1P to 3N1P random is at 1.16 with $2^7$ hidden neurons. One could envision building on this to use the ratios of codes in smaller models to derive scaling predictions for larger models, an approach that merits further research.

\subsection{How Feature Channel Codes Emerge}
\label{section: emerge}

\begin{figure}[h!]
  \centering
    \hspace{1in}
  \begin{subfigure}[b]{0.85\textwidth}
    \centering
\includegraphics[trim=1.4in 1.1in 2.2in 1in, clip=true, width=\textwidth]{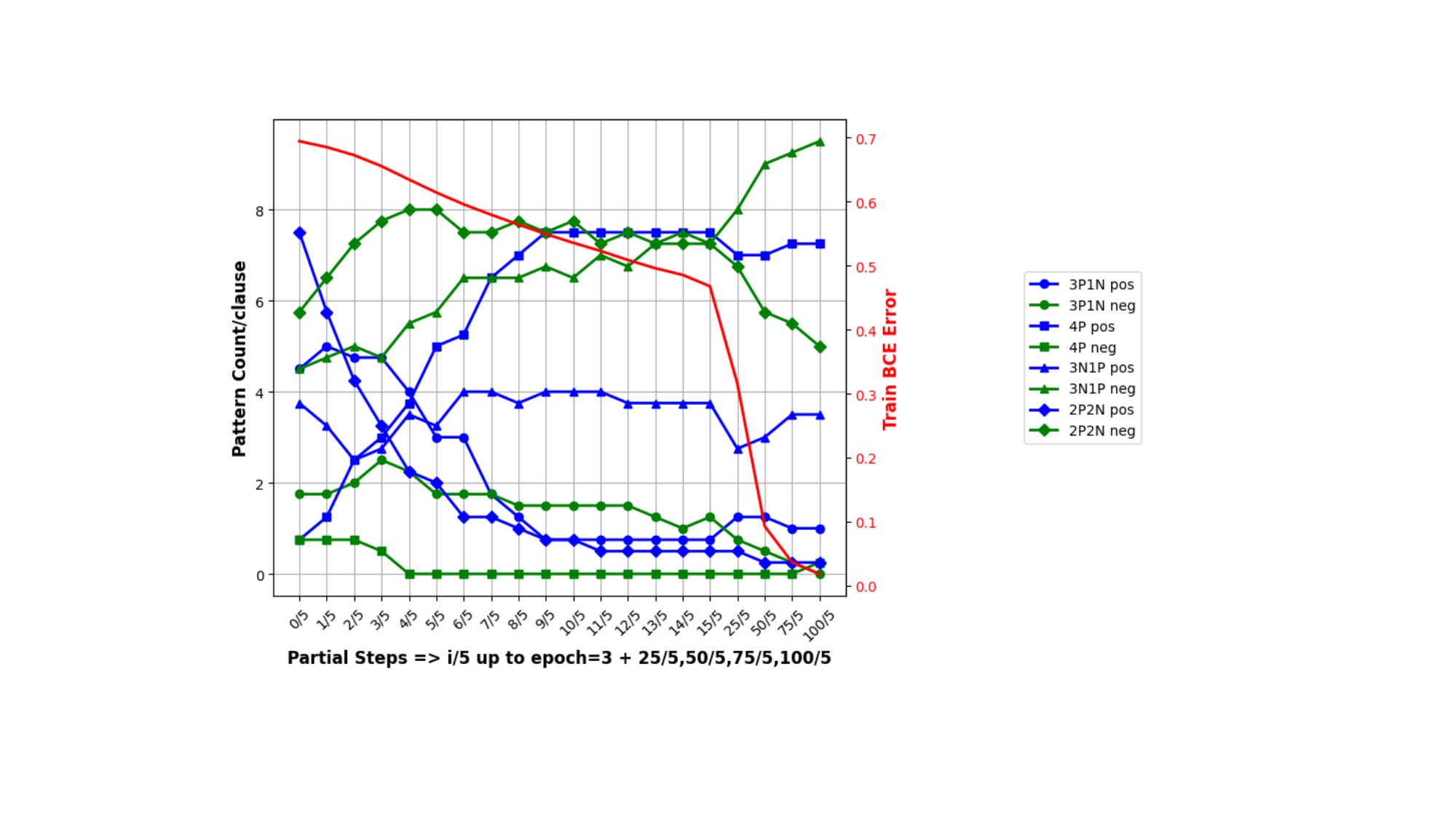}
  \end{subfigure}
  \begin{subfigure}[b]{\textwidth}
    \centering
\includegraphics[width=\textwidth]{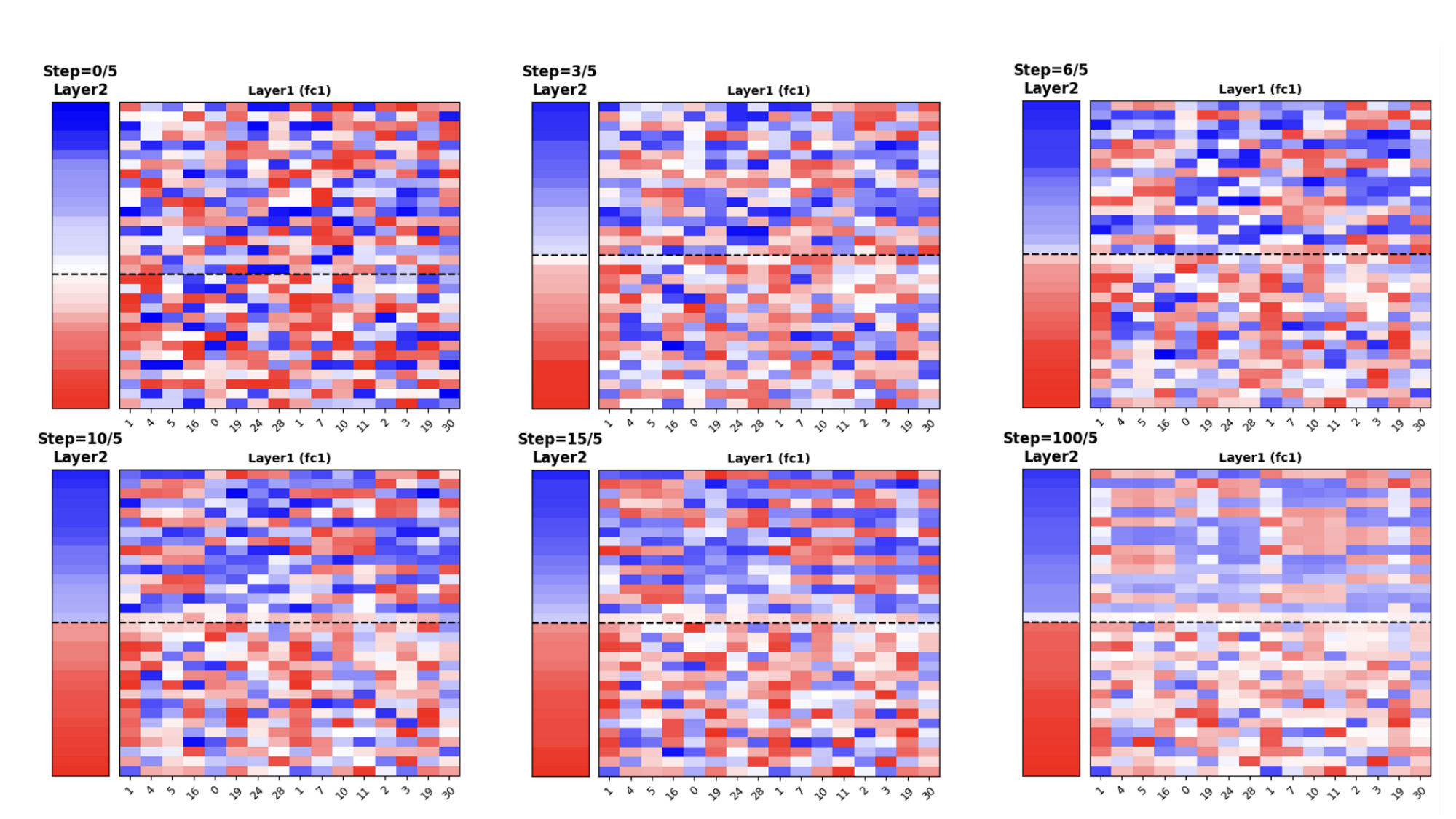}
  \end{subfigure}
  \caption{How codes emerge during training. For clarity, the 20 epochs are not uniformly distributed. Notice in the heat maps of the Layer 1 weights how error decreases as 4P prevelance increases. One can see the emergence of the 4P patterns in the positive witness neurons, and 3N1P in the negative witness neurons.  The values in the graph are an average over 10 runs, but the heat maps are for a single run.}
  \label{emerge matrix}
\end{figure}

We next turn to the question of how the network evolves using gradient descent.  This is depicted in Figure \ref{emerge matrix}.  The results of this graph are interesting in their own right, but they also serve to again highlight the advantage of our combinatorial approach.  We are able to gain insights into ``what is learned when'' during the training process, and these insights also serve to highlight what is important to the model learning process.

In the top part of Figure~\ref{emerge matrix} we track the average number of patterns per clause as a network with $j=32$ neurons and $k=4$ clauses is trained through 20 epochs. To construct this graph, we did the same analysis of patterns types for clauses as we introduced above, but we are now measuring the frequency of the pattern types during the training process. The x-axis shows the number of training epochs, where we split the first 3 epochs into 5 sub-epochs each to allow finer tracking at the start of the training.  Thus, the x-axis is not linear.  The graph also shows, in red, the training error as the training process proceeds. The patterns that emerge are intriguing.  Note that the starting points of the curves are all about where we would expect through randomly setting each matrix entry to either positive or negative with equal probability (which is consistent with our initialization of the training process).  As a result, the training starts with the 2P2N patterns as the most prevalent in both positive and negative rows.  We then see that the network clearly decreases all the positive patterns apart from positive 4P which rapidly increases. Similarly, all the negative patterns apart from 3P1N and 2P2N decrease. This continues until epoch 2, from which point there seems to be little adjustment to the main 4P positive pattern but 2P2N negative starts to transition to 3N1P negative to deliver the final accuracy. 

The final distribution of patterns is similar to what we have seen before, such as in Figure \ref{positive_vs_random}.  In Figure \ref{emerge matrix}, we also depict the heat maps that emerge at the various stages of this training (the graph is an average over 10 runs, but the heat maps are for a single run).  The pattern distributions leave us with a clear understanding of what to expect in the heat maps: the final result is many 4P  patterns as positive witnesses of the clauses and 3P1N patterns as negative witnesses (both appearing more than random), and also many 2P2N negative witnesses, though like all other patterns, they are less than or equal to the number that we would expect to appear at random. One can see that the 4P patterns form clear feature channel codes from step 6.

\remove{
In the next section we will conduct further tests in order to try to better understand the coding patterns that have emerged. However, one thing is pretty clear: feature channel codes are an emergent phenomena resulting from gradient descent.} 

\remove{ 

\begin{figure}[hbt!]
  \centering
  \hspace{1in}\begin{subfigure}[b]{0.65\textwidth}
    \centering
\includegraphics[height=2.5in]{images/plot_pdfs/emerging average 10 runs.png}
  \end{subfigure}
  \caption{How codes emerge during training. The average of 10 runs. As can be seen,  training accuracy is strongly correlated with the emergence of 4P positive  patterns together with 3N1P and 2P2N negative patterns. All other patterns are strictly decreased or not increased beyond their initial random distribution.  }
  \label{emerge matrix}
\end{figure}
} 

One of the most interesting aspects of the graph is the 4P positive curve, which has increased significantly before even the first epoch has completed, and in fact does not increase after the second epoch.  This makes it clear that the 4P codes are an important part of how the model solves the problem and that gradient descent has a relatively easy time of finding them.  More generally, the positive coding patterns converge to their desired result relatively quickly.  However, at the point where those positive rows are stable (around the point labeled 10/5), the error is still around 55\%.  This was surprising to us, since we will see below that in some cases it is possible to get a very good approximation to the learning of the Boolean formula using only the positive rows of the weight matrix.  But in the specific case studied here, the negative rows seem to be crucial.  In those rows, 2P2N gets off to a fast start, but the system eventually realizes that 3N1P is a better pattern to use, and it isn't until 3N1P replaces much of the 2P2N that the training error actually disappears.

\subsection{Clauses With a Negated Variable}
\label{Section: 3P1N}

So far we have shown how feature coding appears in DNFs that have all positive variables. In this section we show how features with negative variables are coded, by studying DNF clauses with 3 non-negated variables and 1 negated variable.  We will see that with this slightly more complex formula, the overall coding behavior is very similar to the fully positive case.  We also use this example to show how computation with feature channel codes proceeds.  

\begin{figure}[h]
  \centering
  \begin{subfigure}[b]{\textwidth}
    \centering    \includegraphics[trim=.5in 0in .5in 0in, clip=true,width=\textwidth]{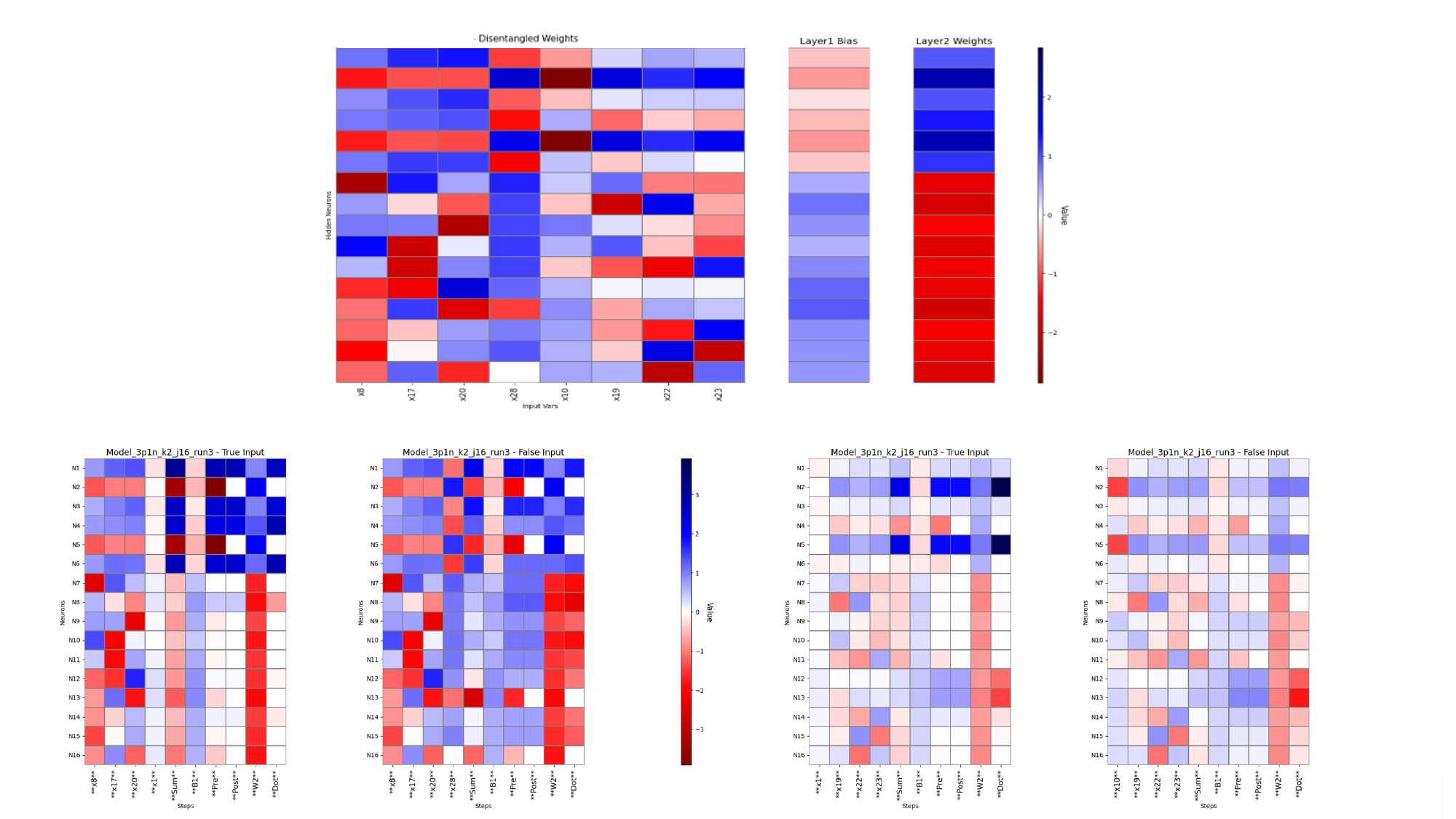}
  \end{subfigure}
  \caption{Execution traces show how channel coding in a model with clauses that have a negated variable each deliver the appropriate positive or negative response. 
  }
  \label{3P1N k=2 j=16}
\end{figure}

We trained the same simple network with a single hidden layer as in our prior experiments, here with 32 input Boolean variables and 16 hidden neurons. In the top part of Figure~\ref{3P1N k=2 j=16}, we show the result of training the model with the formula 
$(x_8 \land x_{17} \land 
x_{20} \land \neg x_{28}) \lor
(\neg x_{10} \land x_{19} \land 
x_{22} \land x_{23})$. For legibility we are only showing the columns corresponding to the 8 variables in the 2 clauses, not all 32 of them. The input features are sorted in the order of the clauses. A quick glace at the matrix shows us the feature channel coding patterns. As before, the network has converged to have a negative bias in Layer 1 neurons corresponding to positive witnesses (positive Layer 2 weights), and positive Layer 1 bias in neurons corresponding to negative witnesses (negative Layer 2 weights). Looking at the Layer 1 matrix entries for the clause  $(x_{8} \land x_{17} \land x_{20} \land \neg  x_{28})$, we see clear positive coding rows emerge in neurons 1, 3, 4 and 6.  For the clause $(\neg  x_{10} \land x_{19} \land x_{22} \land x_{23})$, the positive coding rows are 1, 2, 3, and 5.   Thus, the positive codes overlap in neurons 1 and 3, but the coding for the second clause has smaller magnitude weights in those overlapping rows (which is not a coincidence - see below.)   The Boolean computation performed using these feature channels is as expected: the negated variables have corresponding columns with negative weights while the columns corresponding to positive variables have positive weights.  As before we call this pattern 3P1N.

The bottom part of Figure~\ref{3P1N k=2 j=16} presents four execution traces for the model evaluating two clauses: $(x_8 \land x_{17} \land x_{20} \land \neg x_{28})$ (left two traces) and $(\neg x_{10} \land x_{19} \land x_{22} \land x_{23})$ (right two traces). Each trace details network activation over time (left-to-right), showing initial True variable inputs (first 4 columns), intermediate Layer 1 computations (dot product, bias, pre/post-ReLU values), Layer 2 weights, and the final Layer 1 neuron contributions to the Layer 2 output.

For the first clause, Trace 1 shows a satisfied case ($x_8, x_{17}, x_{20}, x_1$ True and $x_{28}$ (not shown) False). Positive neurons (1, 3, 4, 6) activate according to $x_8, x_{17}, x_{20}$, while negative neurons (7-17) remain mostly inactive, resulting in a correct sigmoid output of 0.9999 (True). Trace 2 shows an unsatisfied case ($x_8, x_{17}, x_{20}, x_{28}$ True). Here, $x_{28}$ inhibits positive neurons (1, 3, 4, 6) via negative weights and activates negative neurons (7-17) via positive weights, resulting in a correct sigmoid output of 0.0015 (False).

For the second clause, Trace 3 demonstrates that in a satisfied case $x_{19}, x_{22}, x_{23}, x_1$ True and $x_{10}$ (not shown) False), positive activation occurs primarily through the set of neurons (2 and 5) that does not overlap with the activations in Trace 1 above.  This indicates something impressive: the network has found feature channel coding that mostly separates clause representations.   Trace 4 shows an unsatisfied case ($x_{10}, x_{19}, x_{22}, x_{23}$ True).  Variable $x_{10}$ suppresses the positive signal in neurons 2 and 5 via negative weights and activates negative witness neurons (9-16) via positive weights, leading to a False output.

As before, we could have hand constructed codes based on two monosemantic neurons, one for each of the clauses, and used large negative biases to provide a clean signal.  However, the technique that gradient descent found involves coding using both positive and negative witnesses, where the negative witnesses might be partly playing the role of the bias in the final computation. In the final result, the coding is clear but uses varying weights, not clean binary values as in the hand crafted combinatorial feature channel codes presented in \cite{adler2024}, most likely due to the pecularities of gradient descent. We believe the binary combinatorial setting may prove to be a good test ground for understanding this process better. 

\begin{figure}[hbt!]
  \centering

  \begin{subfigure}[b]{1.05\textwidth}
    \centering
\includegraphics[trim=0.9in 2in 0in 1.3in, clip=true, width=\textwidth]{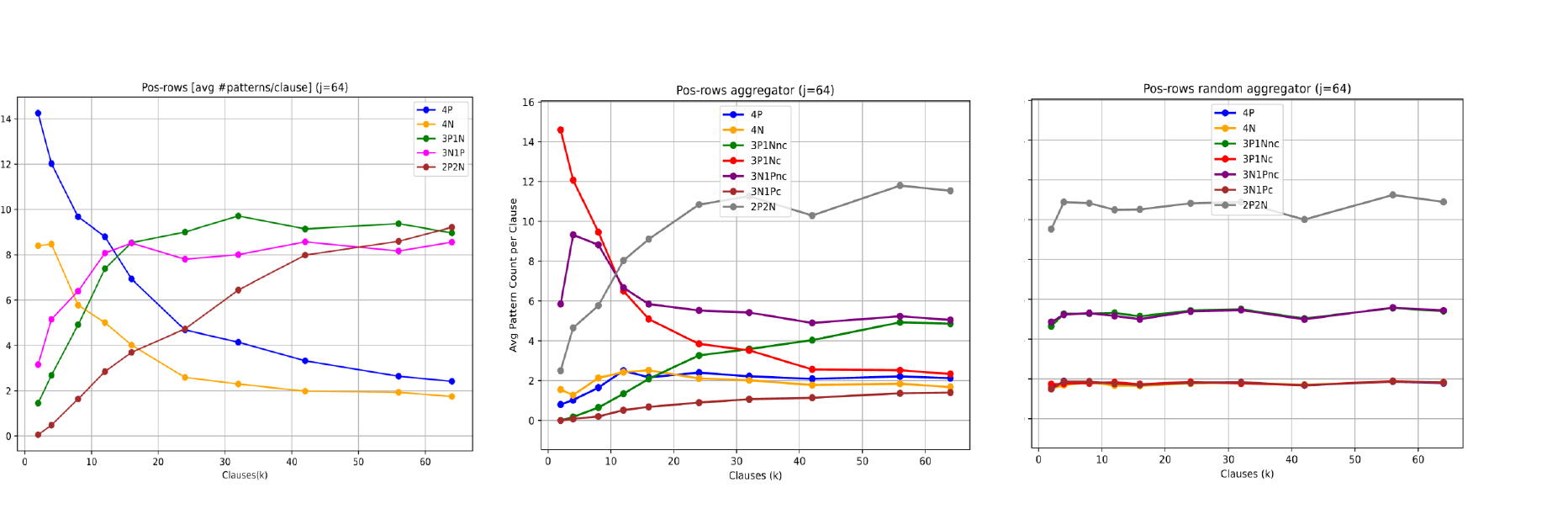}
  \end{subfigure}
  \caption{Left panel: a reminder of what clauses with 4 positive variables look like.  Middle and right panel: Clauses with 3 positive and one negative variable. 
  The middle panel shows the 3P1Nc coding in the positive rows that scales like the 4P pattern we saw for the 4-AND networks in the graph on the left, and very differently than the random distribution on the right.  
  }
\label{3P1N negative_vs_random}
\end{figure}

We also analyzed the prevalence of coding patterns for this case of a DNF with three positive and one negative variable per clause as we did for networks that learn a DNF of clauses with four positive clauses. We ran the same set of tests and benchmarks. We found that the positive witnesses use a 3P1N pattern in which the negative weight in the pattern aligns with the negative variable in the clause. We will call this pattern 3P1Nc, where the $c$ indicates ``aligned with the clause,'' differentiating it from 3P1Nnc, the appearance of a 3P1N pattern where the negative weight is ``not aligned with the clause.'' The left side of Figure~\ref{3P1N negative_vs_random} shows the 4-AND DNF positive rows coded by 4P for the case $j=64$ that we already saw in Figure~\ref{positive_vs_random} (with all positive variables). The middle of Figure~\ref{3P1N negative_vs_random} shows the same type of plot for clauses that include a negative variable, and thus include the patterns 3P1Nc and 3P1Nnc.  The right hand side shows the corresponding random pattern distribution. As can be seen, the neural network is using 3P1Nc as the clear non-random positive coding pattern for clauses with 3 positives and a negative, with a plot in red that looks very similar to the blue 4P pattern on the right, while the 4P patterns in this case look pretty much random, as we would expect. We do not show the coding in the negative rows, but report that it uses 2P2N for lower $k$'s but eventually saturates as $k$ grows, looking similar to random.

\section{Further Examples of Feature Channel Coding}
\label{section : further features}

We next take a look at other examples of problems that the network used feature channel coding to solve: ORs, clauses in CNF form, and a toy vision problem.  These help further clarify this techniques versatility and generality.

\subsection{OR and Conjunctive Normal Form}

We here study the difference in how feature channel coding works between computing an AND and computing an OR.   Within the framework of soft Boolean logic, the difference between an AND and an OR is the bias.  If we ignore the magnitude of the final result, then $x_1 \land x_2 = $ReLU$(x_1+x_2-1)$, while $x_1 \lor x_2 = x_1+x_2$.  Thus, we expect gradient descent to find solutions where AND has significant negative bias, whereas OR have minimal bias.  To study this, we first look at a very simple Boolean formula, where the output is simply the OR of all the variables in the system. We compare this to the type of formula presented in Figure \ref{paired}.  We studied the same network as in that scenario (hidden dimension equals input dimension equals 16, single neuron at the second layer), and we examined the Layer 1 bias for the neurons at layer 1 with positive Layer 2 weights associated with them.  

\begin{table}[ht]
    \centering
    \begin{tabular}{lccc}
        \hline
        & Average layer 1 bias & Maximum absolute layer 1 bias \\
        \hline
        AND & -0.61 & 1.54 \\
        OR & 0.0079 & 0.0306 \\
        \hline
    \end{tabular}
    \caption{Comparison of AND vs OR layer 1 bias for neurons with positive layer 2 weight.}
    \label{and_vs_or}
\end{table}

Table~\ref{and_vs_or} summarizes the result of 10 random network trainings, each consisting of 10000 randomly generated inputs (settings of the variables). 
 For AND, the inputs are as described for Figure \ref{paired}.  For OR, each input is a 1 independently with probability 0.043 and 0 otherwise; this provides approximately equal probability of a positive and negative instance.  In this table, we see that the result are as expected: the AND function has a negative bias, which on average is large enough to have an impact, although as already discussed, the value of the bias tends to be a bit smaller than the theory would predict.  The OR function, on the other hand, has a very small bias, with a negligible impact on the compuatation.

In an effort to further study the OR function, a seemingly good test bed would be to study formulas in Conjunctive Normal Form (CNF), which is the AND of a number of clauses, where each clause is the OR of a number of literals.  The network solves the DNF formulas studied above by computing the AND clauses at the first layer, and then using the single neuron at the second layer to compute an OR of those results, and so our expectation was that the network would solve a CNF formula by computing the OR clauses at the first layer, and then use the second layer to compute the AND of those results.  Our expectation was wrong, but we actually found something more interesting.

To study CNF, our neural network setup is the same as the above, with input dimension 16 and hidden dimension 16.  We study CNF formulas where every clause is the OR of two (positive) variables, and each variable is used exactly once (the \textit{feature influence} \cite{adler2024} of the network is 1). To train any given formula, we want to construct a roughly even split between True inputs and False inputs; we do this simply by choosing every variable to be True independently with probability 0.75. We studied a number of CNF formulas chosen randomly from the set of such formulas, and saw very consistent results.  We here present results for one such randomly chosen formula:

\[
(x_3 \lor x_{11}) \land (x_9 \lor x_{12}) \land (x_2 \lor x_8) \land (x_0 \lor x_{10}) \land (x_{13} \lor x_{14}) \land (x_4 \lor x_7) \land (x_1 \lor x_5) \land (x_6 \lor x_{15}). 
\]

\begin{figure}[h]
  \centering
  \begin{subfigure}[b]{\textwidth}
    \centering

  \includegraphics[trim=1.1in 0in 0.0in 0in, clip=true, width=1.2\textwidth]{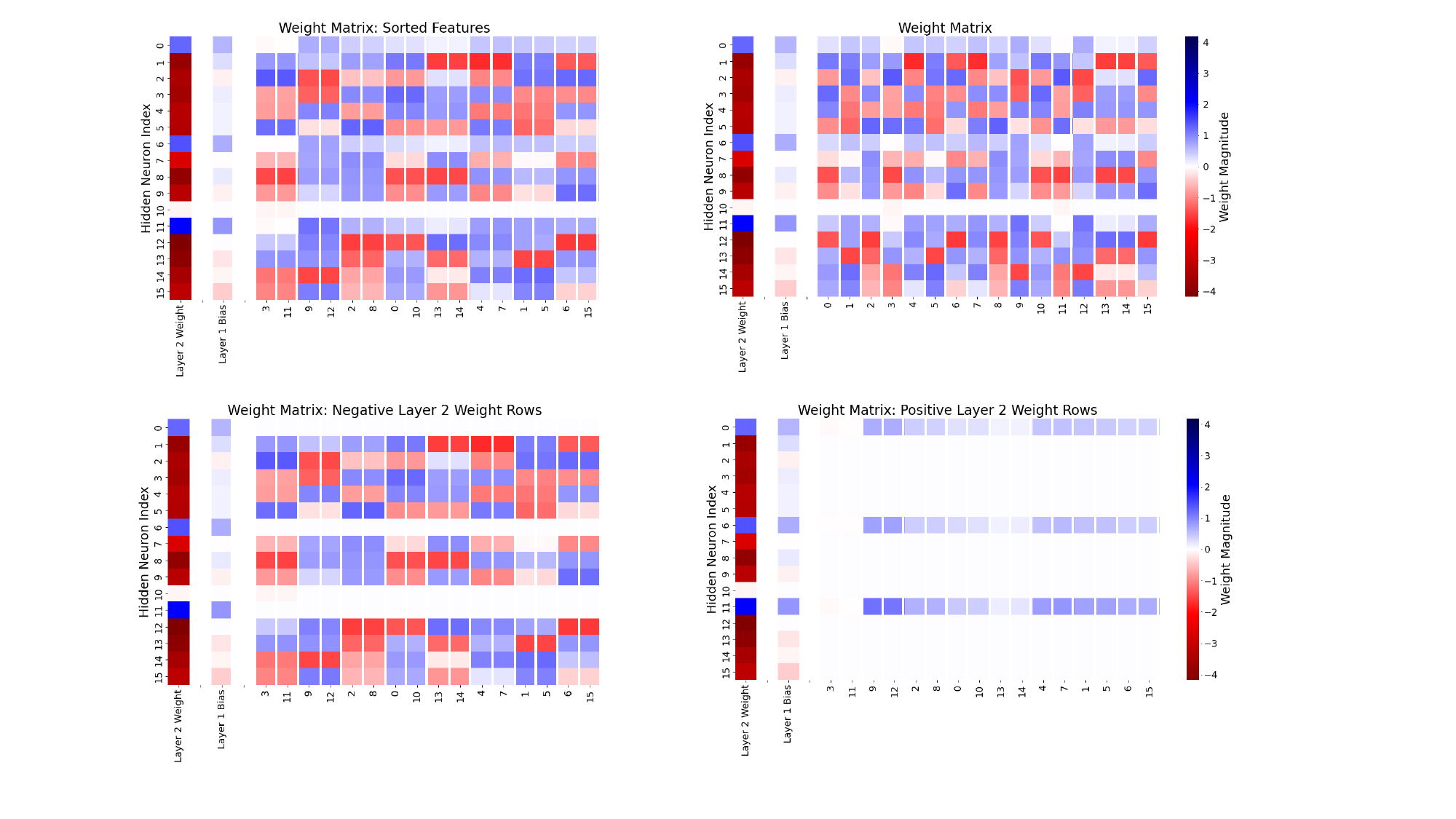}
  \end{subfigure}

  \caption{All weights and biases for a neural network trained on the Boolean formula\\ $(x_3 \lor x_{11}) \land (x_9 \lor x_{12}) \land (x_2 \lor x_8) \land (x_0 \lor x_{10}) \land (x_{13} \lor x_{14}) \land (x_4 \lor x_7) \land (x_1 \lor x_5) \land (x_6 \lor x_{15})$ 
  }
  \label{cnf}
\end{figure}

In Figure \ref{cnf}, we depict what the network learns for this formula with 40,000 randomly chosen inputs.  We see in this figure that the network is not learning the AND as we had expected, and in fact, there is not much that is interesting happening on the small number of positive layer 2 weight rows at all.  On the other hand, the negative rows look much more interesting, and they take up the majority of the neurons.  Furthermore, those rows do fit the pattern of feature channel coding, but they also do not look like they are computing the formula as described.  On closer inspection, we see that the network has actually learned something clever: it has used a different representation of the same logical function, specifically:

\[
\neg \left(
\begin{aligned}
(\neg x_3 \land \neg x_{11}) \;\lor\;
(\neg x_9 & \land \neg x_{12}) \;\lor\;
(\neg x_2 \land \neg x_8) \;\lor\;
(\neg x_0 \land \neg x_{10}) \;\lor\; \\
& (\neg x_{13} \land \neg x_{14}) \;\lor\;
(\neg x_4 \land \neg x_7) \;\lor\;
(\neg x_1 \land \neg x_5) \;\lor\;
(\neg x_6 \land \neg x_{15})
\end{aligned}
\right)
\]

This can be seen to be equivalent to our original presentation of the formula by applying De Morgan's Rule at two different levels: once to the individual OR's of the clauses, and then again to the overall AND of the clauses.  Of course, the network has no actual knowledge of De Morgan's Rule - it is simply given a subset of the truth table for the formula, and it learns some representation of that truth table.  It is interesting to note that based on this example, the network does seem to have a propensity to learn DNF formulas instead of CNF formulas.

In a bit more detail, we see that the network is actually learning this formula as follows.  The outer negation is realized through the utilization of the negative layer 2 weights: if a positive signal comes through the channel codes, then that is converted into a significant negative value by these negative weights.  The neuron at the second layer takes the OR of these negative weights: if any one of them evaluates to a (large enough) negative value, then the entire system evaluates to a negative value.  If none of them are negative, then the small positive weights on the postive layer 2 weight rows are enough to make the whole system positive.  Each of the clauses is computed through a distinct feature channel code, and relies on the fact that $(\neg x_1 \land \neg x_2)$ can be computed in soft lagic as ReLU$(b - x_1 - x_2)$ where $b=1$ in the binary case.   Due to the negative values of the variables, the codes take on negative values (and so are depicted in red in this figure).  We note that the layer 1 bias is usually positive (corresponding to the +1 in that formula), but it is not usually as large as that formula dictates, which is analogous to what we see with DNF formulas, except with the opposite sign.  Here though, some of the rows are negative, which is not analogous to the DNF formulas covered above, where we saw consistently negative bias in the positive weight rows.  However, this is compensated for by positive values that appear elsewhere in the row.  Due to the way the variables were set for the inputs, there is a high likelihood that every row will have an additional positive contribution, to make up for the slightly smaller bias.

\subsection{Code Accuracy in a  One Dimensional Vision Problem}
\label{section: vision}
We next turn our attention to a problem that extends beyond Boolean formulas, into a pattern matching problem that could be viewed as a one dimensional vision problem.  We here consider the {\em consecutive four} problem: given an input string of binary values, are there 4 ones in a row?  Thus the individual input variables are again binary.  We consider this an extremely simple vision problem, in the sense that it deals with pattern matching with locality.  With that locality in mind, we we will train and test our network with a restricted set of inputs to this problem: all inputs have exactly six ones in the string, and they must all be close to each other - within a region of eight consecutive bits of the input string.  We want to see if neural networks trained to solve this problem have properties that are consistent with what we see with pure Boolean problems.

We point out that, like many pattern matching problems, this problem can be solved with a Boolean formula: the OR of all four ways ANDs of four consecutive bits.  And in fact, even though the training process is not given this representation (only the ordered sets of inputs and their labels), using the Combinatorial Interpretability approach, we see that the networks we train to solve this problem do in fact use exactly that formula.

\begin{figure}[h!]
\centering
\makebox[\textwidth][c]{%
\includegraphics[width=1.0\linewidth]{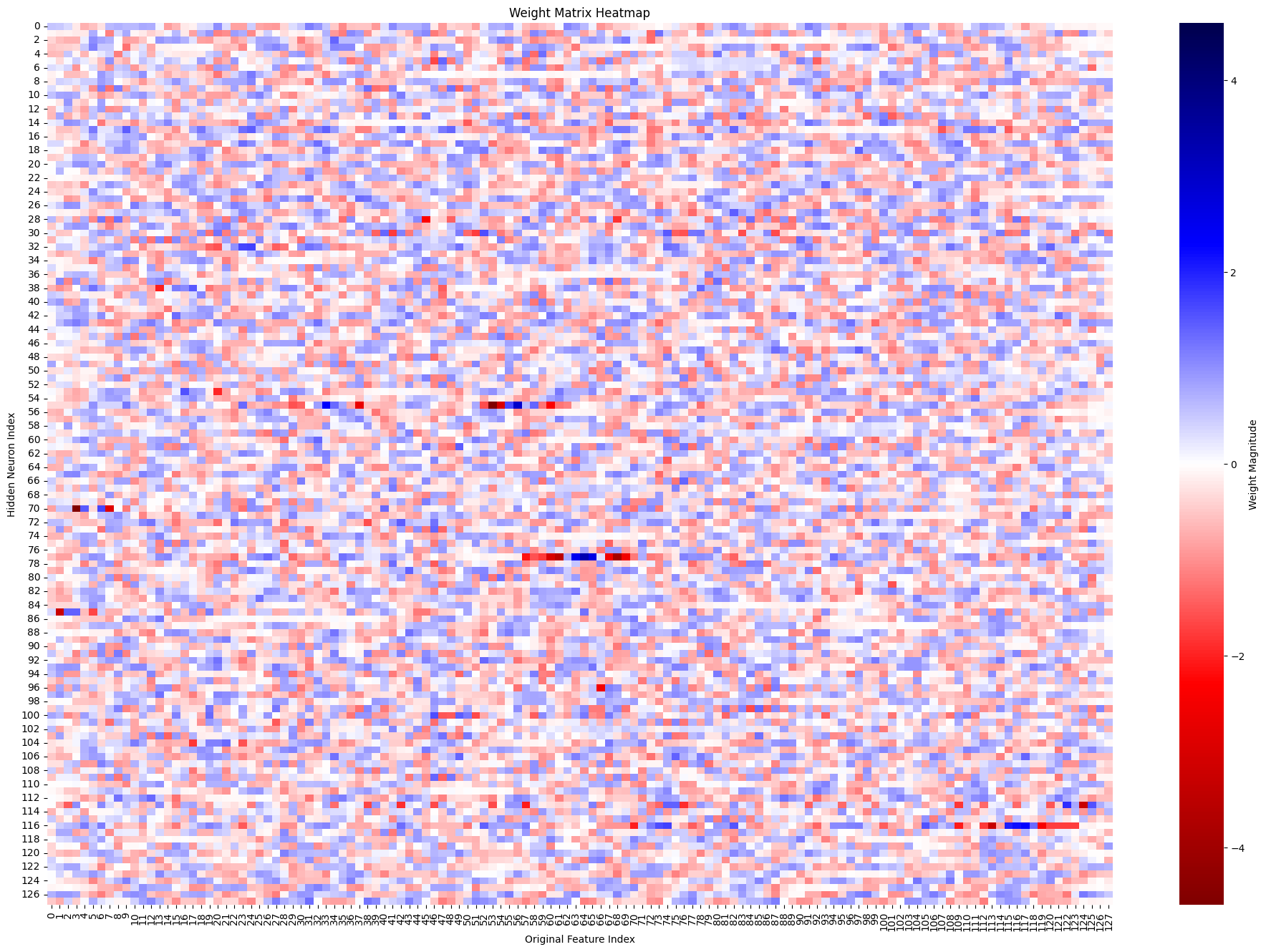}
}
\caption{Layer 1 heat map for the connect four problem. The x-axis corresponds to the pixels of the one dimensional image we the network receives as input.}
\label{connect4all}
\end{figure}

Our setup is as follows.  We use a network with 128 input variables, 128 hidden neurons in the single hidden layer, and a single output neuron.  We generate each training and test data input as follows:
\begin{itemize}
    \item Flip a fair coin for each input to determine if it will be positive or negative.
    \item In either case, choose a sequence of eight consecutive variables uniformly at random.
    \item For a positive input, out of the set of eight chosen variables, pick a consecutive set of four variables uniformly at random.  Also choose two other variables from the eight uniformly at random.  These six variables are set to 1, and all others are set to 0.
    \item For a negative input, randomly pick a consecutive set of four variables out of the eight, but only set three of them to 1.  Then pick another three of the eight variables uniformly at random, and set those variables to 1 as well.  Then test if the result has four consecutive variables set to 1.  If it does, repeat until a negative input is generated, or 20 attempts have failed, in which case use the string of all zeros.
\end{itemize}

We generated 10,000 training inputs, and trained for 20 epochs, after which the network had 0 loss.  Accuracy on the test set was 99\%.  The resulting weight matrix, shown in Figure \ref{connect4all}, at first glance, might not seem to exhibit channel coding.

\begin{figure}[h!]
\centering
\makebox[\textwidth][c]{%
\includegraphics[width=1.0\linewidth]{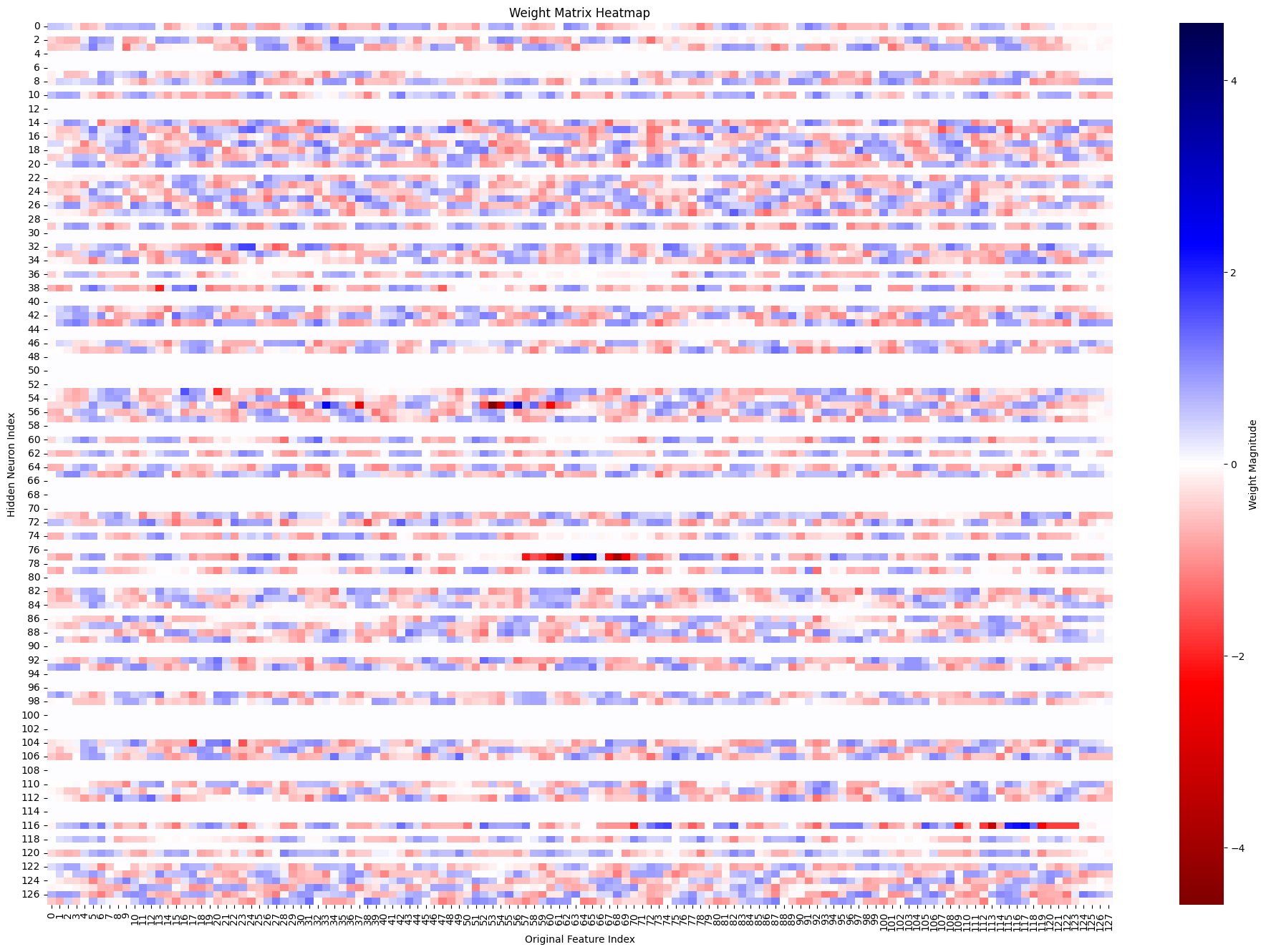}
}
\caption{Layer 1 for connect four: positive witnesses.}
\label{connect4positive}
\end{figure}

However, things become quite a bit clearer when we separate the rows into positive witnesses and negative witnesses, where (as before) a row is considered a positive witness if its Layer 2 weight is positive, and a negative witness if its Layer 2 weight is negative. The positive witness only heat map is depicted in Figure \ref{connect4positive}.  This is a subset of the rows shown in Figure \ref{connect4all}; no other changes have been made.

\begin{figure}[h!]
\centering
\makebox[\textwidth][c]{%
\includegraphics[width=1.0\linewidth]{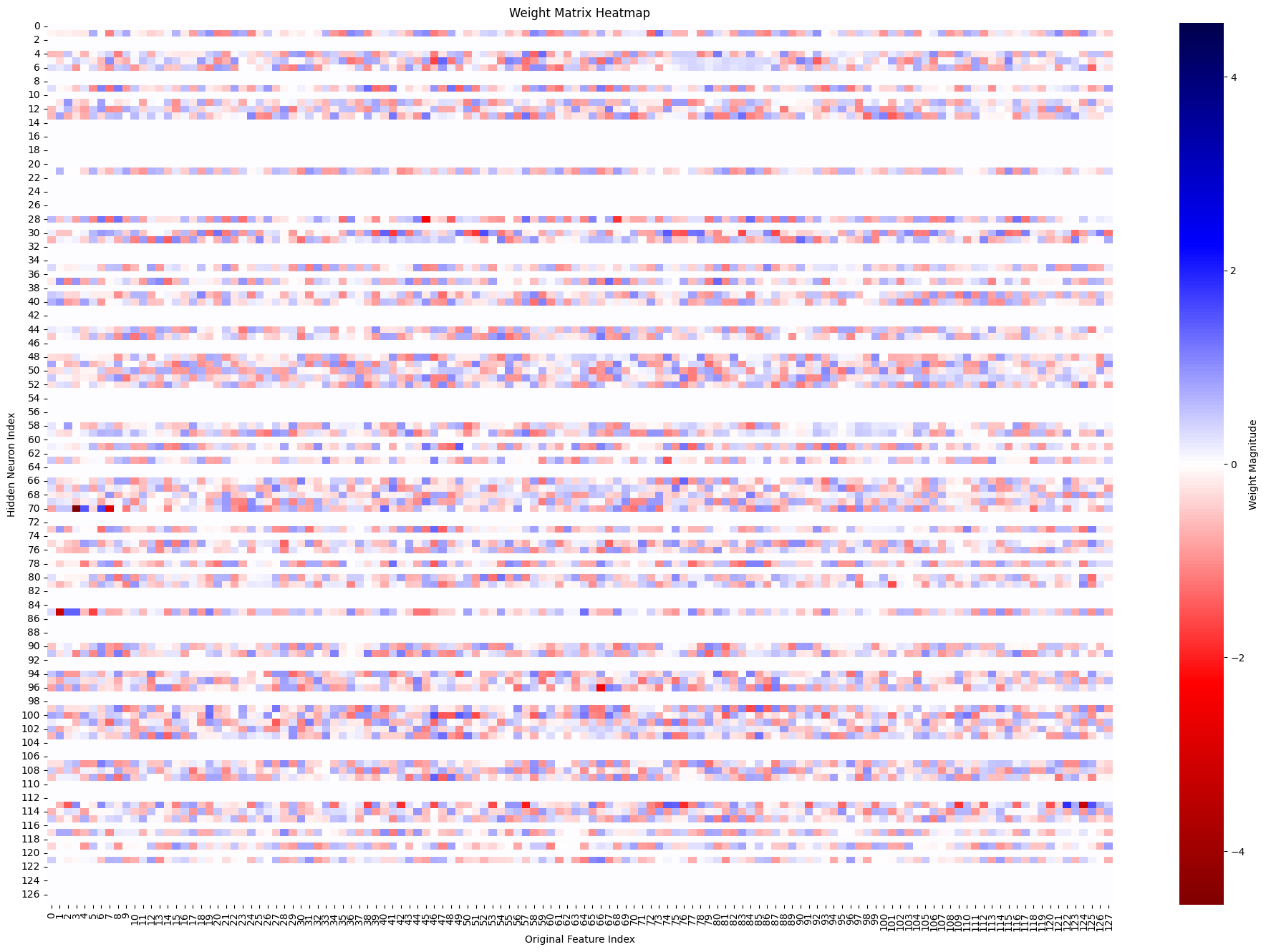}
}
\caption{Layer 1 for connect four: negative witnesses.}
\label{connect4negative}
\end{figure}

We can see visually that the positive witness rows do in fact demonstrate considerable feature channel coding.  And in fact, we see that most of the rows of the (positive only) matrix consist of alternating sequences of four consecutive positive values, followed by a number of negative values.  There are some sequences of ones that are longer than four, but most of them have length exactly four.  This image alone provides convincing evidence that this network uses feature channel coding, but we will demonstrate this with metrics as well.   In this context, we consider a row to be participating in the code for a sequence of four consecutive ones starting at position $i$, if that row (a) corresponds to a positive layer 2 weight, and (b) it has four consecutive positive values starting at position $i$.  The resulting statistics are summarized in Table \ref{connectstatistics}.

\begin{table}[ht]
\centering
\begin{tabular}{@{}lll@{}}
\toprule
\textbf{Statistic}                           & \textbf{Average} & \textbf{Std Dev} \\ \midrule
\multicolumn{3}{l}{\textbf{Codingness Statistics}}                                                 \\ \midrule
Total Number of Unique Coding Rows                              & 70                  & -                           \\
Fraction of Contributions from Coding Rows              & 63.93\%             & 10.29\%                     \\
Number of Coding Rows                                   & 12.77               & 2.73                        \\
Weighted Sum of Coding Rows                             & 9.0139              & 2.1991                      \\ \midrule
\multicolumn{3}{l}{\textbf{Overlap Statistics}}                                                    \\ \midrule
Number of Overlapping Coding Rows                       & 2.35                & 1.64                        \\
Weighted Overlapping Coding Rows                        & 18.24\%             & 12.71\%                     \\ \midrule
\multicolumn{3}{l}{\textbf{ANDness Statistics}}                                                     \\ \midrule
Percentage of Negative L1 Bias Coding Rows              & 100.00\%            & 0.00\%                      \\
\bottomrule
\end{tabular}
\caption{Summary of Codingness, Overlap, and ANDness}
\label{connectstatistics}
\end{table}

These statistics clearly indicate that coding is being used here: an average of 12.77 rows code each possible positive example, and there is only an average overlap of 2.35 between rows.  Furthermore, all coding channels clearly express a compution closely resembling an AND of the four bits that are being coded for: every single coding row has a negative bias (even though the biases are initialized uniformly and independently at random between -1 and 1).

The negative witnesses are depicted in Figure \ref{connect4negative}. We see here that the negative witnesses consist of alternating sequences of one positive value and one negative value, essentially computing an XOR.

We next show that we can extract and perform classification using the codes that are depicted in Figure \ref{connect4positive}, and measured in Table \ref{connectstatistics}.  We can view this process as a combinatorial replacement (for this specific network) of a sparse autoencoder: these codes correspond to how each feature is represented by the neural network by its positive coding rows.  A feature here is any set of four consecutive ones in the input string. Extracting the codes is straightforward: we simply sweep across the different starting positions, and for each starting position extract the codes for that position per the definition of coding rows.  

To use these extracted codes for classification, we construct a decoding algorithm that maps the Layer 1 post-activation values in positive rows of any input to a set of features that are active.   In other words, the algorithm attempts to make a decision as to where there are four consecutive ones in the input string, not based on the entire neural network, but instead only on the Layer 1 post-activation values, and of those values, only those that are associated with positive witnesses.  This serves three purposes: (1) it further demonstrates how to combinatorially interpret the trained model, (2) it demonstrates how well we can do with only that subset of information (and specifically without using any of the negative witnesses, and (3) it also demonstrates that the network is really learning the actual underlying features, as opposed to the summary (binary) classification problem it was asked to solve.

The algorithm proceeds by computing the Layer 1 post-activations of the network, and then comparing these values to the codes.  Any feature is considered to be present if its code is entirely present.  A code is entirely present if all of its rows have a post-activation value that is at least the sum of its four positive elements, minus its bias, minus 1.9 times the maximum value within its sliding window of eight consecutive bits.  We test this algorithm on a set of inputs drawn from the same distribution as both the training set and the test set of inputs, and find that this very simple decoding algorithm does a surprisingly good job of not just differentiating 0 inputs from 1 inputs, but also determining which features are actually present.  The results from a single training run (and accompanying extraction of codes), tested on a set of 40 random test inputs is provided in Table \ref{sampleoutput}.  The actual matches are found simply by scanning the input for four consecutive ones; the code matches are from our extracted codes.

\begin{table}[h!]
\centering
\scriptsize 

\setlength{\columnseprule}{0.4pt} 

\begin{multicols}{2}
\begin{tabular}{c c l l}
\hline
\textbf{Test} & \textbf{Label} & \textbf{Actual Matches} & \textbf{Code Matches} \\
\hline
0  & 1 & [35]             & [35]             \\
1  & 0 & ---              & ---              \\
2  & 1 & [88, 89]         & [88, 89]         \\
3  & 1 & [116, 117]       & [116, 117]       \\
4  & 1 & [11, 12]         & [12]             \\
5  & 0 & ---              & ---              \\
6  & 0 & ---              & [116]            \\
7  & 1 & [18, 19, 20]     & [18, 19, 20]     \\
8  & 0 & ---              & ---              \\
9  & 1 & [40]             & ---              \\
10 & 1 & [15, 16]         & [16]             \\
11 & 1 & [14, 15, 16]     & [14, 15, 16]     \\
12 & 0 & ---              & ---              \\
13 & 0 & ---              & ---              \\
14 & 1 & [47, 48, 49]     & [47, 48, 49]     \\
15 & 0 & ---              & ---              \\
16 & 0 & ---              & ---              \\
17 & 0 & ---              & ---              \\
18 & 1 & [97]             & ---              \\
19 & 1 & [62, 63]         & [62, 63]         \\
\hline
\end{tabular}

\columnbreak

\begin{tabular}{c c l l}
\hline
\textbf{Test} & \textbf{Label} & \textbf{Actual Matches} & \textbf{Code Matches} \\
\hline
20 & 1 & [4]              & [4]              \\
21 & 1 & [106, 107]       & [106, 107]       \\
22 & 0 & ---              & ---              \\
23 & 1 & [29, 30, 31]     & [29, 30]         \\
24 & 0 & ---              & ---              \\
25 & 0 & ---              & ---              \\
26 & 0 & ---              & ---              \\
27 & 1 & [8, 9, 10]       & [8, 9, 10]       \\
28 & 0 & ---              & ---              \\
29 & 1 & [24, 25, 26]     & [24, 25, 26]     \\
30 & 1 & [69]             & [69]             \\
31 & 0 & ---              & ---              \\
32 & 0 & ---              & ---              \\
33 & 0 & ---              & ---              \\
34 & 1 & [80, 81, 82]     & [80, 81, 82]     \\
35 & 0 & ---              & ---              \\
36 & 0 & ---              & ---              \\
37 & 1 & [62]             & [61]             \\
38 & 0 & ---              & ---              \\
39 & 0 & ---              & ---              \\
\hline
\end{tabular}
\end{multicols}
\caption{Sample output from testing: actual matches versus code matches.}
\label{sampleoutput}
\end{table}

We test this more thoroughly by running 50  different training runs, and for each training run testing the accuracy of the algorithm's predictions on a test set of 2000 inputs, drawn from the same distribution as the training set.  We measure success by two criteria: how well did the algorithm predict whether or not there is a matching pattern of four consecutive ones, and how well did the algorithm predict the exact set of matches (which is the empty set if there is no match).  Our results are summarized and compared to the original trained network in Table \ref{codingtests}.

\begin{table}[ht]
    \centering
    \begin{tabular}{lccccc}
        \hline
        & FPR NN & FNR NN & FPR CODES & FNR CODES & FULLY CORRECT \\
        \hline
        Average & 0.71\% & 1.65\% & 6.82\% & 7.66\% & 82.63\% \\
        Std Dev & 0.33\% & 0.37\% & 1.82\% & 0.97\% & 1.80\% \\
        \hline
    \end{tabular}
    \caption{Performance of Coding Interpretability Compared to Original Neural Network.}
    \label{codingtests}
\end{table}

We note that while the results of our algorithm are noticably worse than the original network, it is still over 92\% accurate in terms of the decision problem, and over 82\% accurate in terms of picking out the exact set of matches (which the original network does not do).  These results are important for a number of reasons:
\begin{itemize}
\item They further demonstrate how central the codes are to the computation.
\item Even though the algorithm described above only uses the positive witnesses, they are able to achieve over 92\% accuracy on the decision problem of whether there is a match or not.
\item Even though the neural network was trained just to answer the decision problem, looking at the network through the lens of the resulting codes yields over 82\% accuracy on precisely classifying where all the consecutive four sequences are.  As we see from Table \ref{sampleoutput}, often when it is wrong, it is very close to providing the correct answer (i.e., a sequence of 5 consecutive ones should return two matches, but the coding-based algorithm only returns the first one.)
\end{itemize}

\begin{figure}[h!]
\centering
\makebox[\textwidth][c]{%
\includegraphics[width=0.8\linewidth]{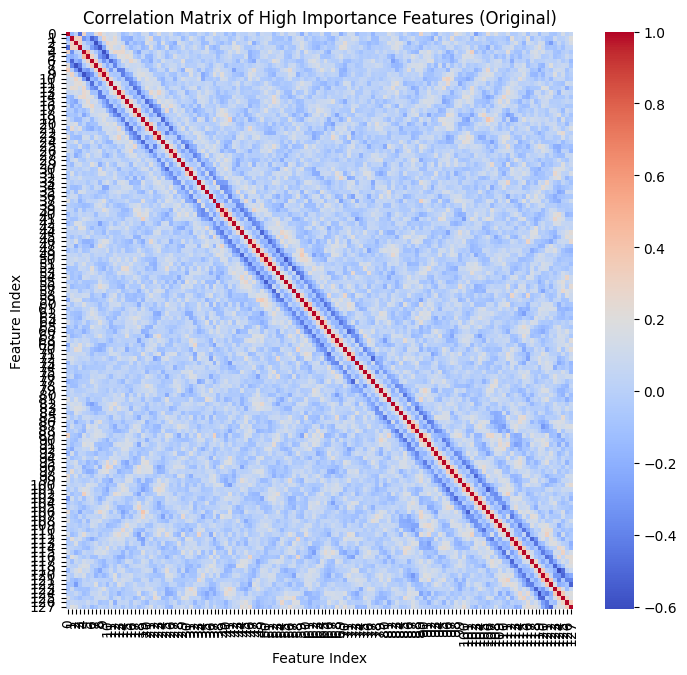}
}
\caption{Correlation Matrix for columns of layer 1 weight matrix.}
\label{fabric}
\end{figure}

We next turn to another phenomena we observed in studying this problem.  This is not directly relevant to our central hypothesis of networks using feature channel coding, but it does demonstrate what can be discovered by using our combinatorial interpretability approach.  To see this phenomena, we looked at all columns of the layer one weight matrix, and we computed all pairwise correlations between them.  This is depicted in Figure \ref{fabric}.  The strong self correlation on the diagonal is of course expected, and we see an expected pattern in the values close to that diagonal.  We did not, however, expect to see such a strong and regular pattern in the upper and lower diagonal regions of this matrix, even for columns that represent inputs that are very far removed from each other.

Our best guess for the cause of this pattern is that there is a strong preference for a new set of four consecutive ones to appear on a different neuron after the current one has finished any existing pattern.  This would cause regions of alternating sequences of “more likely to code” and “less likely to code”, each of length four, which is consistent with the pattern.  This preference presumably is fairly strong for it to not perceptibly fade throughout the entire matrix.

\section{Modeling Computation and Combinatorial Disentanglement}
\label{section: theory}

\begin{figure}[h!]
\centering
\includegraphics[width=1.0\linewidth]{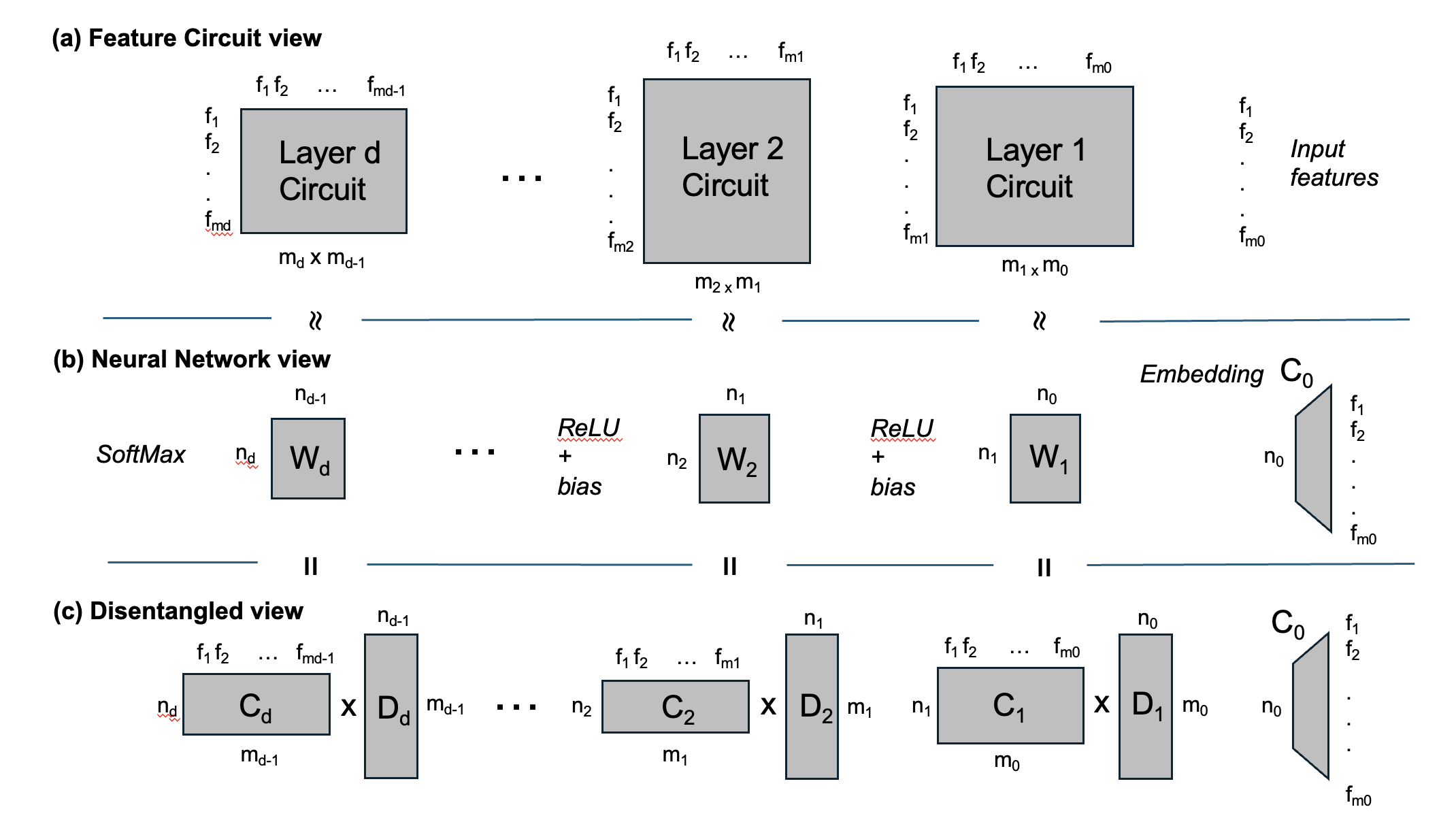}
\caption{The feature circuit coding-decoding view of computation. Computation flows from right to left. The middle sequence (b) is the actual computation through the neural network. The top sequence (a) is the abstract computation mapping features to features in each layer being represented by the neural network. The lower sequence (c) is the feature coding view that bridges the two by thinking of the the weight matrix as implementing a decoding of the prior layer's features to serve as monosemantic inputs to a coding-and-computation of the features of the new one. $W_i = C_i D_i$ is a superposed $n_i \times n_{i-1}$ representation of the $m_i \times m_{i-1}$  $i$-th feature circuit layer.  }
\label{featurelayers}
\end{figure}

Thus far we have illustrated the combinatorial view of a two‑layer network.  Real‐world models—even plain multilayer perceptrons (MLPs)—are usually richer: they begin with an embedding layer and then stack several nonlinear layers. This section introduces a conceptual model for understanding MLP computation based on underlying ``feature circuits.'' We then propose a novel perspective viewing MLP layers as implicitly performing a decode-compute-encode cycle, aiming to disentangle the network and reveal this circuit. Building on this, we outline a potential strategy, termed \textit{cascading feature disentanglement}, for systematically uncovering the feature circuit layer by layer. We present initial steps demonstrating how the first layer of computation can be revealed via disentanglement from an input embedding and also propose a general approach for deeper layers.

\subsection{The Feature Circuit View of Neural Computation}

Our modeling follows the intuitive ideas in \cite{adler2024,toy,vaintrob_superposition}.   We aim to capture the fundamental computational structure implemented by an MLP. We model the computation at each layer $i$ (for $i=1, \ldots, d$) as producing a set of abstract output features $F_i = \{f_{i,1}, f_{i,2}, \ldots, f_{i,m_i}\}$. Each feature $f_{i,j} \in F_i$ is computed by applying a specific function to a subset of the features $F_{i-1} = \{f_{i-1,1}, \ldots, f_{i-1,m_{i-1}}\}$ from the preceding layer (where $F_0$ represents the initial input features, $m_0$ in number). The complete \textit{feature circuit} of the network is the sequence of these computations across all layers, $F_1, F_2, \ldots, F_d$.

\begin{figure}[h!]
  \centering
  \begin{subfigure}[b]{1.0\textwidth}
    \centering
\includegraphics[trim=0.4in 2.8in 0in 2.4in, clip=true, width=\textwidth]{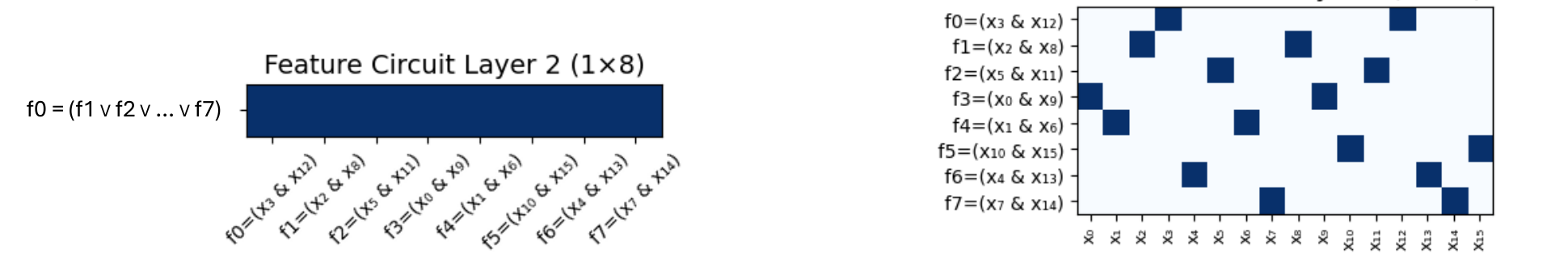}
  \end{subfigure}
  \caption{Two layer feature circuit (as an example of the top row of Figure~\ref{featurelayers}) of the DNF  $(x_3 \land x_{12}) \lor \left( x_2 \land x_{8} \right) \lor \left( x_5 \land x_{11} \right) \lor \left( x_0 \land x_{9} \right) \lor \left( x_{1} \land x_{6} \right) \lor \left( x_{10} \land x_{15} \right) \lor \left( x_4 \land x_{13} \right) \lor \left( x_7 \land x_{14} \right)$. Layer 1 defines the computation of the pairwise AND features and Layer 2 is the single OR feature over the results of the outputs from Layer 1. Together they define the abstract computation that the network in Figure~\ref{disentangled heat map} computes.}
\label{Feature Circuit 1}
\end{figure}

Figure~\ref{featurelayers}(a) provides a visualization of such a multi-layer feature circuit. It takes $m_0$ input features and computes features layer by layer, where the computation of a feature $f_{i,j}$ depends only on a subset of features from $F_{i-1}$. For instance, in our earlier simple 2-layer network example (a fully connected layer followed by a single neuron computing a DNF of 2-ANDs), the feature circuit might resemble Figure~\ref{Feature Circuit 1}.  Note that in that figure, the rectangular portion of the matrix depicts the dependencies (using filled in rectangles), and the formulas to the left of the rectangle depict the actual functions being computed.

This abstract feature circuit finds a concrete implementation in a standard MLP, depicted in Figure~\ref{featurelayers}(b). The $m_0$ input features are first mapped by an embedding matrix $C_0$ (size $n_0 \times m_0$) to an $n_0$-dimensional vector. Subsequent layers $i=1, \ldots, d$ consist of a weight matrix $W_i$ (size $n_i \times n_{i-1}$), a bias vector $b_i$, and a non-linearity (e.g., ReLU), transforming the $n_{i-1}$ neuron activations from layer $i-1$ into $n_i$ neuron activations for layer $i$. (While we use ReLU for illustration, other operations like batch normalization could be used instead). A final readout function (e.g., softmax) produces the network's output. Through training (e.g., via gradient descent), this MLP learns to approximate the underlying feature circuit computation.

We will say that a layer $i$ of a neural network {\em computes in superposition} if $n_i < m_i$, that is, there are fewer neurons than output features in the circuit being computed by this layer. The network as a whole will be said to compute in superposition if its layers compute in superposition. Such networks are sometimes called \textit{polysemantic} because each neuron participates in the computation of more than one output feature, as opposed to monosemantic ones, in which each neuron corresponds to a single output feature. 

As \cite{adler2024,vaintrob_superposition} show, a neural network can use a polysemantic representation with many fewer neurons than features to compute across multi-level neural networks. Part of our observations in this work suggest that polysemanticity can arise in training even when the network does not need to be in superposition, that is, $n_i \geq m_i$ (similar to results shown in \cite{Lecomte2024WhatCauses}).  In some neural networks, such as large language models or convolutional neural networks, the input to the neural network is provided in superposition.  In these cases, the input is encoded into superposition via a first layer that translates text (via tokens) or an image (via downsampling) into an embedding space.  

\subsection{The Disentangled Feature Circuit View}

Our core proposal for interpreting MLPs involves viewing each layer's weight matrix $W_i$ as implicitly factoring into two conceptual matrices, $C_i$ and $D_i$, revealing an intermediate ``disentangled'' representation, shown in Figure~\ref{featurelayers}(c). This perspective aims to bridge the gap between the concrete MLP and the abstract feature circuit.

Specifically, we hypothesize that the computation within layer $i$ can be understood as:
\begin{enumerate}
    \item \textbf{Decompression ($D_i$)}: An implicit matrix $D_i$ (size $m_{i-1} \times n_{i-1}$) acts on the $n_{i-1}$ neuron activations from layer $i-1$. Its role is to transform the $m_{i-1}$ polysemantic features computed by the previous layer ($F_{i-1}$) into approximations of their  monosemantic representations. 
    \item \textbf{Computation and Compression ($C_i$)}: An implicit matrix $C_i$ (size $n_i \times m_{i-1}$), combined with the bias $b_i$ and non-linearity $\sigma_i$ (e.g., ReLU), acts on the monosemantic representations. It performs two roles simultaneously, using feature channel coding:
    \begin{itemize}
        \item It computes the $m_i$ new features for the current layer ($F_i$) based on the $m_{i-1}$ input features.
        \item It \textit{compresses} these $m_i$ computed features back into the $n_i$-dimensional polysemantic represenations for the next layer. 
    \end{itemize}
\end{enumerate}

Thus, the effective transformation by the weight matrix $W_i$ in the standard MLP (Figure~\ref{featurelayers}(b)) is conceptually equivalent to the combined operation $W_i \approx C_i D_i$ in our disentangled view (Figure~\ref{featurelayers}(c)). The entire network can then be viewed as a sequence $$C_d D_d \sigma_{d-1} \ldots \sigma_1 C_1 D_1 (C_0 \cdot \text{input}),$$ where $\sigma_i$ represents the bias and non-linearity applied after the $C_i D_i$ (or equivalently, $W_i$) multiplication. The actual MLP uses the compact $W_i=C_iD_i$ form, hiding the intermediate dimension of size equal to the number of features at layer $i$. The names $D_i$ (Decompress) and $C_i$ (Compress) reflect their roles in moving between the compact polysemantic representation and the potentially larger monosemantic representation. 

This perspective is similar to concepts like sparse autoencoders and transcoders, where encode/decode operations learn similar transformations. Specifically, the $D_i$ matrix resembles an encoder and the $C_i$ matrix a decoder of a transcoder. However, a key distinction is that $C_i$ and $D_i$ are posited as \textit{intrinsic} components of the \textit{existing} trained weight matrix $W_i$, not externally added modules learned via a separate objective like in standard autoencoders or transcoders. Our goal is to \textit{recover} this inherent $C_i D_i$ structure. Consequently, the placement of non-linearities and biases in our model follows the original network architecture, differing from typical autoencoder/transcoder setups.

Interestingly, if one ignores bias and ReLu, then notice that the feature circuit for any layer $i$ could be given by $D_{i} * C_{i-1}$, a matrix of size $m_i \times m_{i-1}$ in which the codes of length $n_i$ are matched between $D_{i}$ and $C_{i-1}$ yielding the layer $i$ feature circuit matrix. So for example, as depicted in Figure~\ref{featurelayers}, the Layer 1 feature circuit would be given by multiplying $D_{2}$ by $C_{1}$.  This matrix is only an approximation because of the bias and ReLU, whereas $C_{1} D_1$ is an accurate representation of the neural network matrix $W_1$. 

\subsection{How the Feature Channels Code Computation}

Let us spend a moment revisiting how feature channel coding defines computation in this view of the neural network. We assume that $f_{i,j}$, a feature to be computed at layer $i$, is a function of a small set of $k$ features $G =\{ f_{i-1,1} \ldots f_{i-1,k} \}$ from layer $i-1$. The features in $G$ of the feature circuit view are each coded (n the neural network view) by a set of neurons from layer $i-1$, and $f_{i,j}$ is coded by a set of neurons from layer $i$. To perform the computation for $f_{i,j}$, the weight matrix $W_i$ maps each of the codes for features in $G$ to the code for $f_{i,j}$.  When a feature at layer $i-1$ is used for multiple features at layer $i$, it gets mapped to each of the codes for the features it is used in.  Thus, prior to the non-linearity for the activation at layer $i$, each of the inputs to $f_{i,j}$ have been mapped to the same set of neurons at layer $i$.  This allows the network to use the code for $f_{i,j}$ as a computational channel for computing $f_0$.

The entangled matrix $W_i$ performs this mapping for the features used by $f_{i,j}$. The disentangled view makes this clear: $D_i$ first maps those features to their monosemantic representation.  $C_i$ then brings them all together to the code for $f_{i,j}$. In a funny way, in terms of feature channel coding, $D_i$ is a \textit{decoding} matrix, moving codes into features, and $C_i$ is a \textit{coding} matrix, encoding them back into new collections of features, exactly the opposite of the naming convention in the autoencoding/transcoding literature \cite{anthropic2023autoencoder,transcoders}. The matrix $C_i$, combined with the bias and the non-linearity, performs the computation required for $f_{i,j}$.  In the examples provided in earlier sections, the inputs were already in their monosemantic representation, so there was no need for a $D$ matrix; the layer one weight matrix serves directly as the $C$ matrix as well.
Note that the channel's code is somewhat independent of the computation being performed - the goal of channel coding is to bring together the necessary inputs onto the same computational channel.  Once they are on the same channel, a number of different functions can be performed on those incoming features. Thus, uncovering the codes for a given neural network would allow us to describe its feature circuit. 

\subsection{Cascading Feature Disentanglement}

We propose the \textit{cascading feature disentanglement} technique as a potential avenue for recovering the feature circuit from a trained MLP. The core hypothesis is that we can iteratively uncover the matrices $C_i$ and $D_i$ layer by layer.

\textbf{Cascading Disentanglement Technique:} For a $d$-layer MLP with weights $W_1, \ldots, W_d$ and input embedding $C_0$:

\begin{enumerate}
    \item \textbf{Initialization (Layer 1):} Estimate $C_1$. Since there is typically no non-linearity between the input embedding $C_0$ (size $n_0 \times m_0$) and the first weight matrix $W_1$ (size $n_1 \times n_0$), the initial disentanglement is simply $C_1 \approx W_1 C_0$. The role of $D_1$ (size $m_0 \times n_0$) is to invert the input embedding, i.e., $C_0 = D_1^*$ (where $D_1^*$ denotes an approximate right inverse of $D_1$, as $D_1$ is generally non-square \cite{invert}). If this holds, then since $W_1 \approx C_1 D_1$, we see that $W_1 C_0 \approx C_1 D_1 C_0 = C_1 D_1 D_1^* \approx C_1$.

    \item \textbf{Iteration (Layer $i > 1$):} Assume we have successfully estimated $C_{i-1}$. Our goal is to estimate $C_i$ from $W_i \approx C_i D_i$. Unlike Step 1, we cannot simply assume $D_i \approx C_{i-1}^*$ because a non-linearity and bias ($\sigma_{i-1}$) were applied after the computation involving $C_{i-1}$.
    The central hypothesis for this step is  that by analyzing the estimated $C_{i-1}$ (size $n_{i-1} \times m_{i-2}$), we can potentially deduce the $m_{i-1}$ features computed by layer $i-1$ and how they are represented by the $n_{i-1}$ neuron activations (after bias and ReLU). This understanding of the layer $i-1$ output features could allow us to construct an estimate for $D_i$ (and hence its approximate inverse $D_i^*$). If we can find $D_i^*$, we can then estimate $C_i \approx W_i D_i^*$.
    Repeating this process for $i = 2, \ldots, d$ would, in principle, decode the entire network's feature circuit via $C_1, \ldots, C_d$.
\end{enumerate}

Developing the precise methods to infer the layer $i-1$ feature vectors from $C_{i-1}$ and construct the corresponding $D_i$ (or $D_i^*$) remains a significant challenge and is a key direction for future research. This cascading approach, however, provides a conceptual roadmap for systematically disentangling deep MLPs.\footnote{Note that we are proposing an algorithmic framework without having the solution algorithm, apart from some initial heuristic attempts. Skeptics might criticize us for this. We note that this has been done before, for example, in the context of public key cryptography \cite{diffiehellman} where Diffie and Hellman proposed a public key cryptography algorithm without the RSA algorithm \cite{RSA} to implement it.} 
We also point out that an alternative approach would be to use a sparse autoencoder to determine the feature encoding being input to layer $i$, and use that to construct the pseudo-inverse $D_i^*$.  In fact, the decode matrix of the sparse autoencoder may itself serve as the pseudo-inverse $D_i^*$, since it essentially performs the same function for layer $i$ as an initial embedding matrix does for layer 1.

\subsection{Cascading Feature Disentanglement with an Embedding Layer}

\begin{figure}[hbt!]
  \centering
  \begin{subfigure}[b]{1.1\textwidth}
    \centering
\includegraphics[trim=1.7in 0.0in 0.3in 0.0in, clip=true, 
width=\textwidth]{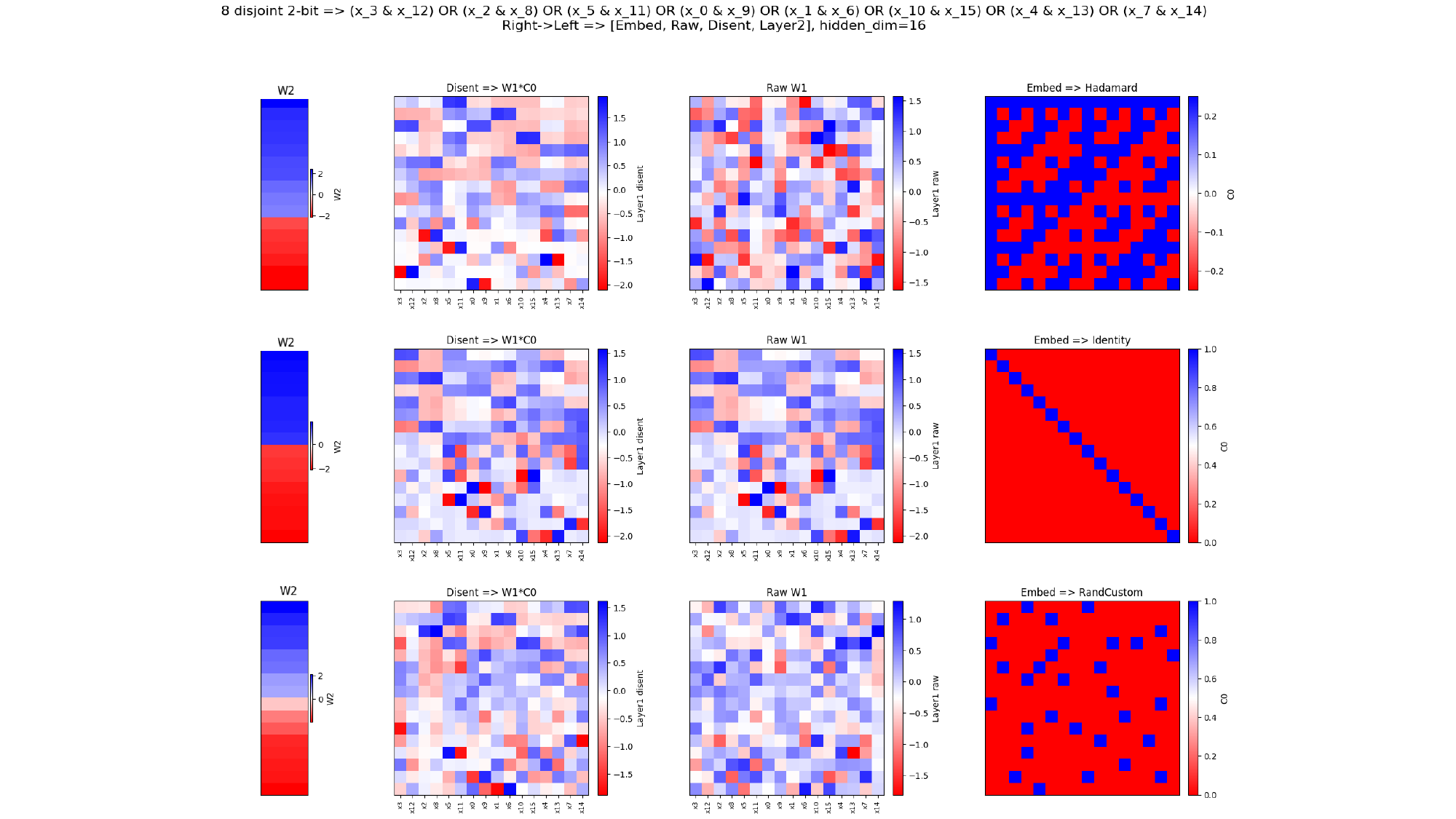}
  \end{subfigure}
  \caption{Disentangling of networks computing the feature circuit depicted in Figure~\ref{Feature Circuit 1}. In top a 2-layer network with an initial Hadamard embedding layer $C_0$ that is trained as before on a DNF of 8 different 2-AND clauses. The raw trained $W_1$ matrix is shown immediately to the left of it. Then to its left we show the disentangled $C_1 = W_1 C_0$. Notice that $C_1$ has the usual expected structure with coding of 2P in the positive witness rows and coding for a XOR using 1N1P in the negative witness rows. Moreover, though having less 2P patterns and positive rows, it has a similar structure as the case where there is no embedding (the embedding is the identity matrix) as shown in the middle row of the figure. The bottom row shows how a symmetric random embedding is  disentangled in a similar fashion.}
  \label{disentangled heat map}
\end{figure}

We demonstrate that that Cascading Disentanglement works for a small example network where we add an embedding layer to the two layer networks we have described earlier in the paper. Specifically,  Figure~\ref{disentangled heat map} depicts three example networks trained via gradient descent, where three different types of embeddings are added to the feature circuit depicted in Figure~\ref{Feature Circuit 1}.  Each row of matrices is a different type of embedding, where the rightmost column depicts the embedding being used: 
in the middle row, the embedding $C_0$ is the identity function (i.e., no embedding), in the top row it is a Hadamard matrix, and the bottom row it is a symmetric random embedding. 

The middle column depicts the raw matrix $W_1$, i.e., prior to disentanglement. In the case of the identity embedding (middle row) feature channel coding is clear. In the case of the Hadamard and random embeddings, the $W_1$ matrices look random.  The leftmost column depicts the result of disentanglement: the matrix obtained by the product $W_1C_0$. 
As described above, this gives us $C_1$ since $W_1 C_0 \approx C_1 D_1 C_0 \approx C_1 D_1 D_1^* \approx C_1$.
This reveals the feature channel encoding hidden within $W_1$ for both the Hadamard embedding and the random embedding.  In other words, multiplying by $C_0$ undoes the implicit embedding that was learned by the network in $W_1$, and reveals the feature channel codes.

This is a good place to pause and contemplate the power of disentanglement. The relative magnitudes of the weights in the raw $W_1$ matrices, before disentanglement, might be quite misleading. For example, in the Hadamard embedding example, it would seem from $W_1$ that for Neuron 3, the weights in columns 2 and 15 are the most important (as they are of the largest magnitude), and yet we know from the disentangled matrix that these weights are just the result of the embedding space: the weights in these columns do not explain what neuron 3 is contributing to the computation. From the disentangled matrix $C_1$ we can tell that Neuron 3 is part of the coding for features $(x_2 \land x_{12})$ and $(x_0 \land x_{9})$. This is the power of combinatorial disentanglement: if we want to decide which weights or neurons to prune, which to apply dropout to during training and so on, doing so in the disentangled space would give us the insights that are missing when using activation space approaches. 

In the above experiments, we have used three different types of embeddings: identity (or no embedding), using a Hadamard matrix, and using a symmetric random embedding. In most of the statistics in Section~\ref{section: scaling} we trained on a network where the input had a Hadamard embedding layer $C_0$ which we then used to get $W_1*C_0$, the coding matrix in which we counted the number of patterns (as a sanity check, we also ran the same training without the Hadamard embedding, getting similar results). In all other sections we used the identity embedding.

Finally, one can understand today's prevailing approach to interpreability via the training of an 
autoencoder~\cite{anthropic2023autoencoder} using our coding-decoding view of computation. The autoencoder approach is a way of learning an approximation of $D_i$ for a given layer $i$. The autoencoder has a very large inner layer to which the output activations of $C_{i-1}$ are mapped, along with a sparsity constraint to separate them so that the features of layer $i-1$ can be read. The difference in the combinatorial approach we propose here is to attempt to decode $C_{i-1}$ and reconstruct $D_{i}$ by analyzing the network weight matrices themselves rather then learning features from the activations using a separate trained network. As mentioned above, we think there is potential in a combined approach, for example where $D_{i}^*$ is learned via an autoencoder (or other activation based learning approaches~\cite{transcoders}), and then this is used to find and interpret $C_i$ combinatorially.  This would allow our approach to be applied to a single layer, without the need for cascading.


\section{Towards Deeper Networks}
\label{section multioutput}

As a bit of a teaser as to the potential directions one can follow with the combinatorial interpretability approach, we examine what happens if there are multiple outputs - i.e., a classification problem with more than 2 possibilities. To do so, we add additional Boolean formulas to learn, and for each additional formula, there is a corresponding additional second layer output neuron responsible for that formula.  The inputs and the first hidden layer are shared between the different outputs, and so the output neurons are computing different formulas on the same set of Boolean variables and on the same hidden layer results.  We still only consider the case where there is a single hidden layer, but we view increasing the size of the output layer as a small step towards having an additional hidden layer, as view computing multiple output features as a potential proxy for computing multiple features for use at the next hidden layer of the network.

We first consider the case with two output neurons.  We tested a range of clause numbers, and found something similar to the single output case: the system is able to train fairly successfully, provided that the total number of clauses (in this case, summed across all outputs) is not larger than the hidden dimension.   We here provide results for one representative example: in this case there are 32 Boolean variables and a hidden dimension of 32.  The formulas are in DNF form with clauses of size 4 (4-AND), and each output is the OR of 8 clauses.  The clauses for each output are non-overlapping in terms of variables, but the clauses for one output are chosen without taking into account the other output.  Thus, each variable is used exactly once in each output but can be used in both outputs.

We generated 50000 inputs, each having between 8 and 10 variables set to true.  The inputs were constructed independently, and each had equal likelihood of being each of the 4 output combinations (00, 01, 10, 11).  Each output possibility was constructed in the same style as used in Section \ref{section: scaling}.

We provide the full neural network parameterization for the following pair of formulas:

\begin{multline*}
f_0(x) =
(x_3 \land x_4 \land x_7 \land x_{10}) \lor
(x_2 \land x_6 \land x_{25} \land x_{27}) \lor
(x_0 \land x_{13} \land x_{30} \land x_{31}) \lor \\
(x_9 \land x_{16} \land x_{20} \land x_{21}) \lor
(x_8 \land x_{12} \land x_{14} \land x_{28}) \lor
(x_1 \land x_5 \land x_{17} \land x_{24}) \lor \\
(x_{11} \land x_{22} \land x_{26} \land x_{29}) \lor
(x_{15} \land x_{18} \land x_{19} \land x_{23})
\end{multline*}

\begin{multline*}
f_1(x) =
(x_{10} \land x_{15} \land x_{23} \land x_{26}) \lor
(x_9 \land x_{13} \land x_{18} \land x_{19}) \lor
(x_0 \land x_4 \land x_8 \land x_{14}) \lor \\
(x_6 \land x_{20} \land x_{28} \land x_{31}) \lor
(x_3 \land x_{21} \land x_{22} \land x_{30}) \lor
(x_2 \land x_{11} \land x_{12} \land x_{24}) \lor \\
(x_1 \land x_5 \land x_{25} \land x_{29}) \lor
(x_7 \land x_{16} \land x_{17} \land x_{27})
\end{multline*}

\begin{figure}[hbt!]
  \centering
  \begin{subfigure}[b]{0.45\textwidth}
    \centering
    \includegraphics[width=\textwidth]{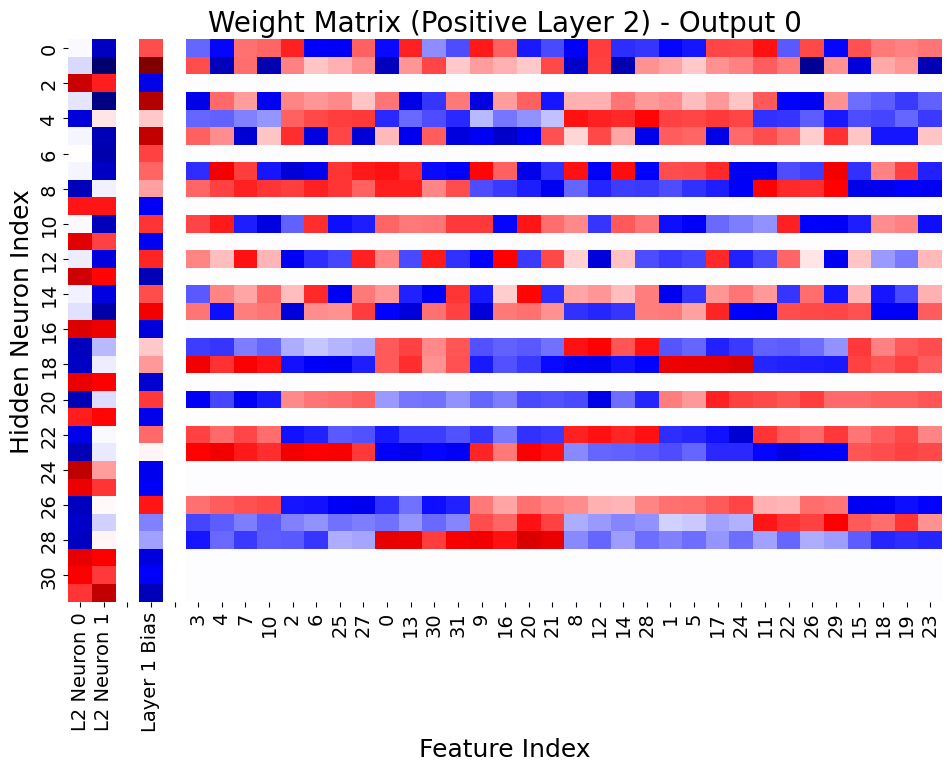}
  \end{subfigure}
  \hfill
  \begin{subfigure}[b]{0.523\textwidth}
    \centering
    \includegraphics[width=\textwidth]{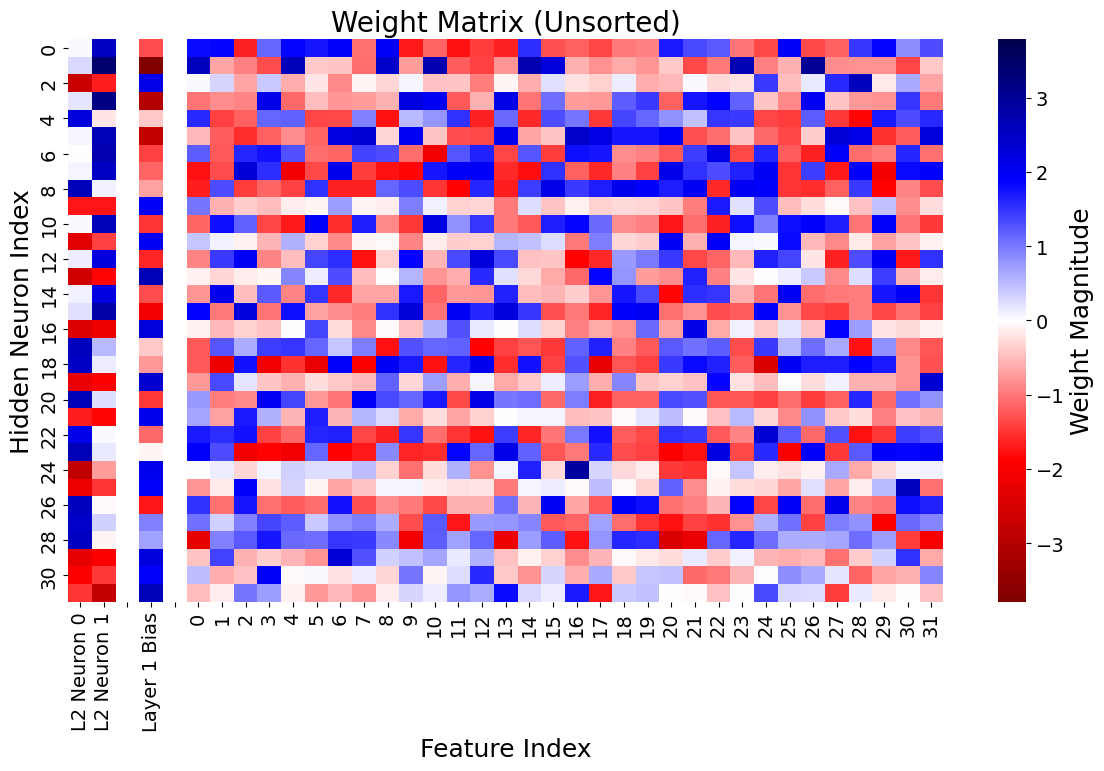}
  \end{subfigure}
  \begin{subfigure}[b]{0.45\textwidth}
    \centering
    \includegraphics[width=\textwidth]{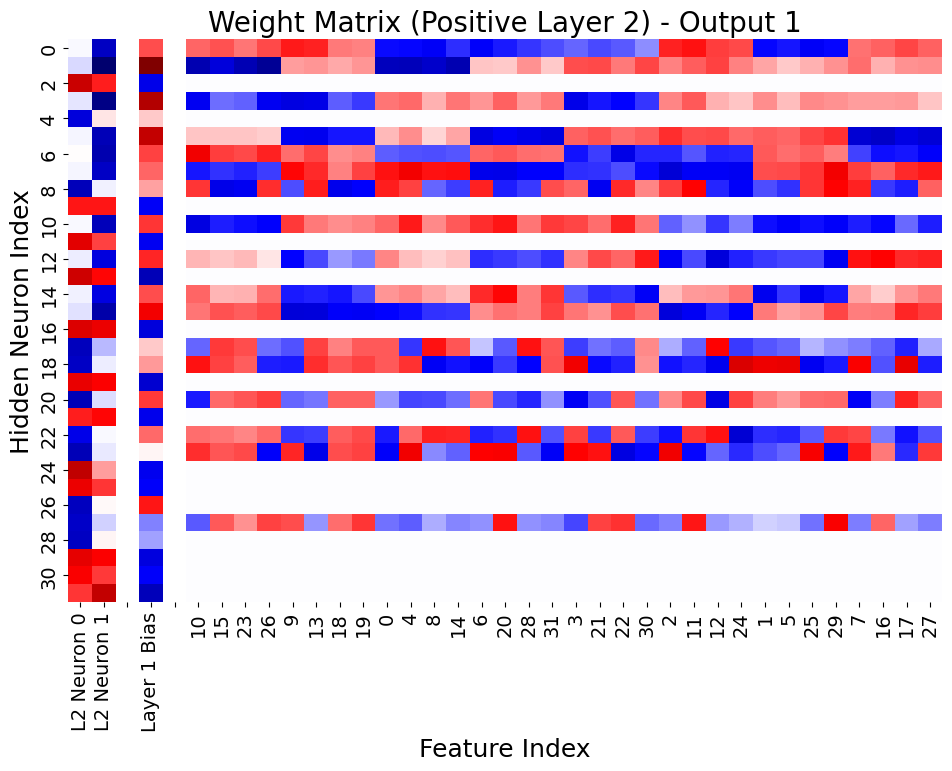}
  \end{subfigure}
  \hfill
  \begin{subfigure}[b]{0.523\textwidth}
    \centering
    \includegraphics[width=\textwidth]{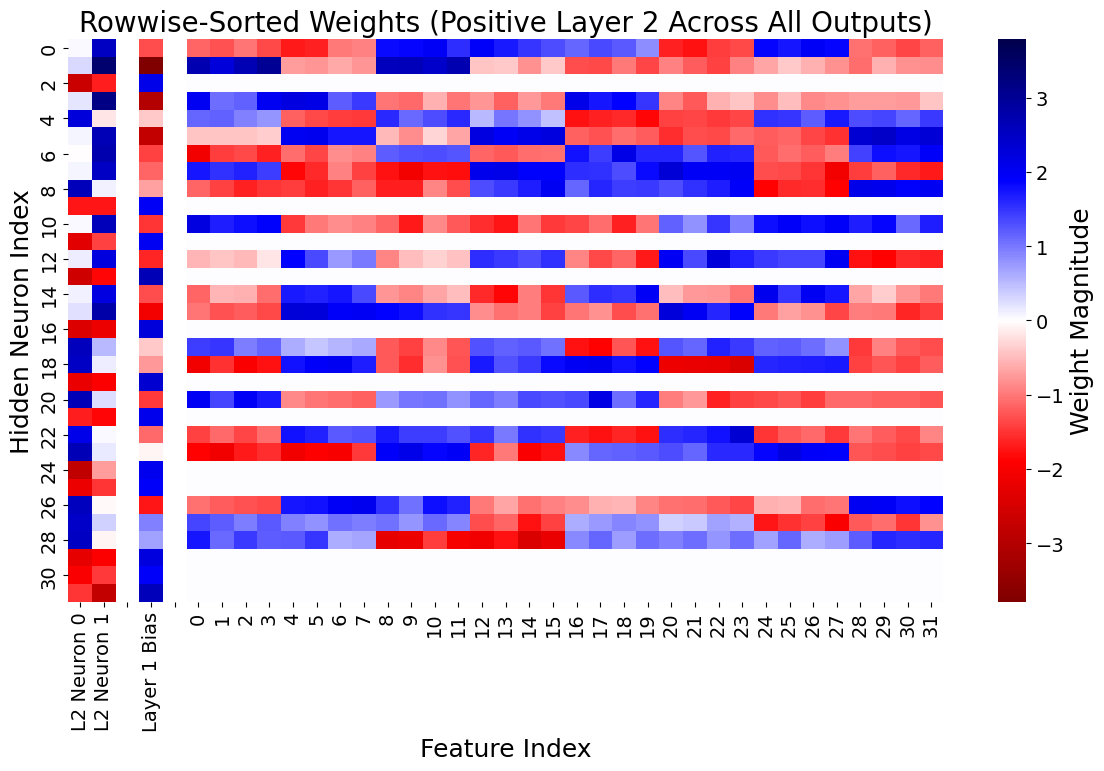}
  \end{subfigure}
  \caption{Trained neural network for $f_0$ and $f_1$.  The top right is the unsorted version of the weight matrix, with all weights and biases.  The top left is sorted by the clauses of $f_0$ (computed by output 0), and, as above, filtered to show only those neurons that have a positive edge to output neuron 0.  The bottom left is the equivalent version for output neuron 1 and $f_1$.  The bottom right has all rows with a positive edge to either output neuron, and each such row is sorted by the clauses corresponding to the output neuron with the larger weight to that row/hidden layer neuron.}
  \label{multi 2}
\end{figure}

This computation is depicted in Figure \ref{multi 2}.  This network achieved training loss of 3.42\% after 1000 epochs, and at that time had overall accuracy (both bits correct) on the test set of 97.30\%.  Not only did the network learn this network very clearly, we see clear fature channel coding, despite having to do that coding for two different Boolean functions simultaneously.

A few observations about this parameterization:
\begin{itemize}
\item The biases are right in line with our expectations: if {\em either} output neuron has a positive weight to a hidden layer neuron, then the bias for the hidden layer neuron is negative.  (Neurons 27 and 28 are exceptions and slightly positive.)  All rows where both output layer weights are negative have a positive hidden layer bias.
\item  The positive row filtering for a given neuron sometimes codes exactly as expected, and other times it seems random.  However, on closer examination, in every single case where it seems random, the weight from the other output neuron to that row is significantly larger.  It is really coding for the other neuron in those rows, and simply looks random because we are sorting by the wrong set of clauses.
\item The output neuron weights partition the hidden layer neurons into three categories: (1) positive and larger weight from output neuron 0, (2) positive and larger weight from output neuron 1, and (3) negative from both neurons.  Neurons of type (1) code for output neuron 0 and function $f_0$.  Neurons of type (2) code for  neuron 1.  Neurons of type (3) prove negative witnesses.  This is visualized clearly in the bottom right panel of Figure \ref{multi 2}.  There, each row is sorted according to which cell of the partition is in: for rows of type (1), we sort by the clauses of $f_0$, for rows of type (2), we sort by the clauses of $f_1$, and rows of type (3) are filtered out.  We see that the result is nearly perfect coding.
\item Once the system has chosen which output a row is coding for, it effectively ignores the other output for that row - the other weight has a much smaller magnitude weight, and that weight has a tendancy to be (slightly) positive.
\item In Section \ref{section: scaling}, we saw that even as we increased the number of clauses, the network was not able to learn a weight distribution with more than roughly half positive rows. We here see a fairly even divide between the three cells of the partition, and so the number of negative rows is only 12.  This is additional evidence that the system does not need as many negative rows as the solution that it is converging to has.  For this case where there are 2 output neurons, one could also argue that it is seeking equality between the positive rows and the negative rows, and the negative rows are reused between the two output neurons.  However, (a) we do not see that same effect with 3 output neurons (below), and (b) even if that were the case, it does imply a limit on how much logic can be built into the negative rows (since they are the same between the two outputs, with relatively minor differences in how negative the output layer weight is).
\end{itemize}

\begin{figure}[hbt!]
  \centering
  \begin{subfigure}[b]{0.45\textwidth}
    \centering
    \includegraphics[width=\textwidth]{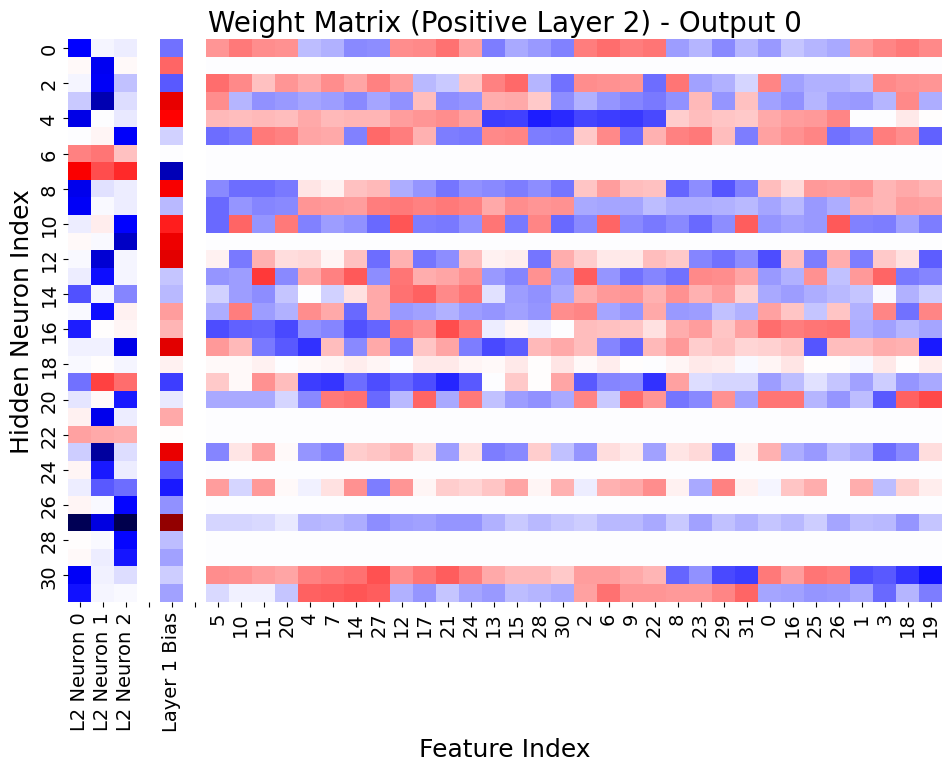}
  \end{subfigure}
  \hfill
  \begin{subfigure}[b]{0.523\textwidth}
    \centering
    \includegraphics[width=\textwidth]{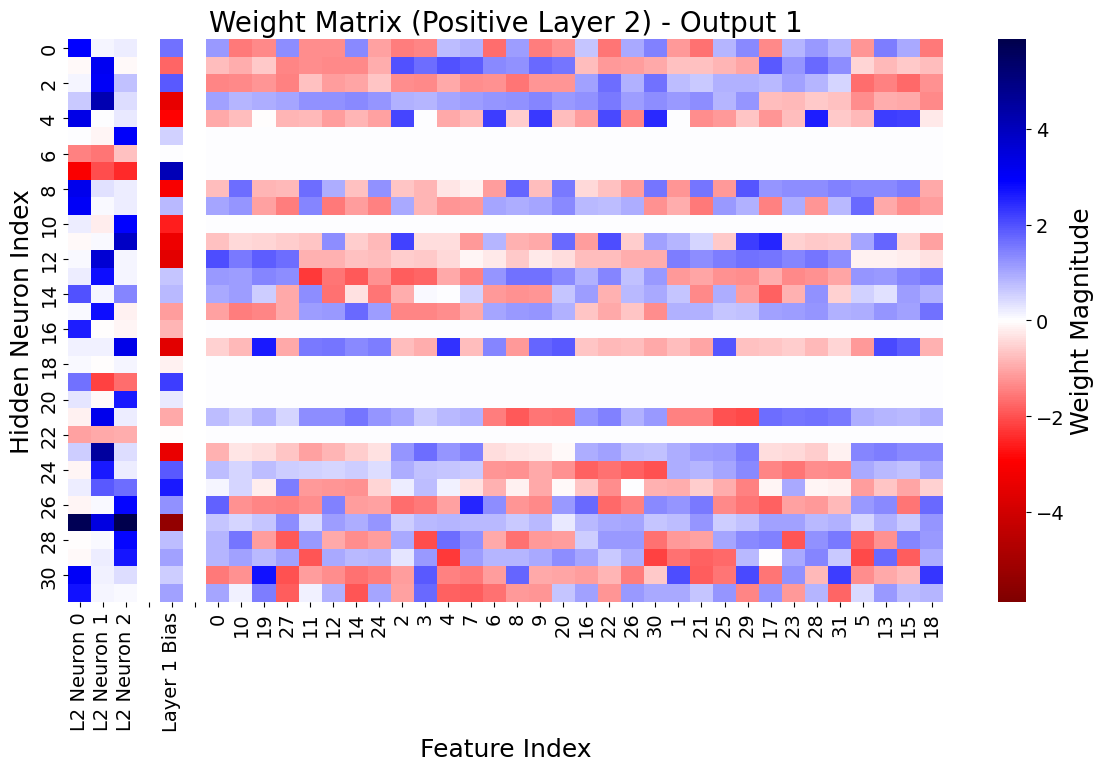}
  \end{subfigure}
  \begin{subfigure}[b]{0.45\textwidth}
    \centering
    \includegraphics[width=\textwidth]{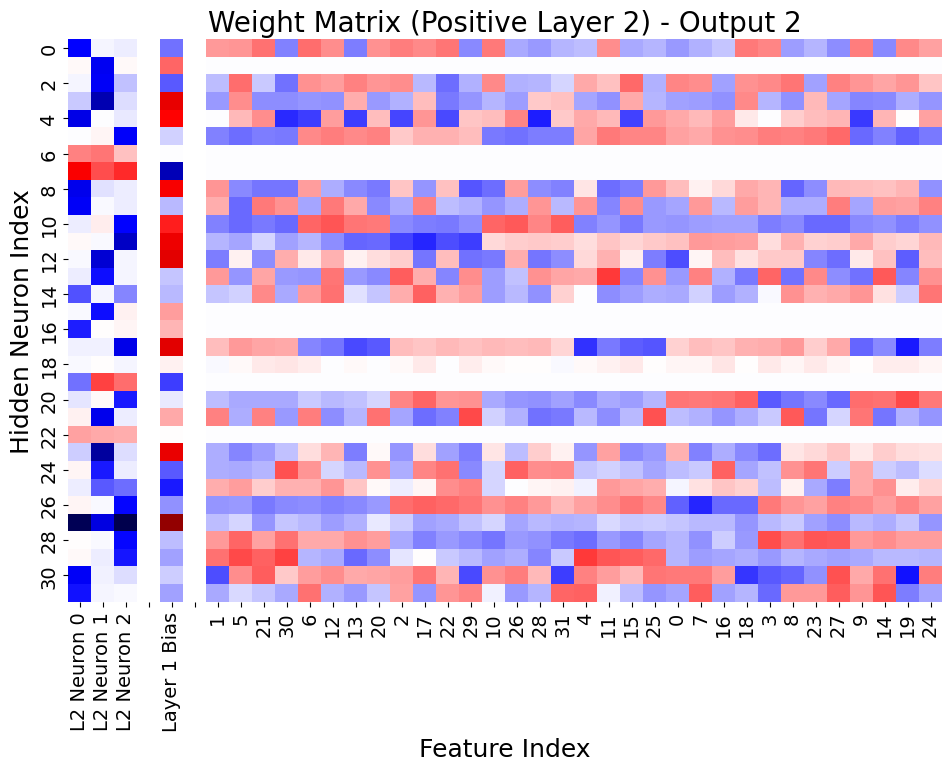}
  \end{subfigure}
  \hfill
  \begin{subfigure}[b]{0.523\textwidth}
    \centering
    \includegraphics[width=\textwidth]{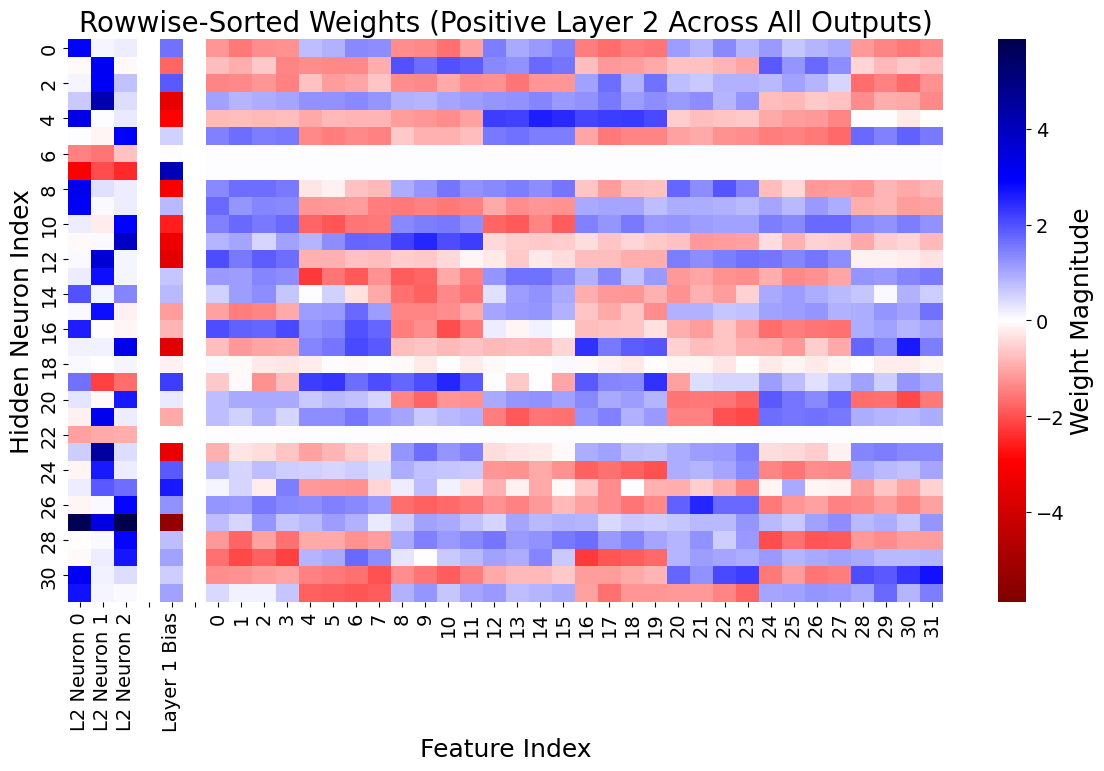}
  \end{subfigure}
  \caption{Trained neural network for $g_0$, $g_1$, and $g_2$.  The top right is sorted by the clauses of $g_0$ (computed by output 0), and filtered to show only those neurons that have a positive edge to output neuron 0.  The top left is the equivalent version for output neuron 1 and $g_1$, and the bottom left is the equivalent for output neuron 2 and $g_2$.  The bottom right has all rows with a positive edge to any output neuron, and each such row is sorted by the clauses corresponding to the output neuron with the largest weight to that row/hidden layer neuron.}
  \label{multi 3}
\end{figure}

We next turn to the case where we have 3 output neurons.  The setup is the same otherwise, except that we now train with more data (200,000 inputs).  The result was slightly worse performance for this more difficult case (since we now have 24 total clauses, instead of 16).  Specifically, after 1000 epochs, we have training loss 5.86\% and test set accuracy (all 3 outputs correct) of 95.05\%.  Figure \ref{multi 3} depicts the results for the following set of formulas:

\begin{multline*}
g_0(x) =
(x_5 \land x_{10} \land x_{11} \land x_{20}) \lor
(x_4 \land x_7 \land x_{14} \land x_{27}) \lor
(x_{12} \land x_{17} \land x_{21} \land x_{24}) \lor \\
(x_{13} \land x_{15} \land x_{28} \land x_{30}) \lor
(x_2 \land x_6 \land x_9 \land x_{22}) \lor
(x_8 \land x_{23} \land x_{29} \land x_{31}) \lor \\
(x_0 \land x_{16} \land x_{25} \land x_{26}) \lor
(x_1 \land x_3 \land x_{18} \land x_{19})
\end{multline*}

\begin{multline*}
g_1(x) =
(x_0 \land x_{10} \land x_{19} \land x_{27}) \lor
(x_{11} \land x_{12} \land x_{14} \land x_{24}) \lor
(x_2 \land x_3 \land x_4 \land x_7) \lor \\
(x_6 \land x_8 \land x_9 \land x_{20}) \lor
(x_{16} \land x_{22} \land x_{26} \land x_{30}) \lor
(x_1 \land x_{21} \land x_{25} \land x_{29}) \lor \\
(x_{17} \land x_{23} \land x_{28} \land x_{31}) \lor
(x_5 \land x_{13} \land x_{15} \land x_{18})
\end{multline*}

\begin{multline*}
g_2(x) =
(x_1 \land x_5 \land x_{21} \land x_{30}) \lor
(x_6 \land x_{12} \land x_{13} \land x_{20}) \lor
(x_2 \land x_{17} \land x_{22} \land x_{29}) \lor \\
(x_{10} \land x_{26} \land x_{28} \land x_{31}) \lor
(x_4 \land x_{11} \land x_{15} \land x_{25}) \lor
(x_0 \land x_7 \land x_{16} \land x_{18}) \lor \\
(x_3 \land x_8 \land x_{23} \land x_{27}) \lor
(x_9 \land x_{14} \land x_{19} \land x_{24})
\end{multline*}

This solution is very similar to the case with 2 output neurons, and as we can see in Figure \ref{multi 3}, it is clearly using feature channel coding.  However the structure of the network is not quite as clean.  This is not surprising, given (a) the poorer performance, and (b) the fact that we are starting to approach the limit we discussed in Section \ref{section: scaling}, where the total number of clauses equals the size of the hidden dimension.  Specifically, we see that the layer 1 biases are not as consistently negative for the positive rows, and that not every positive row chose a clear output to code for (see neuron 27, which just contributes a bit of positive value from all input variables.  It is however, interesting to note that the number of negative rows is only 3 (or 4 if you count row 19), leaving each output about 9 rows of positive coding, again providing evidence that the single output solutions are over investing in negative rows.

We did also experiment with 4 output neurons, but our initial attempts at that did not yield particularly effective networks.  In that case the number of clauses was equal to the size of the hidden layer; the trained networks tended to have error rates around 1/3.  Still even in these poorly performing networks, it was clearly trying to use feature channel coding.

\section{Further Research}
We believe the combinatorial approach and its first application through feature channel coding, can have implications even before we have the ability to apply it to large real world networks. In particular, in the same way we used it to understand scaling laws, one can attempt to understand other important neural computation aspects such as sparsity, quantization, and the linear representation hypothesis, to name a few. Understanding to what extent computation in real production neural networks contains Boolean features and tehir codes is also of great interest. Finally, it has not escaped our notice that the specific coding we have postulated immediately suggests a possible computation mechanism for neural tissue. 


\bibliographystyle{plain}
\bibliography{bibliography}
\end{document}